\documentclass[11pt]{article}
\usepackage{fullpage,amsmath,amssymb,bm,url,mathrsfs}
\usepackage[round,colon,authoryear]{natbib}
\usepackage{algorithm}
\usepackage{amsthm}
\usepackage{hyperref}
\usepackage{comment}
\usepackage{enumitem}
\usepackage{bbm}
\hypersetup{
    colorlinks=true,
    linkcolor=blue,
    citecolor=blue,
    filecolor=magenta,      
    urlcolor=cyan,
}
\RequirePackage{epsfig, graphicx, color}

\usepackage{stmaryrd}

\usepackage{amsmath}
\DeclareMathOperator*{\argmax}{arg\,max}
\DeclareMathOperator*{\argmin}{arg\,min}
\usepackage{stmaryrd}

\usepackage{graphics,graphicx}
\usepackage{subcaption}
\usepackage{xcolor,crossreftools}

\usepackage{algorithm}
\usepackage{algpseudocode}

\newcommand{\abs}[1]{\lvert#1\rvert}

\def\calE{{\cal  E}} 
 
\def\calG{{\cal  G}}

\def\calK{{\cal  K}} 
 
\def\calM{{\cal  M}}

\def\calV{{\cal  V}} 
 
\def\calX{{\cal  X}}

\newcommand{\bfm}[1]{\ensuremath{\mathbf{#1}}}

   \def\bA{\bfm A}  
     
   \def\bC{\bfm C}  
     
     \def\EE{\mathbb{E}}

     \def\OO{\mathbb{O}}

\def\br{\bfm r}     \def\RR{\mathbb{R}}
     \def\SS{\mathbb{S}}

   \def\bX{\bfm X}

  \def\bTheta{\bfm \Theta}

\def\hat{\widehat}

\newcommand{\RN}[1]{%
  \textup{\uppercase\expandafter{\romannumeral#1}}%
}
\newcommand\fro[1]{\| #1 \|_{\rm F}}
\newcommand\op[1]{\|#1\|}

\def\bcalV{\boldsymbol{\mathcal{V}}}
\def\bcalG{\boldsymbol{\mathcal{G}}}
\def\bcalE{\boldsymbol{\mathcal{E}}}

\newtheorem{theorem}{Theorem}
\newtheorem{remark}{Remark}

\newtheorem{assumption}{Assumption}
\newtheorem{lemma}{Lemma}
\newtheorem{corollary}{Corollary}

\theoremstyle{remark}

\newcommand{\E}{\mathbb{E}}
\newcommand{\Prob}{\mathbb{P}}
\newcommand{\R}{\mathbb{R}}
\newcommand{\cM}{\mathcal{M}}

\newcommand{\oLambda}{\overline{\Lambda}}

\newcommand\numberthis{\addtocounter{equation}{1}\tag{\theequation}}

\begin{document}
\title{Latent Space Model for Higher-order Networks and Generalized Tensor Decomposition\footnote{Zhongyuan Lyu and Dong Xia's research was partially supported by Hong Kong RGC Grant ECS 26302019 and GRF 16303320.}}
\author{Zhongyuan Lyu$^{\dagger}$, Dong Xia$^{\dagger}$ and Yuan Zhang$^{\ddagger}$\\
{\small$^\dagger$Hong Kong University of Science and Technology}\\
{$^\ddagger$}\small Ohio State University}

\date{}
\maketitle

\begin{abstract}
{
	We introduce a unified framework, formulated as general latent space models, to study complex higher-order network interactions among multiple entities.  
	Our framework covers several popular models in recent network analysis literature, including mixture multi-layer latent space model and hypergraph latent space model. 
	We formulate the relationship between the latent positions and the observed data via a generalized multilinear kernel as the link function.
	While our model enjoys decent generality, its maximum likelihood parameter estimation is also convenient via a generalized tensor decomposition procedure.
	We propose a novel algorithm using projected gradient descent on Grassmannians.
	We also develop original theoretical guarantees for our algorithm.
	First, we show its linear convergence under mild conditions. 
	Second, we establish finite-sample statistical error rates of latent position estimation, determined by the signal strength, degrees of freedom and the smoothness of link function, for both general and specific latent space models. 
	We demonstrate the effectiveness of our method on synthetic data.
	We also showcase the merit of our method on two real-world datasets that are conventionally described by different specific models in producing meaningful and interpretable parameter estimations and accurate link prediction.
	
}
\end{abstract}

\section{Introduction}
{\it Networks} \citep{newman2018networks} capturing the {\it dyadic} or {\it pairwise} interactions between a set of entities/vertices have been an active research field for more than half a century, leading to millions\footnote{Google Scholar reports $\sim5.8$ million results to the search query "network analysis".} of publications and technical reports in related disciplines and a wide spectrum of applications.
To date, various aspects of networks, e.g. fundamental theories, statistical models, efficient algorithms and so forth, have been well developed through joint contributions from distinct scientific communities -- physics, computer science, mathematics, statistics, to name a few.  However, the recent decade has witnessed a fast growing demand in processing and analyzing more complex systems where interactions among a set of entities are {\it polyadic} or {\it non-linear}. These complex networks pose fresh challenges on understanding and exploiting the joint interactions among entities. 

The recent boom in data science gives rise to numerous categories of complex networks where relations among entities are far beyond being dyadic. More concretely, we focus on three specific types of complex networks -- {\it multi-layer} networks \citep{kivela2014multilayer}, {\it hypergraph} networks \citep{ghoshal2009random} and {\it dynamic}/{\it temporal} networks \citep{goldenberg2010survey}, each of which is an independent sub-field of study and has tremendous applications. Multi-layer networks arise when two vertices can present multiple types of relations, for instance, friendship networks \citep{dickison2016multilayer,wang2020learning} on LinkedIn, Instagram and Facebook among the same set of people can differ drastically; trading patterns of different commodities \citep{jing2021community, cai2021generalized} among the same set of countries are distinct. Other notable examples of multi-layer networks include brain fMRI images \citep{arroyo2019inference, paul2020random, tang2017robust, le2018estimating, wang2019common},  genetic networks and protein-protein interaction networks \citep{hore2016tensor,larremore2013network, lei2020consistent, zhang2017finding}, transportation networks \citep{cardillo2013emergence,cardillo2013modeling} and etc. Note that the interactions (node $i$, node $j$) in multi-layer networks on each layer are still dyadic. But they can be viewed as polyadic interactions (node $i$, node $j$, layer $l$) if layers are treated as an independent set of entities. Hypergraph networks refer to the complex systems whose vertex interactions are representable by hypergraphs consisting of a set of vertices and a set of hyper-edges.  Each hyper-edge can connect multiple (more than $2$) vertices exhibiting a polyadic relationship among these vertices, say (node $i$, node $j$, node $k$). A hypergraph is said to be $m$-{\it uniform} if every hyper-edge connects exactly $m$ vertices. Unlike the pairwise interaction of an edge, a hyper-edge captures the higher-order interaction which often carries more insightful information. In \citep{benson2016higher}, the authors discover that, by incorporating high-order interactions in the airport network, the spectral clustering algorithm reveals geographic proximity between airports which is unseen if only dyadic relationships are used. Hypergraph networks are typically observed in co-authorship networks \citep{cai2021generalized, ji2016coauthorship,newman2011structure}, legislator network \citep{ke2019community,lee2017time}, proton emission networks \citep{zhen2021community}, circuit networks \citep{ghoshdastidar2014consistency} and so on. Lastly, dynamic/temporal networks \citep{wang2018optimal} naturally model dynamic systems where interactions between the same set of vertices evolve through time. They resemble multi-layer networks in the sense that layers are now indexed in a meaningful order, such as a discrete time flow. At a fixed time point, the relationship between vertices is still pairwise. Clearly, dynamic networks can be treated as networks of polyadic interactions, say (node $i$, node $j$, time-stamp $t$), if the discrete time flow is viewed as a separate set of entities. Typical examples of dynamic networks include, for instance, the Senate cosponsorship network \citep{wang2017fast}, Enron email network \citep{park2012anomaly, wang2013locality}, social interactions between animals \citep{matias2015statistical} and student friendship network \citep{chen2015graph}.

The goal of this paper is to investigate the aforementioned complex and unweighted networks -- (mixture) {\it multi-layer} networks, {\it hypergraph} networks and {\it dynamic}/{\it temporal} networks in a unified framework. Since these networks all involve joint interactions of multiple entities, we collectively refer to these networks as {\it higher-order networks}. We note that, during the preparation of this work, the same concept was also coined by \citep{bick2021higher}.

Stochastic block model (SBM)  \citep{holland1983stochastic} is a prevalent approach for modelling the latent group structures of vertices in {\it networks}. At the core of SBM is the assumption that vertices belonging to the same group are {\it stochastically equivalent}. The group structure of SBM intrinsically impose low-rank constraint on the expected adjacency matrix which naturally popularizes the spectral methods \citep{rohe2011spectral,lei2015consistency, zhang2016community, zhang2020detecting}. Undoubtedly, numerous variants of SBM have been proposed to treat higher-order networks. The multi-layer SBM was proposed in \cite{lei2020consistent, paul2020spectral, arroyo2019inference} assuming the same group assignments across all layers. A random effect multi-layer SBM was proposed in \cite{paul2020random} allowing for heterogeneous group assignments for different layers. More recently, \cite{jing2021community} introduced a novel mixture multi-layer SBM to simultaneously cluster networks and identify global and local group memberships of vertices. Among these prior works, the low-rankness of adjacency matrix and tensor is the primary ingredient in their methods. Similarly, hypergraph SBM was proposed and theoretically investigated in \cite{ghoshdastidar2015provable,ghoshdastidar2017consistency,chien2018community,kim2018stochastic,pal2019community,yuan2018testing}, where the expected adjacency tensor admits a low-rank decomposition. Meanwhile, \cite{ke2019community} introduced a degree corrected hypergraph SBM to accommodate the degree heterogeneity commonly observed in practice. The authors also proposed a low-rank tensor-based spectral method for community detection. For modelling the group structures in dynamic networks, SBM is also much favored. For instance, \cite{pensky2019dynamic, pensky2019spectral} studied a dynamic SBM model and a spectral method for community detection. Aside from vertices clustering, another practically relevant problem in dynamic SBM is to detect change points in the sense that, for example, when network structure suddenly shifts. See, e.g., \cite{park2012anomaly,wang2013locality,wilson2019modeling,wang2017fast, wang2018optimal} for more details. All the aforementioned SBM extensions were designed for treating high-order networks. Without loss of generality, we will collectively refer to them as the {\it high-order SBM}.

Higher-order SBM enjoys structural simplicity, motivates diverse new statistical methods and demonstrates effective performances in identifying clusters. However, the stringent model assumptions of SBM may hamper or even jeopardize its effectiveness in handling more general higher-order networks. First of all, SBM enforces transitivity ($i \textrm{ connects to } j, j \textrm{ connects to } k\Rightarrow i \textrm{ connects to } k$ with high probability) via the cluster structure, i.e., nodes in the same cluster tend more likely to connect. However, such strong clustering phenomenon may not be prevalent, especially in high-order networks. Recent advances in analyzing multi-layer networks, such as change point detection in dynamic networks, no longer limit themselves to block model structures \citep{wang2018optimal}. Secondly, higher-order SBM usually makes the impractical assumption that nodes in the same cluster are stochastically {\it equivalent}. As an example, the trading flows of commodities between countries in Section~\ref{sec:real-data}; even though China, Germany and USA share similar trading patterns of industrial commodities with other countries, and are identified as being {\it close} by a clustering algorithm, they clearly should not be regarded as {\it equivalent} in view of the striking technological gaps between these three economies. Finally, due to the linear relations, higher-order SBM usually results into an expected adjacency tensor admitting a low-rank decomposition \citep{ke2019community, jing2021community}. Unfortunately, oftentimes, the observed adjacency tensor presents many moderate-magnitude singular values rendering the low-rank presumption questionable. 

As argued in \cite{hoff2002latent}, the transitivity of relations in networks may be better characterized by the proximity between vertices in an {\it unobserved} latent space, where each entity/vertex is associated with a vector of characteristics, named {\it latent position}, in this space.  It is therefore referred to  as the latent space model (LSM). Compared with SBM, the learned latent features from LSM \citep{ma2020universal, levin2017central, zhang2020flexible, macdonald2020latent} is sometimes more useful in downstream tasks such as node visualization, link prediction and community detection. Meanwhile, LSM allows for non-linear relations with a general link or kernel function. In this paper, we propose a unified framework based on LSM to treat higher-order networks -- thus the name {\it higher-order latent space model} (hLSM). Without loss of generality, we focus on higher-order networks with triadic interactions among vertices. Let $\calV_1, \calV_2, \calV_3$ be three sets of ``vertices"  so that a triadic interaction of vertices $i_1\in \calV_1, i_2\in\calV_2, i_3\in\calV_3$ is notationally regarded as a tuple $(i_1,i_2,i_3)$. We emphasize the abstraction of ``vertices" in our framework since they can stand for conceptually different subjects in different contexts. In a hypergraph network, $\calV_1, \calV_2, \calV_3$ are the same set of vertices and the tuple $(i_1,i_2,i_3)$ just represents a hyper-edge connecting the three vertices. For a multi-layer or dynamic network, $\calV_1$ and $\calV_2$ can be the same set of vertices while $\calV_3$ is viewed as the index set of layers or time-stamps, respectively. Underlying our hLSM is the major assumption that each vertex $i_k\in\calV_k$, for $k=1,2,3$, is associated with a latent position in a {\it low-dimensional} space $\calX_k$. Conditioning on the latent positions, hLSM assumes that three vertices positioned $u_{i_1}^{\ast}, v_{i_2}^{\ast}, w_{i_3}^{\ast}$ would form triadic interaction $(i_1, i_2, i_3)$, independently of others, with probability $\rho(u_{i_1}^{\ast}, v_{i_2}^{\ast}, w_{i_3}^{\ast})$. Here, $\rho(\cdot): \calX_1\times \calX_2\times \calX_3\mapsto [0,1]$ is called the kernel function of hLSM. 
The latent positions are treated as fixed points for all vertices whereas we note that our framework can be easily generalized to the case of random latent positions \citep{athreya2017statistical}. The central task in hLSM is to estimate the latent positions. This inevitably relies on the identifiability of latent positions and the regularity conditions of the kernel function, which shall be unfolded with more details in Section~\ref{Sec:model}. At last, we remark that many aforementioned higher-order SBM's are special cases of hLSM.  By choosing a linear kernel, hSLM reduces to the multi-layer random dot product graph of \cite{levin2017central}. With a logistic link and shared latent positions, hLSM reproduces the multi-layer LSM of \cite{zhang2020flexible}. The hypergraph embedding model proposed in \cite{zhen2021community} is a special case of hLSM with a joint inner product  of latent positions and a transformed logistic link. A special case of hypergraphon is studied in \cite{Balasubramanian2021nonparametric}.  

We then investigate a unified framework for estimating the latent positions via generalized low-rank tensor decomposition. At the core of our framework is the assumption that the kernel $\rho$ is a generalized multilinear function in the sense that $\rho(u_{i_1}^{\ast}, v_{i_2}^{\ast}, w_{i_3}^{\ast})=g(\langle \bC^{\ast}, u_{i_1}^{\ast}\otimes v_{i_2}^{\ast}\otimes w_{i_3}^{\ast}\rangle)$, where $g(\cdot)$ is a known {\it link function} and $\bC^{\ast}$ is an unknown {\it interaction tensor}.  Under the independent-edge assumption, the adjacency tensor $[\bA]_{i_1,i_2,i_3}\stackrel{{\rm ind.}}{\sim} \textrm{Bernoulli}\big(g([\bTheta^{\ast}]_{i_1,i_2,i_3})\big)$ for an unknown low-rank tensor $\bTheta^{\ast}=\bC^{\ast}\cdot \llbracket U^{\ast}, V^{\ast}, W^{\ast}\rrbracket$. Here $\cdot \llbracket,, \rrbracket$ represents multilinear product, see formal definition in the last paragraph of this section. 
We estimate the latent positions $U^{\ast}, V^{\ast}, W^{\ast}$ via the maximum likelihood estimator which is formulated as a problem of generalized low-rank tensor decomposition. Unfortunately, the objective function is highly non-convex and can be solved only locally. Due to the orthogonality assumptions, the latent positions can be treated as points on Grassmann manifolds. We then propose a projected gradient descent algorithm on the Grassmannians. The algorithm is partially inspired by the tensor completion literature \citep{xia2017polynomial} where its convergence analysis is missing. Here, we investigate this algorithm in a more generalized tensor decomposition framework to treat binary observations. Under mild conditions on the link function, we prove that, even with a constant stepsize, the algorithm converges linearly to a locally optimal solution. This is, to our best knowledge, the first rigorous proof of the fast convergence of the gradient descent algorithm on Grassmannians. Moreover, we also characterize the statistical error of the final estimates of latent positions for general high-order LSM's. The error rate, determined by the signal strength of interaction tensor and the smoothness of the link function, is optimal in terms of the degrees of freedom. These results are applicable to a {\it novel} mixture multi-layer latent space model (MMLSM) and the hypergraph latent space model (hyper-LSM) since they are special cases under our general framework. 
In particular, our framework is capable of detecting heterogeneous latent positions in multi-layer networks and cluster the layers of networks which might admit similar latent positions. Finally, we also apply our method to a simple dynamic latent space model for change point detection.

Our main contributions can be summarized as follows. First, we introduce a general latent space model, called hLSM in short, to characterize {\it polyadic} interactions in higher-order networks, where the participating entities can be real actors in networks  or {\it virtual } ``vertices". Second, in order to treat heterogeneous multi-layer networks, we propose a novel mixture multi-layer LSM. Unlike the existing literature on multi-layer LSM, our model allows distinct latent positions across layers, prevalent in many real-world applications. Other special cases of hLSM, including hypergraph LSM and dynamic LSM, are presented as well. Third, we formulate a general framework to estimate the latent positions by the maximum likelihood estimator, and propose a projected gradient descent algorithm on Grassmannians. We prove that the algorithm converges linearly if initialized well, and establish the statistical error of final estimates for both general and specific hLSM's. Finally, the effectiveness of our algorithm is validated on comprehensive simulations and two real-world datasets. We showcase the merits of latent space models in the tasks of node embedding and link prediction.

\paragraph{Notation and Preliminaries on Tensors} Througout the paper, we use $c,c_0,c_1,\dots$ and $C,C_0,C_1,\dots$ to denote small and large absolute and positive constants, respectively. We write $x\asymp y$ indicating that positive $x$ and $y$ are of same order, i.e., $c y\le x\le C y$. Denote $e_j$ the $j$-th canonical base vector whose dimension might vary, depending on the context. 
For an integer $m$, denote $[m]:=\{1,\cdots,m\}$. Let $\mathbb{O}_{n,p} = \{X\in \mathbb{R}^{n\times p}: X^T X = I_{p\times p}\}$ be the collection of all column-orthonormal $n\times p$ matrices. We use uppercase fonts, e.g., $U,W$, to denote matrices and bold uppercase fonts, e.g., $\bA,\bTheta$, for tensors. Denote the $(i,j,k)$-th entry of $\bA$ by $[\bA]_{i,j,k}$. For any matrix $A$ with ${\rm rank}(A)=r$, let $\sigma_1(A)\ge\sigma_2(A)\ge\cdots\sigma_r(A)>0$ denote its non-zero singular values. Define $\sigma_{\max}(A):=\sigma_1(A)$ and $\sigma_{\min}(A):=\sigma_r(A)$. Denote $\|A\|$, $\|A\|_\infty$ the spectral norm and max norm of the matrix $A$, respectively.  We write $\|A\|_{\rm F}$ ($\|\bA\|_{\rm F}$) for the Frobenius norm of the matrix $A$ (tensor $\bA$). Define $\|A\|_{2,\infty}:=\max_j \|e_j^{\top}A\|$.

For an $n_1\times n_2\times n_3$ tensor $\bA$, its $1$-st matricization (also called unfolding) $\calM_1(\bA)\in \RR^{n_1\times (n_2n_3)}$ is defined by $[\calM_1(\bA)]_{i_1, (i_2-1)n_3+i_3}=[\bA]_{i_1,i_2,i_3}$ for $\forall i_j\in [n_j]$. The $2$-nd and $3$-rd matricization of $\bA$ are defined in a similar fashion. The Tucker ranks of $\bA$ are defined by ${\rm rank}(\bA)=\big({\rm rank}(\calM_1(\bA)), ({\rm rank}(\calM_2(\bA)), ({\rm rank}(\calM_3(\bA))\big)$. Given a matrix $T\in\RR^{r_1\times n_1}$, the multi-linear product, denoted by $\times_1$, between $\bA$ and $U$ is defined by $[\bA\times_1 T]_{i_1, i_2,i_3}=\sum_{j=1}^{n_1} [\bA]_{j, i_2, i_3}[T]_{i_1,j}$ for $i_1\in[r_1]$, $i_2\in[n_2]$ and $i_3\in[n_3]$. The other multi-linear products $\times_2$ and $\times_3$ are defined similarly. 
If $\bA$ has Tucker ranks $(r_1, r_2, r_3)$, there exists an $r_1\times r_2\times r_3$ tensor $\bC$, $U\in\OO_{n_1, r_1}$, $V\in \OO_{n_2,r_2}$ and $W\in\OO_{n_3,r_3}$ such that 
\begin{align}\label{eq:tucker}
\bA=\bC\cdot \llbracket U, V, W\rrbracket:= \bC\times_1 U\times_2 V\times_3 W,
\end{align}
This is often referred to as the Tucker decomposition of $\bA$. 
We use $\oLambda(\bA):=\max\{\|\mathcal{M}_k(\bA)\|,k=1,2,3\}$ and $\underline{\Lambda}(\bA):=\min\{\sigma_{\min}(\mathcal{M}_k(\bA)),k=1,2,3\}$ to denote the largest and smallest singular values of the tensor $\bA$.

\section{Higher-order Latent Space Model}\label{Sec:model}
For ease of exposition, we only present the hLSM for third-order networks, that is, all interactions among ``vertices" are triadic. Its extension to higher-order $(\geq 3)$ networks is conceptually straightforward. 
Without loss of generality, consider that there exist three sets of ``vertices" $\calV_1, \calV_2$ and $\calV_3$ with size $n_k=|\calV_k|$. Here ``vertices" are abstractions of ``actors" in higher-order networks that can stand for even virtual subjects such as the index of layers in multi-layer networks and time-stamps in dynamic networks. 

The observed third-order network is denoted by $\bcalG=(\bcalV,\bcalE)$ with a set of vertices $\bcalV=\{\calV_1, \calV_2, \calV_3\}$ and a set of triadic interactions $\bcalE$.  A triadic interaction is a tuple $(i_1, i_2, i_3)$ with vertex $i_k\in \calV_k$. We say the triadic interaction among the vertices $i_1, i_2, i_3$ occurs if $(i_1,i_2,i_3)\in\bcalE$. The occurrences of distinct triadic interactions are assumed independent akin to the independent-edge random hypergraph \citep{ke2019community}. In hLSM, each vertex is associated with a latent position in an unobserved low-dimensional space characterizing inherent natures of the subjects, e.g. the latent factor for the conservative {\it versus} liberal political ideology of senators \citep{chen2021note}. For any  tuple $(i_1, i_2, i_3)$, let $u_{i_1}^{\ast}\in\RR^{r_1}, v_{i_2}^{\ast}\in\RR^{r_2}$ and $w_{i_3}^{\ast}\in\RR^{r_3}$ be the latent positions of these vertices. Here $r_k$ denotes the dimension of the latent space and it usually does not grow as the network size increases, for instance, the political ideology of a senator can be described by a $2$-dim vector -- conservatism {\it versus} liberalism.  Nevertheless, our framework still applies to the cases where $r_k$ grows with the network size. 

We introduce a \emph{kernel function} $\rho(\cdot): \RR^{r_1}\times \RR^{r_2}\times \RR^{r_3}\mapsto [0,1]$ such that the triadic interaction $(i_1,i_2,i_3)$ is generated with probability $\rho(u_{i_1}^{\ast},v_{i_2}^{\ast},w_{i_3}^{\ast})$. Fixing the kernel function, the connection probability is determined {\it solely} by the latent positions. Denote $\bA\in \{0,1\}^{n_1\times n_2\times n_3}$ the {\it binary} adjacency tensor of $\bcalG$ whose entries are $[\bA]_{i_1,i_2,i_3}={\bf 1}\big((i_1,i_2,i_3)\in\bcalE\big)$. Under hLSM,  we have
\begin{equation}\label{eq:bA-def}
[\bA]_{i_1,i_2,i_3} \stackrel{{\rm ind.}}{\sim} \textrm{ Bernoulli}\big(\rho(u_{i_1}^{\ast}, v_{i_2}^{\ast}, w_{i_3}^{\ast})\big),\quad \forall i_k\in \calV_k
\end{equation}
Denote $U^{\star}=[u_1^{\star},\cdots, u_{n_1}^{\ast}]^{\top}\in\RR^{n_1\times r_1}$ (also $V^{\star}, W^{\star}$ {\it resp.}) the collection of all latent positions of $\calV_1$ (also $\calV_2, \calV_3$, {\it resp.}). By observing the adjacency tensor $\bA$ obeying eq. (\ref{eq:bA-def}), our goal is to estimate the latent positions $U^{\ast}, V^{\ast}$ and $W^{\ast}$. 

The general class of kernel functions is too large to estimate. For simplicity, we assume that $\rho(\cdot)$ is a {\it generalized multi-linear function} in the sense that 
\begin{equation}\label{eq:rho-form}
\rho(u^{\ast}, v^{\ast}, w^{\ast})=g\big(\langle\bC^{\ast}, u^{\ast}\otimes v^{\ast}\otimes w^{\ast}\rangle\big)
\end{equation}
where $\bC^{\ast}\in\RR^{r_1\times r_2\times r_3}$ is an {\it unknown} parameter, called the {\it interaction tensor}, to be estimated. Here $\otimes$ denotes tensor product and $\langle\cdot,\cdot\rangle $ is the Euclidean inner product. The function $g(\cdot)$ is a {\it known} link function, for instance the logistic function $g(x)=(1+e^{-x})^{-1}$ and the probit function $g(x)=\Phi(x)$ where $\Phi(\cdot)$ is the {\it c.d.f.} of standard normal random variable. With eq. (\ref{eq:bA-def}) and (\ref{eq:rho-form}), we write the expected adjacency tensor by
$
\EE \bA= g\big(\bC^{\ast}\cdot \llbracket U^{\ast}, V^{\ast}, W^{\ast}\rrbracket\big),
$
where we slightly abuse the notation and let $g(\cdot):\mathbb{R}\to\mathbb{R}$ also apply entry-wisely on a tensor. If $g(x)=x$ and the latent positions have cluster structures, the model reduces to a higher-order SBM where the expected adjacency tensor admits a low-rank decomposition. Under hLSM with a general link function, $\EE\bA$ can be full rank while $g^{-1}(\EE\bA)$ is low-rank. 
Denote $\bTheta^{\ast}=\bC^{\ast}\cdot \llbracket U^{\ast}, V^{\ast}, W^{\ast}\rrbracket$ and then 
\begin{equation}\label{eq:hLSM}
[\bA]_{i_1,i_2,i_3}\stackrel{{\rm ind.}}{\sim} \textrm{ Bernoulli}([g(\bTheta^{\ast})]_{i_1,i_2,i_3}),\quad \forall i_k\in \calV_k.
\end{equation}
Note that the independence of entries might hold only for a subset of all entries, e.g., the off-diagonal entries for undirected graphs.  
Clearly, $\bTheta^{\ast}$ can be uniquely determined by $\EE\bA$ if the function $g(\cdot)$ is monotonic. However, the latent positions are un-identifiable even with a given $\bTheta^{\ast}$. Without loss of generality, we assume {\it orthonormal} latent positions so that $n_1^{-1}U^{\ast\top}U^{\ast}, n_2^{-1}V^{\ast\top}V^{\ast}$ and $n_3^{-1}W^{\ast\top}W^{\ast}$ are all {\it identity} matrices. We remark that the latent positions sometimes can possess additional structural properties, among which the {\it incoherence} is the most prevailing \citep{jing2021community, ke2019community, han2020optimal, cai2021generalized}. The {\it incoherence constant} of latent position $U^{\ast}$ is defined by
\begin{equation}\label{eq:incoh}
\textsf{Incoh}(U^{\ast}):=n_1^{1/2}\|U^{\ast}\|_{\rm F}^{-1}\cdot \max_{1\leq j\leq n_1}\|e_j^{\top}U^{\ast}\|
\end{equation}
Basically, if $\textsf{Incoh}(U^{\ast})$ is upped bounded by a constant, it implies that the majority rows of $U^{\ast}$ have comparable and small magnitudes. It also means that the information $\bTheta^{\ast}$ carries is fairly spread over all its entries. 
 
 For ease of references, we refer to hLSM($\bC^{\ast}, U^{\ast}, V^{\ast}, W^{\ast}, g(\cdot)$) as the higher-order LSM with parameters $\bC^{\ast}\cdot \llbracket U^{\ast}, V^{\ast}, W^{\ast}\rrbracket$ and link function $g(\cdot)$. 
 We now illuminate specific examples of hLSM for mixture multi-layer networks, hypergraph networks and dynamic networks. 
 
\subsection{Mixture Multi-layer Latent Space Model}\label{sec:MMLSM}
A multi-layer network often consists of multiple networks on the {\it same} set of vertices. 
Denote by $\bcalG=(\calV,\cup_{l=1}^L \calE_l)$ a multi-layer network that
is composed of $L$ layers on the set of vertices $\calV$ of size $|\calV|=n$. The $l$-th layer of network, denoted by $\calG_l=(\calV,\calE_l)$, is an {\it undirected} binary graph. This is a special third-order network with $\calV_1=\calV_2=\calV$ and $\calV_3=[L]$, and $n_1=n_2=n, n_3=L$. 
Then, its adjacency tensor $\bA\in\{0,1\}^{n\times n\times L}$ with its $l$-th slice $[\bA]_{:,:,l}$ being the adjacency matrix of the $l$-th layer.

In \cite{zhang2020flexible}, the authors introduced a multi-layer LSM assuming the {\it unchanged} latent positions of vertices across {\it all} layers. However, in practice, similarities between vertices can shift drastically on different layers. For instance, when trading industrial commodities with other countries, China and USA are quite similar; whereas these two countries are in completely different positions when trading natural products with other countries. This suggests that a more reasonable model should allow heterogeneous latent positions across different layers. Towards that end, 
we propose a novel generative model, called {\it mixture multi-layer latent space model} (MMLSM). It can be regarded as a generalization of the mixture multi-layer SBM \citep{jing2021community}. 

Suppose that there exists a mixture of $m$ LSMs and each layer $\mathcal{G}_l$ is independently sampled from one of these LSM's. Now each layer has a latent label indicating which class of LSM it is sampled from. 
More specifically, for each $j\in[m]$, the $j$-th class LSM is described by the latent positions $U_j\in\RR^{n\times q_j}$ with $n^{-1}U^{\top}_jU_j$ being identity and by a $q_j\times q_j$ interaction matrix $C_j$. Given a link function $g(\cdot)$, if $\calG_l$ is sampled from the $j$-th class LSM, its expected adjacency matrix is simply $g(U_jC_jU_j^{\top})$. For simplicity, we denote
\begin{itemize}
\item LSM($U_j, C_j, g(\cdot)$) --- the $j$-th class LSM with parameter $U_j$, $C_j$ and link function $g(\cdot)$. 
\item $s_l\in[m]$  --- the latent label of $l$-th layer for any $l\in [L]$. Denote $\SS=\{s_1,\cdots, s_L\}$.
\item $L_j=\#\{l: s_l=j, l\in[L]\}$ --- the number of layers generated by the $j$-th class LSM. 
\end{itemize}
Throughout this paper, we regard the layer labels $\SS$ as being fixed. Consequently, the observed adjacency tensor obeys
$$
[\bA]_{i_1, i_2, l} \stackrel{{\rm ind.}}{\sim} \textrm{Bernoulli}\big(g([U_{s_l}C_{s_l}U_{s_l}^{\top}]_{i_1,i_2})\big),\quad \forall (i_1, i_2, l)\in [n]\times [n]\times [L].
$$
We call $U_j$ the {\it local} latent positions of the $j$-th class LSM. Vertices $i_1$ and $i_2$ are {\it locally} similar in the $j$-th class LSM if the $i_1$-th and $i_2$-th rows of $U_j$ are close.   Let $\bar U=(U_1,\cdots,U_m)\in \RR^{n\times \bar{q}}$ be the collection of all {\it local} latent positions where $\bar{q}=\sum_{j=1}^m q_j$. The closeness between the $i_1$-th and $i_2$-th row of $\bar{U}$ implies the {\it global} similarities of vertices $i_1$ and $i_2$ across all layers.  MMSLM can be written in the form of hLSM. 
Define the $\bar{q}\times \bar{q}\times m$ interaction tensor $\bC$ such that its $j$-th slice $[\bC]_{:,:,j}$ equals ${\rm diag}(0_{q_1},\cdots, 0_{q_{j-1}}, C_j, 0_{q_{j+1}},\cdots, 0_{q_m})$, where $0_q$ denotes the $q\times q$ all-zero matrix.
Denote the $L\times m$ layer-label matrix $W=(e_{s_1},\cdots, e_{s_L})^{\top}$ with $e_j$ being the $j$-th canonical basis vector in $\RR^{m}$. Thus we can write 
$\bTheta^*=\bC \cdot \llbracket {\bar U} ,{\bar U}, W \rrbracket$  and $\EE\bA=g(\bTheta^{\ast})$.

Let $W^{\ast}:=L^{1/2}\cdot W\text{diag}(L_1^{-1/2},\cdots,L_m^{-1/2})$ be the layer latent position matrix such that $L^{-1}\cdot W^{\ast\top}W^{\ast}=I_m$. The latent position $W^{\ast}$ reflects how layer label, as an independent ``actor", affects vertex interactions. But $\bar{U}$ may be rank deficient and thus inappropriate to be treated as {\it global} latent positions. Denote $r=\textrm{rank}(\bar U)$ and $n^{-1/2}\bar {U}^{\ast}$ the top-$r$ left singular vectors of $\bar U$ so that $n^{-1}\bar U^{\ast\top}\bar U^{\ast}$ is the identity matrix. We refer to $\bar U^{\ast}$ as the {\it global} latent positions of vertices. Therefore, $\bTheta^{\ast}$ can be re-parameterized and written as $\bTheta^{\ast}=\bC^{\ast}\cdot \llbracket \bar U^{\ast}, \bar U^{\ast}, W^{\ast}\rrbracket$ where the new interaction tensor $\bC^{\ast}$ is of size $r\times r\times m$. Clearly, $nL^{1/2}\bC^{\ast}$ is attainable by multiplying $\bC$ with singular values and right singular vectors of $\bar U$ in the 1-st and 2-nd modes, and with $\text{diag}(L_1^{1/2},\cdots,L_m^{1/2})$ in the 3-rd mode, accordingly. Finally, we write
\begin{equation}\label{decom:multilayer}
[\bA]_{i_1,i_2,l}\stackrel{{\rm ind.}}{\sim} \textrm{Bernoulli}\big([g(\bC^{\ast}\cdot \llbracket \bar U^{\ast}, \bar U^{\ast}, W^{\ast}\rrbracket)]_{i_1,i_2,l}\big),\quad 1\leq i_1\leq i_2\leq n, l\in[L]
\end{equation}
implying that the MMLSM is an hLSM with parameters $\bC^{\ast}, \bar U^{\ast}, W^{\ast}$ and the link function $g(\cdot)$. In MMLSM, we aim to estimate the local latent positions $U_j$'s, layer latent positions $W^{\ast}$ and global latent positions $\bar U^{\ast}$.

	We remark that, although we focus on undirected networks, there is no substantial difficulty to generalize our framework to directed cases, in which the entries of parameter tensor can be written in the form $\bTheta^*=\bC^* \cdot \llbracket \bar U^*, \bar V^*, W^*\rrbracket$.

\subsection{Hypergraph Latent Space Model}\label{sec:hLSM} 
A hypergraph network models higher-order interactions, called hyperedges, among a set of vertices. Without loss of generality, we focus on {\it $3$-uniform} hypergraph where each hyperedge connects exactly $3$ vertices. 
We now propose the hypergraph latent space model (hyper-LSM). 
Let $\mathcal{G}=(\calV,\calE)$ be a 3-uniform undirected binary hypergraph with $\calV=[n]$ being the set of vertices and $\calE$ being the set of hyperedgs, i.e., $(i_1, i_2, i_3)\in \calE$ if there exists a hyperedge among vertices $i_1$, $i_2$ and $i_3$. 

In hyper-LSM, each vertex $i\in \calV$ is associated with an {\it unknown} latent position vector $u_i^{\ast}\in \RR^{r}$. Similarly, the probability of generating hyperedge $(i_1,i_2,i_3)$ only depends solely on the latent positions. Suppose $U^{\ast}=(u_1^{\ast},\cdots, u_n^{\ast})^{\top}$ satisfying $n^{-1}U^{\ast\top}U^{\ast}=I_r$ for identifiability. Let $\bA\in\{0,1\}^{n\times n\times n}$ be the adjacency tensor of $\calG$. We assume there exists an {\it unknown} interaction $r\times r\times r$ tensor $\bC^{\ast}$ and a {\it known} link function such that
$$
[\bA]_{i_1,i_2,i_3} \stackrel{{\rm ind.}}{\sim} \textrm{Bernoulli}\big(\big[g(\bC^{\ast}\cdot \llbracket U^{\ast}, U^{\ast}, U^{\ast} \rrbracket)\big]_{i_1,i_2,i_3}\big), \quad \forall 1\leq i_1\leq i_2\leq i_3\leq n
$$
implying that $\EE\bA=g(\bTheta^{\ast})$ where $\bTheta^{\ast}=\bC^{\ast}\cdot \llbracket U^{\ast}, U^{\ast}, U^{\ast}\rrbracket$. Therefore, the hyper-LSM is an hLSM with parameters $\bC^{\ast},  U^{\ast}$ and the link function $g(\cdot)$

If $g(x)=x$ and $U^{\ast}\in \{0,1\}^{n\times r}$ is a {\it membership} matrix such that $U^{\ast}1_r=1_n$, the hyper-LSM reduces to the hypergraph stochastic block model \citep{ghoshdastidar2017consistency, kim2018stochastic, chien2018community,yuan2018testing}. Moreover, if $U^{\ast}$ is the product of a diagonal matrix and a membership matrix, the hyper-LSM becomes the degree corrected block model \citep{ke2019community}.

\subsection{Dynamic Latent Space Model}\label{sec:DLSM}
A dynamic network is a times sequence of networks on the same set of vertices. There exist several approaches to model the temporal transition of network structures in dynamic networks \citep{xu2015stochastic,sewell2015latent,sarkar2005dynamic,matias2015statistical}. For simplicity, we only consider a simple dynamic network model which was often studied for change point detection in dynamic networks \citep{bhattacharjee2018change}.

Let a dynamic network $\bcalG=\{\calG_t\}_{t=1}^T$ compose of a sequence of $T$ networks on the same set of $n$ vertices $\calV$, where the binary graph $\calG_t:=(\calV,\calE_t)$ represents the interaction at time $t$. Denote $\bA\in\{0,1\}^{n\times n\times T}$ the adjacency tensor of $\bcalG$ whose $t$-th slice $[\bA]_{:,:,t}$ is the adjacency matrix of $\calG_t$. For simplicity, 
we assume the network structures only change at $m\ll T$ {\it unknown} time points $\{t_j\}_{j=1}^m$, called change points \citep{bhattacharjee2018change,wang2018optimal}. Here, $t_1=1$ and hence the initial network is always identified as a change point. The main task is to identify the other $m-1$ change points and also recover underlying network structures, e.g., the latent positions. For each $j\in[m]$ and $t\in[t_{j},t_{j+1})$, we assume $\mathcal{G}_t$ is generated from the same latent space model with the {\it local} latent positions $U_j^{\ast}\in \RR^{n\times q_j}$  and interaction matrix $C_j^{\ast}\in \mathbb{R}^{q_j\times q_j}$. We assume $n^{-1}U_j^{\ast\top }U_j^{\ast}=I_{q_j}$ for identifiability and the network layer at each time point is independently sampled from the others. Denote $\bar U=(U_1^{\ast},\cdots,U_m^{\ast})\in\mathbb{R}^{n\times \bar{q}}$ with $\bar{q}=\sum_{j=1}^m q_j$, whose rows reflect the {\it global} similarity between vertices throughout all the time. One can similarly define the interaction tensor as MMLSM of Section~\ref{sec:MMLSM}. Consequently, this simple dynamic LSM can be viewed as a special case of MMLSM in that the network layers between two consecutive time change points are sampled from an identical latent space model.

\section{Maximum Likelihood Estimation by Tensor Decomposition}
In hLSM, with the observed adjacency tensor generated by model (\ref{eq:hLSM}), our goal is to estimate the latent positions. 
In view of the low-rank structure of $\bTheta^{\ast}$, a natural solution is the maximum likelihood estimator (MLE) with low-rank constraint.
 Let $\ell_n(\cdot):\mathbb{R}^{n_1\times n_2\times n_3}\mapsto \mathbb{R}$ be the negative log-likelihood for the distribution in hLSM, depending on the choice of a link function $g(\cdot)$. Given the observed {\it binary} adjacency tensor $\bA$ and a choice of latent parameters $\bTheta\in \RR^{n_1\times n_2\times n_3}$, the corresponding negative log-likelihood is1
\begin{align}\label{eq:lnTheta-def}
\ell_n(\bTheta)&=-\sum_{i_j\in [n_j], j\in[3]}\Big([\bA]_{i_1,i_2,i_3}\log g([\bTheta]_{i_1,i_2,i_3})+(1-[\bA]_{i_1,i_2,i_3})\log(1-g([\bTheta]_{i_1,i_2,i_3}]))\Big)
\end{align}
Under mild regularity conditions on $g(\cdot)$, e.g. strictly increasing monotonicity, the loss function $\ell_n(\bTheta)$ is convex in $\bTheta$.  More details can be found in Section~\ref{sec:theory}. 
Based on (\ref{eq:lnTheta-def}), we formulate the rank-constrained as follows
 \begin{align}\label{eq:opt-prob}
\min_{\bTheta}\ell_n(\bTheta) \quad\quad \text{subject to} \quad \text{rank}(\bTheta)\le (r_1,r_2,r_3),
\end{align}
 where ${\rm rank}(\cdot)$ denotes the Tucker ranks of a tensor. 
 While the (unconstrained) objective function in problem (\ref{eq:opt-prob}) is usually convex, the rank-constrained feasible set is non-convex. This rank constraint implies the existence of a low-rank decomposition $\bTheta=\bC\cdot \llbracket U,V,W\rrbracket$ with $U\in\RR^{n_1\times r_1}, V\in\RR^{n_2\times r_2}$ and $W\in\RR^{n_3\times r_3}$. Thus the problem (\ref{eq:opt-prob}) is essentially boiled down to a generalized low-rank tensor decomposition which has been intensively investigated in the literature, e.g. the penalized jointly gradient descent in \cite{han2020optimal}, the Riemannian gradient descent in \cite{cai2021generalized}, the alternating minimization in \cite{wang2020learning} and so on. These prior works all take advantage of the specific forms of decomposition of $\bTheta$. 
 
We propose a local algorithm for solving problem (\ref{eq:opt-prob}) by projected gradient descent on Grassmannian. The Grassmannian $\textsf{Gr}(n,r)$ is the collection of all $r$-dimensional subspaces in $\RR^n$. The Stiefel manifold $\textsf{St}(n,r)=\{U: U^{\top}U=I_r\}$ is the set of orthonormal $r$-frames in $\RR^n$. $\textsf{Gr}(n,r)$ can be obtained by identifying those matrices in $\textsf{St}(n,r)$ whose columns
span the same subspace (a quotient manifold), \citep{edelman1998geometry}. Note that any $U\in \textsf{Gr}(n,r)$ satisfies that $U^{\top}U$ is identity. Thus $\textsf{Gr}(n,r)$ naturally serves as the feasible set for the latent positions in hLSM (\ref{eq:hLSM}) where $n^{-1/2}U^{\ast}\in \textsf{Gr}(n_1,r_1)$. Sometimes $U$ has a bounded incoherence constant so that its row-wise norm is small. To this end, for any $\delta\in(0,1)$, we denote $\textsf{Gr}(n,r,\delta)$ the set of $U\in\textsf{Gr}(n,r)$ such that $\|U\|_{2,\infty}\leq \delta$. 
Equipped with Grassmannians and by taking advantage of the incoherence property, we reformulate the problem (\ref{eq:opt-prob}) as
\begin{gather}\label{eq:opt-prob-ncx}
\min_{\bC, U, V, W} \ell_n\big(\bC\cdot \llbracket U,V,W\rrbracket\big)\\
 \text{subject to} \quad U\in \textsf{Gr}(n_1, r_1,\delta_1), V\in \textsf{Gr}(n_2,r_2,\delta_2), W\in \textsf{Gr}(n_3,r_3,\delta_3),\nonumber
\end{gather}
where $\delta_j\in(0,1)$ are tuning parameters.  
We show in Section~\ref{sec:theory} that, under mild conditions and given fixed $U, V, W$, the objective function of (\ref{eq:opt-prob-ncx}) is convex with respect to $\bC$. 
Since $\bC$ is low-dimensional, optimizing $\bC$ is computationally efficient. 
Thus the major computation challenge lies in the search for optimal $U, V$ and $W$. 

The problem (\ref{eq:opt-prob-ncx}) is still highly non-convex and solvable only locally where the gradient descent algorithm is often favored. Unfortunately, a naive gradient descent algorithm cannot ensure that the iterated estimations still 1) remain on Grassmannian; and 2) comply with the incoherence condition. The first issue can be resolved by considering the geodesic gradient descent on Grassmannian \citep{edelman1998geometry, xia2017polynomial}, but this approach is typically burdensome in computation and greatly complicates theoretical analysis. The second issue is simpler to resolve, for instance, by penalization \citep{xia2017polynomial} or projection \citep{ke2019community, han2020optimal}.

We now propose our approach, based on the projected gradient descent on Grassmannians, for locally optimizing the problem (\ref{eq:opt-prob-ncx}). Our algorithm consists of three steps in every iteration.

\begin{itemize}
\item {\it Step~1}. At $t$-th iteration, given the current estimate $\bTheta^{(t)}=\bC^{(t)}\cdot \llbracket U^{(t)}, V^{(t)}, W^{(t)}\rrbracket$, we calculate the gradients $\nabla _{U}\ell_n(\bTheta^{(t)}), \nabla _{V}\ell_n(\bTheta^{(t)})$ and $\nabla_{W}\ell_n(\bTheta^{(t)})$. With a properly chosen stepsize $\eta>0$, we update the estimate by gradient descent and obtain $\check U^{(t)}$ by the left singular vectors of $U^{(t)}-\eta \nabla _{U}\ell_n(\bTheta^{(t)})$. This is equivalent to projecting $U^{(t)}-\eta \nabla _{U}\ell_n(\bTheta^{(t)})$ onto the Grassmannian and thus $\check U^{(t)}\in \textsf{Gr}(n_1, r_1)$. 
	
\item {\it Step~2}. The updated $\check U^{(t)}$ from {\it Step~1} may have a large incoherence coefficient. To reinstate incoherence, 
we impose a regularization that rescales all row $\ell_2$-norms higher than $\delta$ down to $\delta$. Formally, for any $U\in \textsf{Gr}(n,r)$, define the regularization operator by 
$$
\textsf{Reg}_\delta({U}):=D_U U,\quad \text{where}  \quad D_U=\text{diag}\left(\frac{\min\left\{\delta,\|[U]_{1,:}\|\right\}}{\|[U]_{1,:}\|},\cdots, \frac{\min\left\{\delta,\|[U]_{n,:}\|\right\}}{\|[U]_{n,:}\|}\right)
$$
By definition, the output satisfies $\|\textsf{Reg}_{\delta}(U)\|_{2,\infty}\leq \delta$. 
Then we set $U^{(t+1)}$ to be the left singular vectors of $\textsf{Reg}_{\delta_1}(\check U^{(t)})$, which provably satisfies $U^{(t+1)}\in \textsf{Gr}(n_1,r_1,2\delta_1)$.

\item[*] Update $V^{(t+1)}$ and $W^{(t+1)}$ using the same procedure described in {\it Step 1} and {\it Step 2}. 

\item {\it Step~3}. With the updated $U^{(t+1)}, V^{(t+1)}$ and $W^{(t+1)}$, we find the core tensor $\bC^{(t+1)}$ by solving $\arg\min_{\|\bC\|_{\rm F}\leq \xi} \ell_n(\bC\cdot \llbracket U^{(t+1)}, V^{(t+1)}, W^{(t+1)}\rrbracket)$ where $\xi>0$ is a tuning parameter. Recall that $\bC\in\mathbb{R}^{r_1\times r_2\times r_3}$ is low-dimensional and the objective function is convex (see more details in Section~\ref{sec:theory}) in $\bC$, the update $\bC^{(t+1)}$ can be efficiently found by Newton-Raphson algorithm.
\end{itemize}
 The implementation details of our algorithm are summarized in Algorithm~\ref{algo:gd}. We note that the gradient can be explicitly computed by 
 $$
 \nabla_U\ell_n(\bTheta)=\calM_1\big(\nabla \ell(\bTheta)\big)(W\otimes V)\calM_1^{\top}(\bC),
 $$
 where recall that $\bTheta=\bC\cdot \llbracket U, V, W\rrbracket$. 
 
 \begin{algorithm}
\caption{Projected Gradient Descent on Grassmannians}
\label{algo:gd}
\begin{algorithmic}[20]
\Require Tuning parameters $\delta_1, \delta_2,\delta_3, \xi>0$; learning rate $\eta>0$; maximum iterations $t_{\max}$; initialization ${U}^{(0)}\in \textsf{Gr}(n_1,r_1,\delta_1)$, ${V}^{(0)}\in \textsf{Gr}(n_2,r_2,\delta_2)$, $W^{(0)}\in \textsf{Gr}(n_3,r_3,\delta_3)$; ${\mathbf{C}}^{(0)}\leftarrow \arg \min _{\|\bC\|_{\rm F}\leq \xi}\ell_n(\bC\cdot \llbracket U^{(0)},V^{(0)},W^{(0)}\rrbracket)$
\Ensure $\widehat \bTheta$, $\hat U ,\hat V, \hat W$
\For{$t=1,2,\cdots,t_{\max}$}
\State 1. ${\bTheta}^{(t-1)}\leftarrow{\bC}^{(t-1)}\cdot \llbracket {U}^{(t-1)}, {V}^{(t-1)}, {W}^{(t-1)}\rrbracket$
\State 2. (Gradient descent)
\vspace{-0.2cm}
\begin{align*}
\check U^{(t-1)}& \leftarrow {\rm SVD}\big(U^{(t-1)}-\nabla_U \ell_n(\bTheta^{(t-1)})\big)\\
\check V^{(t-1)}& \leftarrow {\rm SVD}\big(V^{(t-1)}-\nabla_V \ell_n(\bTheta^{(t-1)})\big)\\
\check W^{(t-1)}& \leftarrow {\rm SVD}\big(W^{(t-1)}-\nabla_W \ell_n(\bTheta^{(t-1)})\big)
\end{align*}
\vspace{-0.8cm}
\State 3. (Regularization)
$$
{U}^{(t)}\leftarrow\text{SVD}\big(\textsf{Reg}_{\delta_1}(\check{U}^{(t-1)})\big); {V}^{(t)}\leftarrow\text{SVD}\big(\textsf{Reg}_{\delta_1}(\check{V}^{(t-1)})\big); {W}^{(t)}\leftarrow\text{SVD}\big(\textsf{Reg}_{\delta_1}(\check{W}^{(t-1)})\big)
$$
\State 4. Compute ${\mathbf{C}}^{(t)}\leftarrow \arg \min _{\|\bC\|_{\rm F}\leq \xi} \ell_{n}(\bC\cdot \llbracket U^{(t)}, V^{(t)}, W^{(t)}\rrbracket)$
\EndFor
\State Set $\widehat \bC\leftarrow\bC^{(t)}$,  $\widehat U \leftarrow U^{(t)}, \hat V\leftarrow V^{(t)}, \hat W\leftarrow W^{(t)}$; ${\widehat \bTheta}\leftarrow{\widehat \bC}\cdot \llbracket \hat U, \hat V, \hat W\rrbracket$
\end{algorithmic}
\end{algorithm}

\section{Convergence and Estimation Accuracy}\label{sec:theory}
We now present the convergence performances of Algorithm~\ref{algo:gd} and the general statistical error bounds of the final estimates of latent positions. Then we apply these results to several specific hLSM models and elucidate the accuracy of estimated latent positions.  
\subsection{Regularity conditions on the link and loss functions and properties}
The link function $g(\cdot)$ plays a decisive role in the convergence of Algorithm~\ref{algo:gd}. It determines the geometry of loss function $\ell_n(\cdot)$. 
Define
$$
g_+(x):=\left(\frac{g^{\prime}(x)}{g(x)}\right)^2-\frac{g^{\prime \prime}(x)}{g(x)}\quad {\rm and}\quad g_-(x):=\left(\frac{g^{\prime}(x)}{1-g(x)}\right)^2+\frac{g^{\prime \prime}(x)}{1-g(x)},
$$
which are the second order derivative of $\log g(x)$ and $\log(1-g(x))$. 
\begin{assumption}\label{assump:link}
Assume that for any small $\alpha>0$, there exist $\gamma_{\alpha},\beta_{\alpha}>0$ depending only on $\alpha$ and $g(\cdot)$ such that 
\begin{gather*}
\min\left\{\inf _{|x| \leq \alpha}g_+(x),\inf _{|x| \leq \alpha}g_-(x)\right\}\ge \gamma_\alpha\\
\max\left\{\sup _{|x| \leq \alpha}g_+(x), \sup _{|x| \leq \alpha}g_-(x) \right\} \le \beta_\alpha
\end{gather*}
\end{assumption}
The quantities $\gamma_{\alpha}$ and $\beta_{\alpha}$ are often sensitive to $\alpha$. For instance, if $g(x)=(1+e^{-x/\sigma})^{-1}$ is the logistic link with a global scaling $\sigma>0$, we have $\gamma_{\alpha}=e^{\alpha/\sigma}[\sigma(1+e^{\alpha/\sigma})]^{-2}$ and $\beta_{\alpha}=1/(4\sigma^2)$; if $g(x)=\Phi(x)$ is the probit link, we have $\gamma_{\alpha}\asymp (\alpha+0.1)(2\pi)^{-1/2}\cdot e^{-\alpha^2}$ and $\beta_{\alpha}\geq 0.6$. These examples suggest that $\beta_{\alpha}\gamma_{\alpha}^{-1}$ increases fast as $\alpha$ becomes larger. 

We now state our main assumption on the latent positions $U^{\ast}, V^{\ast}, W^{\ast}$ and the underlying low-rank tensor $\bTheta^{\ast}=\bC\cdot \llbracket U^{\ast}, V^{\ast}, W^{\ast}\rrbracket$ of hLSM (\ref{eq:hLSM}). 
\begin{assumption}\label{assump:Theta}
Assume that $U^{\ast}, V^{\ast}, W^{\ast}$ are incoherent with constants upper bounded by $\mu_0>0$, i.e. $\textsf{Incoh}(U^{\ast}), \textsf{Incoh}(V^{\ast}), \textsf{Incoh}(W^{\ast})\leq \mu_0$. 
	Also, the largest singular value of $\bC^*$ is upper bounded by  ${\oLambda}(\bC^*)\le  \alpha \mu_0^{-3}(r_1r_2r_3)^{-1/2}$. 
\end{assumption}
In hLSM, Assumption~\ref{assump:Theta} implies that $\|U^{\ast}\|_{2,\infty}\leq \mu_0r_1^{1/2}$, $\|V^{\ast}\|_{2,\infty}\leq \mu_0r_2^{1/2}$ and $\|W^{\ast}\|_{2,\infty}\leq \mu_0r_3^{1/2}$. Together with the upper bound of $\oLambda(\bC^{\ast})$, Assumption~\ref{assump:Theta} implies that $\|\bTheta^{\ast}\|_{\infty}\leq \alpha$. Then, Assumption~\ref{assump:link} implies that entry-wisely, $\gamma_{\alpha}\leq g_+(\bTheta^{\ast})\leq \beta_{\alpha}$ and $\gamma_{\alpha}\leq g_-(\bTheta^{\ast})\leq \beta_{\alpha}$. This is crucial to ensure the strongly convexity and smoothness of the loss function around the truth. The following lemma is straightforwardly implied by Assumption~\ref{assump:link}, thus we omit its proof.

\begin{lemma}\label{lem:lnproperty} Under Assumption~\ref{assump:link}, the loss function $\ell_n(\cdot)$ is $\gamma_\alpha$-strongly convex and $\beta_{\alpha}$-smooth on the set $\calK_{\alpha}:=\{\bTheta\in\RR^{n_1\times n_2\times n_3}: \|\bTheta\|_{\infty}\leq \alpha\}$, i.e.,
\begin{align*}
\big<\nabla \ell_n (\bTheta_1) - \nabla \ell_n(\bTheta_2),  \bTheta_1-\bTheta_2\big>\geq& \gamma_{\alpha}\|\bTheta_1-\bTheta_2\|_{\rm F}^2\\
\|\nabla \ell_n (\bTheta_1) - \nabla \ell_n(\bTheta_2)\|_{\rm F}\leq& \beta_{\alpha} \|\bTheta_1-\bTheta_2\|_{\rm F}
\end{align*}
for any $\bTheta_1, \bTheta_2\in \calK_{\alpha}$. 
\end{lemma}
The next lemma investigates the update of $\bC$ in the main iteration of Algorithm \ref{algo:gd} and quantifies the convexity of the objective function in $\bC$, given properly updated $U,V$ and $W$.  We relegate its proof to the appendix (Section \ref{sec::proof::lemma::2}). 
\begin{lemma}\label{lem:solveC}
Suppose Assumption~\ref{assump:link} holds. Let $U\in\textsf{Gr}(n_1,r_1,\delta_1), V\in\textsf{Gr}(n_2,r_2,\delta_2), W\in\textsf{Gr}(n_3,r_3,\delta_3)$ be fixed with $\delta_j\leq \mu_0(r_j/n_j)^{1/2}$. If we view $\ell_n(\bC\cdot \llbracket U,V,W\rrbracket)$ as a function of $\bC$, then it is $\gamma_{\alpha}$-strongly convex on the set $\big\{\bC\in\RR^{r_1\times r_2\times r_3}: \|\bC\|_{\rm F}\leq \alpha \mu_0^3(n_1n_2n_3r_1^{-1}r_2^{-1}r_3^{-1})^{1/2}\big\}$. 
\end{lemma}
By Lemma~\ref{lem:solveC}, with a properly chosen tuning parameter $\xi$, the objective function in the optimization program for updating $\bC$ is strongly convex. 

\subsection{Error bounds of latent position estimates under general hLSM}
Let $\{U^{(t)}\}_{t=1}^{t_{\max}}, \{V^{(t)}\}_{t=1}^{t_{\max}}, \{W^{(t)}\}_{t=1}^{t_{\max}}$ be the iterative updates by Algorithm~\ref{algo:gd}.  Notice that, due to the column-normalization of SVD, $U^{(t)}$ (and similarly, $V^{(t)}$ and $W^{(t)}$) estimates $n_1^{-1/2}U^{\ast}$ rather than $U^{\ast}$, up to an unknown right-rotation. Therefore, we measure the error of $U^{(t)}$ by the chordal Frobenius-norm distance. Formally, define
$$
d_{\textsf{f}}(U^{(t)}, n_1^{-1/2}U^{\ast}):=\min_{O\in\OO_{r}} \| U^{(t)}-n_1^{-1/2}U^{\ast}O\|_{\rm F},
$$
where $\OO_r$ is the set of $r\times r$ orthogonal matrices. Define the error measurements for $V^{(t)}$ and $W^{(t)}$ similarly. Then we define the overall error $\textsf{D}_t$ at the $t$-th iteration to be
$$
\textsf{D}_t^2:=d_{\textsf{f}}^2(U^{(t)}, n_1^{-1/2}U^{\ast})+d_{\textsf{f}}^2(V^{(t)}, n_2^{-1/2}V^{\ast})+d_{\textsf{f}}^2(W^{(t)}, n_3^{-1/2}W^{\ast})
$$
The statistical error of the final estimate $(\hat U, \hat V, \hat W)$ depends on the gradient of the loss function at the truth $\bTheta^{\ast}$. Let $\br=(r_1,r_2,r_3)$ denote the Tucker ranks of $\bTheta^{\ast}$. The stochastic error of the final estimate in hLSM is determined by 
$$
\textsf{Err}_{\br}:=\sup_{\|\bX\|_{\rm F}\le 1, {\rm rank}(\bX)=\br}\langle \nabla\ell_n(\bTheta^*)-\E \nabla\ell_n(\bTheta^{*}),\bX\rangle
$$
where recall that $\ell_n(\bTheta)$ depends on the random $\bA$. 
Under hLSM, we have $\E \nabla \ell_n(\bTheta^{\ast})=0$.  To see this, first recall that
$$
[\nabla \ell_n(\bTheta^{\ast})]_{i_1,i_2,i_3}=\frac{g([\bTheta^{\ast}]_{i_1,i_2,i_3})-[\bA]_{i_1,i_2,i_3}}{g([\bTheta^{\ast}]_{i_1,i_2,i_3})\big(1-g([\bTheta^{\ast}]_{i_1,i_2,i_3})\big)}\cdot g'([\bTheta^{\ast}]_{i_1,i_2,i_3}).
$$
Also recall that $\|\bTheta^{\ast}\|_{\infty}\leq \alpha$ under Assumption~\ref{assump:Theta}. Define $\zeta_{\alpha}:=\sup_{|x|\leq \alpha} |g'(x)|\big(g(x)(1-g(x))\big)^{-1}$. Then, the entries of $\nabla \ell_n(\bTheta^{\ast})$ are independent centered sub-Gaussian random variables which are uniformly upper bounded by $\zeta_{\alpha}$. The following lemma characterizes the magnitude of the stochastic error $\textsf{Err}_{\br}$, whose proof is deferred to the appendix. 
\begin{lemma}\label{lem:xibound}
Under Assumption~\ref{assump:Theta}, there exist an absolute constant $c_0>0$ such that with probability at least $1-\exp\big(-c_0(r_1r_2r_3+n_1r_1+n_2r_2+n_3r_3)\big)$, 
$$
\textsf{Err}_{\br} \le {\zeta_\alpha}\cdot\Big({r_1r_2r_3+\sum\nolimits_{k=1}^3n_kr_k}\Big)^{1/2}
$$
\end{lemma}
Denote $\kappa_0:=\oLambda(\bC^*)/\underline{\Lambda}(\bC^*)$ the condition number of $\bC^*$ and $\bar r=\max_{1\leq j\leq 3} r_j$. 
The following theorem shows that, with good initializations and appropriately chosen tuning parameters, Algorithm~\ref{algo:gd} converges linearly and the error of final outputs only depends on the signal strength $\underline{\Lambda}(\bC^{\ast})$ and the stochastic error $\textsf{Err}_{\br}$. 
\begin{theorem}\label{thm:main}
Suppose Assumption~\ref{assump:link}-\ref{assump:Theta} hold in hLSM (\ref{eq:hLSM}) and $\beta_{\alpha}\leq \gamma_{\alpha}^2/(6\kappa_0^2)$. Assume that 
\begin{enumerate}[label=(\alph*)]
\item Initialization error: $\textsf{D}_0^2\leq c_1\kappa_0^{-8}/\bar{r}$;
\item Signal-to-noise ratio: $(n_1n_2n_3)^{1/2}\cdot \underline{\Lambda}(\bC^{\ast})/ \textsf{Err}_{\br}\geq C_1{\kappa_0^4\bar{r}}/(\sqrt{c_1}\land c_2)$,
\end{enumerate}
where $c_1,c_2,c_3\in(0,1)$ and $C_1>0$ are constants depending only on $\alpha$ and $c_2< c_3$. Let the tuning parameters be $\delta_j=C_j^{\prime}\mu_0(r_j/n_j)^{1/2}$ for $1\leq j\leq 3$ and $\xi=C_4^\prime\alpha \mu_0^3 (n_1n_2n_3r_1^{-1}r_2^{-1}r_3^{-1})^{1/2}$ for some absolute constants $C_1^\prime, C_2^\prime, C_3^\prime, C_4^\prime>0$. If we choose step size $\eta=\eta_0\kappa_0^{-4}\underline{\Lambda}^{-2}(\bC^{\ast})/\bar{r}$ with $\eta_0\in[c_2,c_3]$, we have for all $t=1,\cdots,t_{\max}$ that
\begin{align*}
\textsf{D}_t^2\le\Big(1-\frac{\eta_0\gamma_{\alpha}}{8\kappa_0^6}\Big)^t\cdot \textsf{D}_0^2+\frac{C_2\bar{r}\cdot\textsf{Err}_{\br}^2}{n_1n_2n_3\cdot \underline{\Lambda}^2(\bC^*)},
\end{align*}
where $C_2>0$ depends only on $\alpha$. 
 Then, after at most $t_{\max}=O\big(\log(n_1n_2n_3\underline{\Lambda}(\bC^{\ast})/(\bar{r}\cdot \textsf{Err}_{\br}))\big)$ iterations, we have
$$
\textsf{D}_{t_{\max}}^2 \le \frac{C_3\bar{r}\cdot\textsf{Err}_{\br}^2}{n_1n_2n_3\cdot \underline{\Lambda}^2(\bC^*)}
$$
where $C_3>0$ depends only on $\alpha$.
\end{theorem}

By treating $\eta_0$ and $\gamma_{\alpha}$ as constants, the proof of Theorem \ref{thm:main} implies that the joint error of the latent positions estimates by Algorithm \ref{algo:gd} contracts as $\textsf{D}_{t+1}^2\leq (1-c_0/\kappa_0^{6})\textsf{D}_t^2+\textsf{ statistical error}$, where the contraction rate $1-c_0\kappa_0^{-6}$ is strictly smaller than $1$ with a fixed stepsize. Therefore, Algorithm \ref{algo:gd} converges linearly to a locally optimal solution. The initialization condition is also mild. In the case $\kappa_0, \bar r=O(1)$, our theorem only requires $\textsf{D}_0\leq c_4<1$ for a universal constant $c_4.$

By combining Theorem \ref{thm:main} and Lemma \ref{lem:xibound}, we obtain
$$
\textsf{D}_{t_{\max}}^2\leq \frac{C_3\zeta_{\alpha}\cdot \bar r(r_1r_2r_3+\sum_{k=1}^3 n_kr_k)}{n_1n_2n_3\cdot \underline{\Lambda}^2(\bC^{\ast})}
$$
If $\bar r, \zeta_{\alpha}=O(1)$ and $n_1\geq n_2\geq n_3$, it implies that $\textsf{D}_{t_{\max}}$ converges to zero as long as $n_2n_3\underline{\Lambda}^2(\bC^{\ast})\to \infty$.  Put it differently, the estimation error of the latent positions diminishes very quickly as the network size grows, which also matches our observations in simulation studies. 

\subsection{Error bound of latent position estimates for specific hLSMs}\label{sec:specific-hLSM}
We now apply Theorem~\ref{thm:main} to the specific examples of hLSM.  Here and after, for notational simplicity, we assume the maximum number of iterations $t_\text{max}=O\big(\log(n_1n_2n_3\underline{\Lambda}(\bC^{\ast})/(\bar{r}\cdot \textsf{Err}_{\br}))\big)$ under the general LSM model (\ref{eq:hLSM}). Throughout this section, we assume the initialization condition of Theorem \ref{thm:main} holds. 

\subsubsection{Application 1: Mixture multilayer latent space model (MMLSM)}

Let $\bar U=(n^{-1/2}\bar U^{\ast})\Sigma_{\bar U}R_{\bar U}^{\top}$ denote the thin SVD of $\bar U$, where the $r\times r$ diagonal matrix $\Sigma_{\bar U}$ contains the singular values of $\bar U$. 
To characterize the signal strength of $\bC^\ast$, define $\bar{\bC}=\bC\times _1 R_{\bar U}^{\top} \times _2 R_{\bar U}^{\top}$. Simple algebra shows that $\underline{\Lambda}(\bC^*)\ge n^{-1}L^{-1/2}\sigma_{\min}^2(\bar{U})\underline{\Lambda}(\bar{\bC})\sqrt{\min_{1\le j\le L} L_j}$. Denote $\kappa_{\bar{U}}$ the condition number of $\bar U$. 
The following corollary is an immediate conclusion from combining Theorem \ref{thm:main} and Lemma \ref{lem:xibound}, whose proof is straightforward and thus omitted.

\begin{corollary}[Error bounds of estimating latent positions in MMLSM]\label{col:MMLSM_latent}
Suppose Assumption~\ref{assump:link}-\ref{assump:Theta} hold and $\beta_{\alpha}\leq \gamma_{\alpha}^2/(6\kappa_0^2)$. 
Let $\widehat{U}:=U^{(t_m)}$  be the output of Algorithm \ref{algo:gd}. Denote the signal strength of $\bar\bC$ by $c_{\ast}=\underline{\Lambda}(\bar{\bC})$. 
If the  network cluster sizes are balanced $\min_{1\le j\le m}L_j\asymp L/m$, then there exists an absolute constant $c_0>0$ such that with probability at least $1-\exp\big(-c_0(2nr+Lm+mr^2)\big)$
\begin{align}\label{eq:cor_MMLSM_err}
d_\textsf{f}^2(\widehat{U},n^{-1/2}\bar{U}^*)\le C_3\zeta_\alpha^2\kappa_{\bar{U}}^4\frac{(r\vee m)\left(2nr+Lm+mr^2\right)}{c_*^2n^2L{m}} 
\end{align}
provided that 
$$
n^2L\ge C_3 \zeta_\alpha^2\kappa_{\bar{U}}^4\left(\frac{r}{{m}}\vee 1\right)\left(2nr+Lm+mr^2\right)c_*^{-2}
$$ 
where $C_3$ is a constant depending only on $\alpha$.
\end{corollary}

	
To gain more insight, let us consider a simple setting where $r, m, \zeta_{\alpha}, \kappa_{\bar U}=O(1)$.  The error rate in (\ref{eq:cor_MMLSM_err}) simplifies to $(n+L)/(c_{\ast}^2n^2L)$, and we observe an interesting phase transition: (1). when the number of layers $L$ is small compared to $n$, that is $L=O(n)$, the error rate is dominated by the first term $1/(c_{\ast}^2nL)$. In this phase, increasing the number of nodes or the number of layers can both improve the estimation of latent positions.  (2). when $L\gg n$, 
the error rate would be bottlenecked by the second term $1/(c_{\ast}^2n^2)$, which does not depend on $L$ anymore.  Consequently, increasing the number of layers can no longer improve the estimate of latent positions. 
This phase transition is also empirically confirmed by our simulation studies, see Section \ref{Sec:simulation}. 
The latter phase seems unexpected since it implies that, beyond certain threshold, increasing the number of layers brings diminishing benefits to the estimation of latent positions. This result, actually, is an outcome due to both the difficulty of the mixture model and the limitation of tensor methods. The mixture nature of MMLSM underlines the importance of estimating the $L\times m$ matrix $W^{\ast}$. However, our tensor method jointly estimates $\bar{U}^{\ast}$ and $W^{\ast}$, and the errors of $\hat U$ and $\hat W$ are thus intertwined. Clearly, when $L \gg n$, estimating $W^{\ast}$ is more difficult than estimating $\bar{U}^{\ast}$.  Therefore, in the latter phase, the error rate reflects the difficulty of recovering $W^{\ast}$ rather than estimating $\bar U^{\ast}$. { This phenomenon can be easily understood from Theorem \ref{thm:main} under general hLSM's.  Indeed, one can expect that for a more general tensor $\bA\in\mathbb{R}^{n_1\times n_2\times n_3=L}$, this error bound would become $(n_1+n_2+n_3)/(n_1n_2n_3)$.  Without loss of generality if $n_1\gg n_2,n_3$, then the dominating term would be $1/(n_2n_3)$ and increasing $n_1$ would only bring diminishing benefits.}

We can also recover the network classes $\mathbb{S}$ by applying standard K-means clustering to the rows of $\widehat W:=\widehat W^{(t_m)}$ from Algorithm~\ref{algo:gd}. Given an $\widehat{\mathbb{S}}=\{\hat s_l\}_{l=1}^L$, the estimator of $\mathbb{S}$, we use the average Hamming distance to measure its accuracy:
$$\mathcal{L}(\widehat{\mathbb{S}}, \mathbb{S})=\underset{\tau: \text { a permutation on }[{m}]}{\min } \frac{1}{L}\sum_{l=1}^L\mathbbm{1}\left(s_l\ne \tau(\hat s_l)\right)$$
\begin{theorem}[Error bounds of network clustering in MMLSM]\label{MMLSM:network}
Under the conditions of Corollary \ref{col:MMLSM_latent}, there exists a global constant $c_0>0$ such that with probability at least $1-\exp\big(-c_0(2nr+Lm+mr^2)\big)$,
$$
\mathcal{L}(\widehat{\mathbb{S}}, \mathbb{S})\le C_3\zeta_\alpha^2\kappa_{\bar{U}}^4\frac{(r\vee m)\left(2nr+Lm+mr^2\right)}{c_*^2n^2L{m}^{2}}
 $$
provided that 
$$
n^2L\ge C_3\zeta_\alpha^2\kappa_{\bar{U}}^4\left(\frac{r}{{m}}\vee 1\right)\left(n\cdot \frac{2r}{{m}}+L+r^2\right)c_*^{-2}
$$ 
where $C_3$ is a constant depending only on $\alpha$.
\end{theorem}

Corollary~\ref{col:MMLSM_latent} and Theorem~\ref{MMLSM:network} suggest that, under similar mild signal strength conditions, both the global latent positions and the layer labels can be consistently recovered.
{
Here we have the similar understanding as in Corollary \ref{col:hLSM_latent} that the accuracy is bottlenecked by the asymptotically smaller  one between $n$ and $L$.
}

\begin{remark}
After obtaining the layer labels, one can further estimate the local latent positions for each ${\rm LSM}(U_j, C_j)$. Since the layers with equal labels are assumed to be sampled from the same {\rm LSM}, it is unnecessary to apply tensor methods (the factor corresponding to the third dimension becomes trivially constant). Interested readers may refer to \cite{zhang2017finding, zhang2020flexible} and references therein for more details. 
\end{remark}

\subsubsection{Application 2: Hypergraph latent space model (hyper-LSM)}

Similar to Corollary \ref{col:MMLSM_latent}, we have the following result. 
\begin{corollary}[Error bounds of estimating latent position in hyper-LSM]\label{col:hLSM_latent}
Suppose Assumption~\ref{assump:link}-\ref{assump:Theta} hold and $\beta_{\alpha}\leq \gamma_{\alpha}^2/(6\kappa_0^2)$. 
Let $\widehat{U}:=U^{(t_m)}$  be the output of Algorithm \ref{algo:gd}. Denote the signal strength of $\bC^{\ast}$ by $c_{\ast}=\underline{\Lambda}(\bC^{\ast})$. 
Then there exists some absolute constant $c_0>0$ such that with probability at least $1-\exp\big(-c_0(r^3+3nr)\big)$,
 \begin{align}\label{eq:cor_hLSM_err}
 d_\textsf{f}(\widehat U,n^{-1/2}U^\ast)\le C_3\zeta_\alpha^2\cdot \frac{3nr+r^3}{n^3c_*^2}
 \end{align}
provided that
$$
n^2 \ge C_3 r\left(3+\frac{r^2}{n}\right)c_*^{-2}
$$ 
with the constant $C_3>0$ depending only on $\alpha$.
\end{corollary}
If $\zeta_{\alpha}, r=O(1)$, the error rate (\ref{eq:cor_hLSM_err}) simplifies to $1/(n^2c_{\ast}^2)$, where we recall that in an hyper-LSM, by definition $L=n$.  
This bound diminishes quadratically in $n$.
Similarly, the minimal signal strength requirement $c_{\ast}$ also decreases {linearly} with respect to $n$. 

\subsubsection{Application 3: Dynamic latent space model (dynamic LSM)}\label{sec:thm-dynamic-LSM}
Lastly, we consider the change point detection in dynamic latent space model. With the output $\widehat W:= W^{(t_\text{max})}$ of Algorithm \ref{algo:gd}, we perform a row-wise screening to  identify the change points $\{t_m\}_{m=1}^M$. More specifically, we iteratively compare the difference of two consecutive rows of $\widehat W$ in $\ell_2$ norm, and for all $t\in[T]$, $t+1$ is identified as a change point if and only if   
$$
\big\|[\widehat W]_{t,:}-[\widehat W]_{t+1,:}\big\|_2\ge \epsilon
$$
for some tuning parameter $\epsilon>0$. Define the $r\times r\times m$ tensor $\bar{\bC}$ in the same fashion as in MMLSM, and we can have the following result.
\begin{theorem}[Exact detection of change points in dynamic LSM]\label{DLSM:latent}
Suppose Assumption~\ref{assump:link}-\ref{assump:Theta} hold and $\beta_{\alpha}\leq \gamma_{\alpha}^2/(6\kappa_0^2)$. 
Denote the signal strength of $\bar\bC$ by $c_{\ast}=\underline{\Lambda}(\bar{\bC})$. 
If the time intervals between neighboring change points are balanced  $\min_{1\le j\le m}T_j\asymp T/m$, then there exist absolute constants $c_0, c_1,c_2> 0$ such that by choosing $\epsilon\in\big[c_1(T/m)^{1/2},c_2(T/m)^{1/2}\big]$, all change points $\{t_m\}_{m=1}^M$ can  be exactly detected with probability at least $1-\exp\big(-c_0(2nr+Tm+mr^2)\big)$, provided that
$$
nT\ge C\kappa_{\bar U}^2(r\vee m)^{1/2}\left(2nr+Tm+{mr^2}\right)^{1/2}\cdot c_*^{-1}
$$
where $C>0$ is constant depending only on $\alpha$.
\end{theorem}
If $\kappa_{\bar U}, r=O(1)$ and $m=O(n)$, by Theorem~\ref{DLSM:latent}, in order to exactly detect those change points, the minimal signal strength requirement becomes $c_{\ast}^2n T^2\geq Cm$ for some absolute constant $C>0$.  

{
Our result characterizes the probability of exact change point recovery, which is a natural consequence of accurate latent position estimation, and is different from the noisy recovery error measurement in \cite{wang2018optimal}.  Therefore, the signal strength assumption of our Theorem \ref{DLSM:latent} and the counterpart of \cite{wang2018optimal} are not directly comparable.  In fact, our result provide richer information about changes in network evolution that are not limited to \emph{sudden changes}.  For instance, our method is capable of revealing a dynamic network that shows rapid but continuous changes during \emph{change periods} rather than \emph{change points}.  This pattern is not covered by most change detection literature in network analysis.
}

\section{Simulations on Synthetic Higher-order Networks}\label{Sec:simulation}


In this section, we showcase the performances of Algorithm \ref{algo:gd} on synthetic higher-order networks. 
We first focus on the general higher-order LSM.  Then we generate synthetic data from the three application scenarios, namely multi-layer, hypergraph and dynamic networks, discussed in Section \ref{Sec:model}, and evaluate the numerical performances.

\subsection{Simulation 1: general higher-order LSM's}

Without loss of generality, we only consider third-order networks for the general higher-order latent space model (\ref{eq:hLSM}). The network sizes are fixed at $n_k\equiv n=50$ and the dimension of latent space is fixed at $r_k\equiv r= 3$ for $k=1,2,3$. We generate the low-rank parameter tensor $\bTheta^{\ast}$ as follows. 
We first generate a truncated standard normal tensor $\widetilde \bTheta\in \mathbb{R}^{n\times n\times n}$  with $[\widetilde \bTheta]_{ijk}\overset{{\rm i.i.d.}}{\sim}\text{TruncNorm}(0,1;[-3,3]),i,j,k\in[n]$,  and then apply higher-order SVD to $10\cdot\widetilde \bTheta$ with multilinear ranks $(r,r,r)$, which produces the core tensor $(n_1n_2n_3)^{1/2}\bC^*\in \mathbb{R}^{r\times r\times r}$ and factor matrices  $n_1^{-1/2}U^*,n_2^{-1/2}V^*,n_3^{-1/2}W^*$. The parameter tensor is then set to be $\bTheta^*=\bC^*\cdot \llbracket U^*,V^*,W^*\rrbracket$. The observed data tensor $\bA$ has independent entries sampled from $\text{Bernoulli}(g(\bTheta^*/\sigma))$ {\it entry-wisely}, where we set the link function $g(\cdot):={\rm logit}(\cdot;\sigma)$ with a global scaling parameter $\sigma\in\{0.1,0.5,1\}$.

\begin{figure}
	\centering
	\begin{subfigure}[b]{.49\linewidth}
		\includegraphics[width=\linewidth]{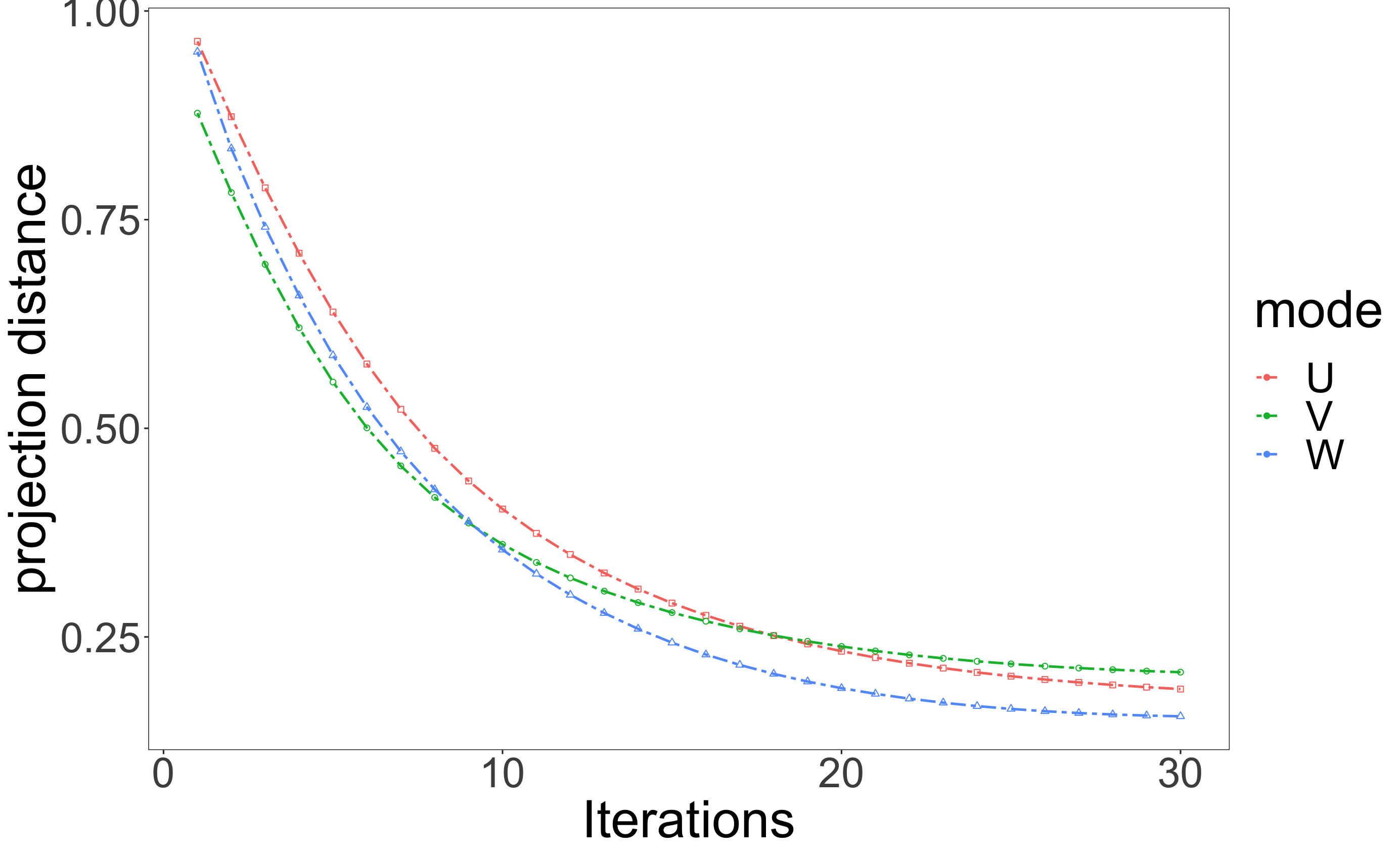}
		\caption{\tiny{Algorithm \ref{algo:gd}}}
	\end{subfigure}
	\begin{subfigure}[b]{.49\linewidth}
		\includegraphics[width=\linewidth]{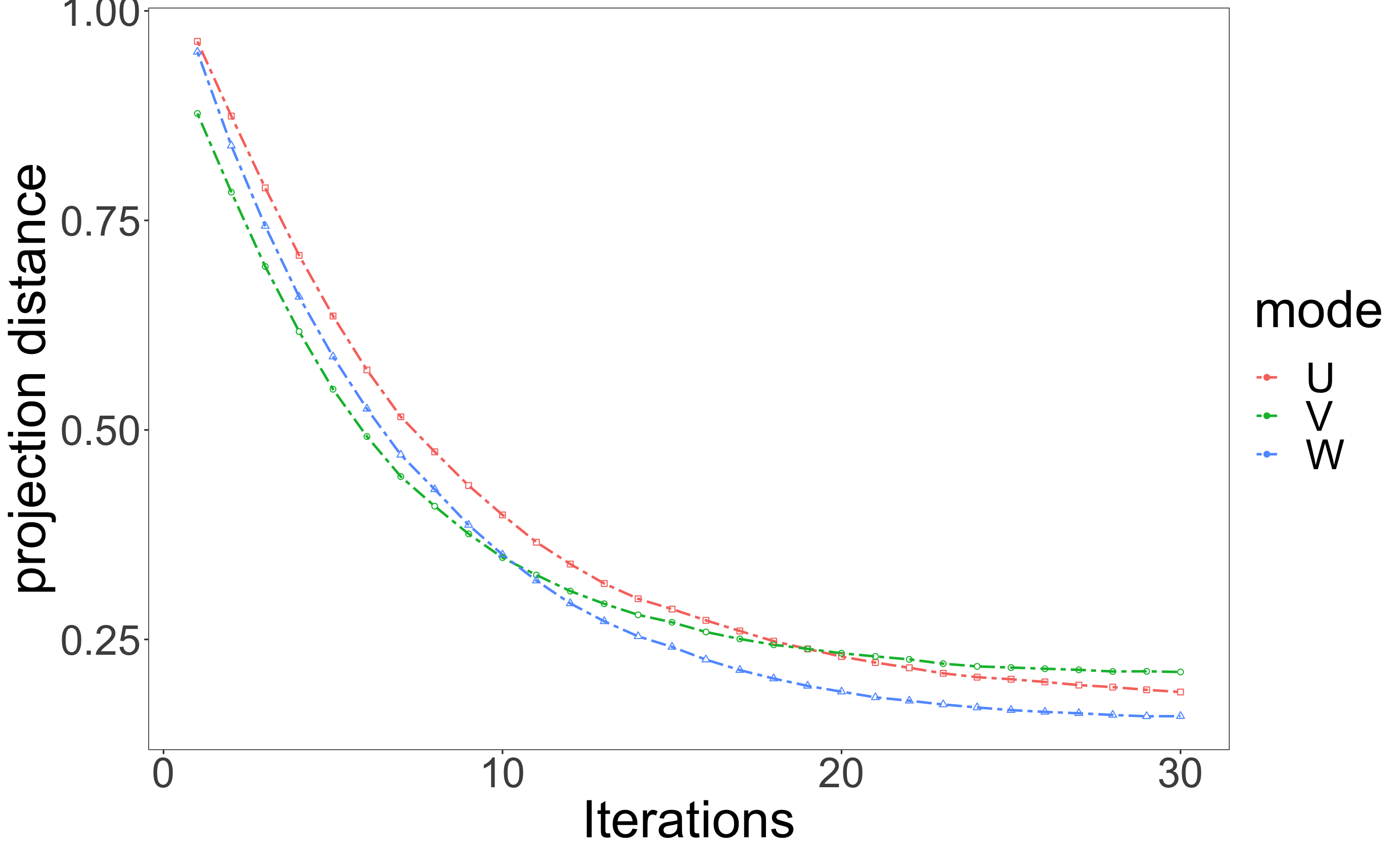}
		\caption{\tiny{Accelerated version of Algorithm \ref{algo:gd}}}
	\end{subfigure}
	\caption{Simulation 1-1 for general hLSM: the convergence of projection error $\fro{U^{(t)}U^{(t)\top}-n_1^{-1}U^*U^{*\top}}^2$ (also for $V,W$ \textit{resp.}).}
	\label{fig:general_projdist}
\end{figure}

\begin{figure}
	\centering
	\begin{subfigure}[b]{.49\linewidth}
		\includegraphics[width=\linewidth]{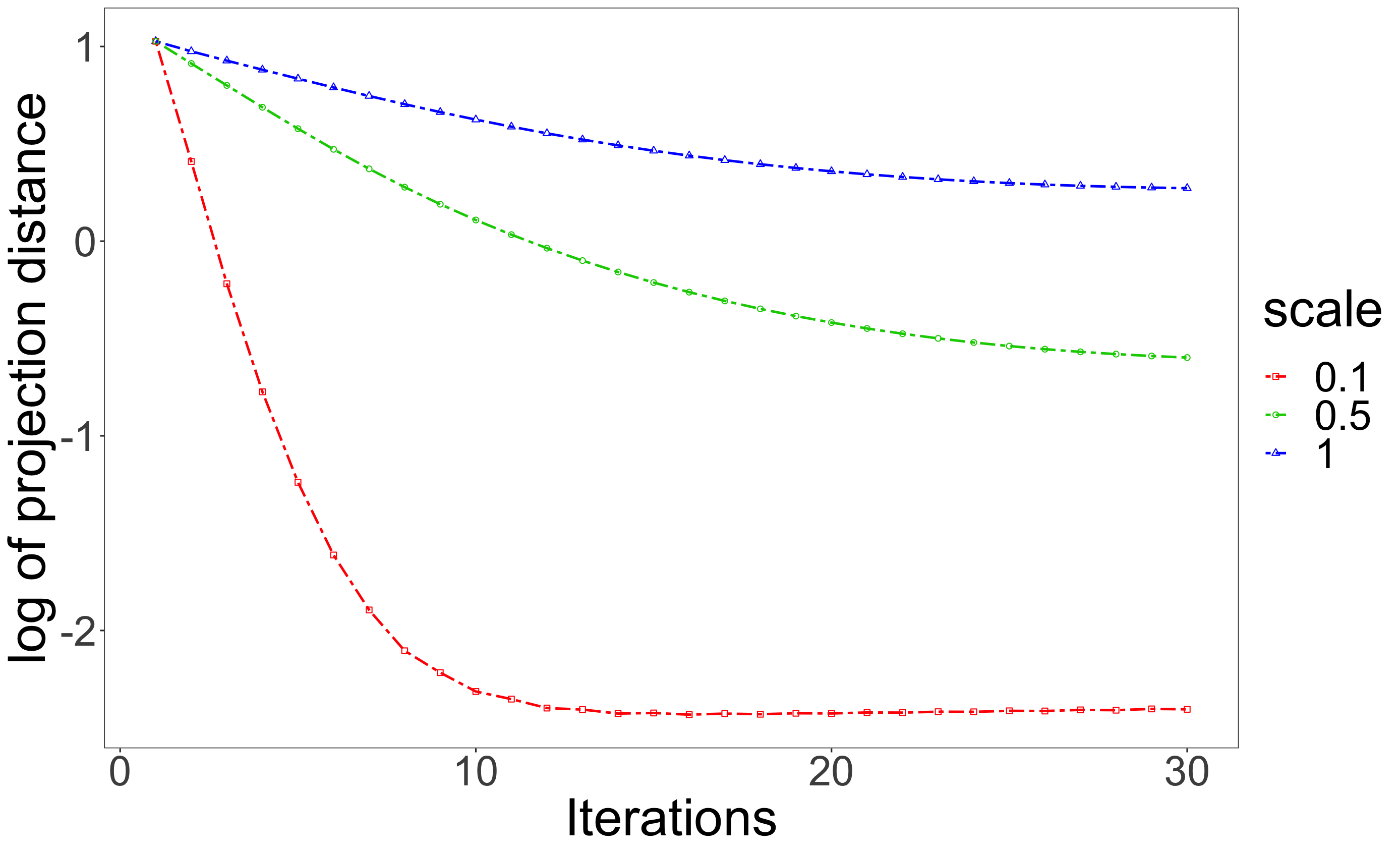}
		\caption{\tiny{Algorithm \ref{algo:gd}}}
	\end{subfigure}
	\begin{subfigure}[b]{.49\linewidth}
		\includegraphics[width=\linewidth]{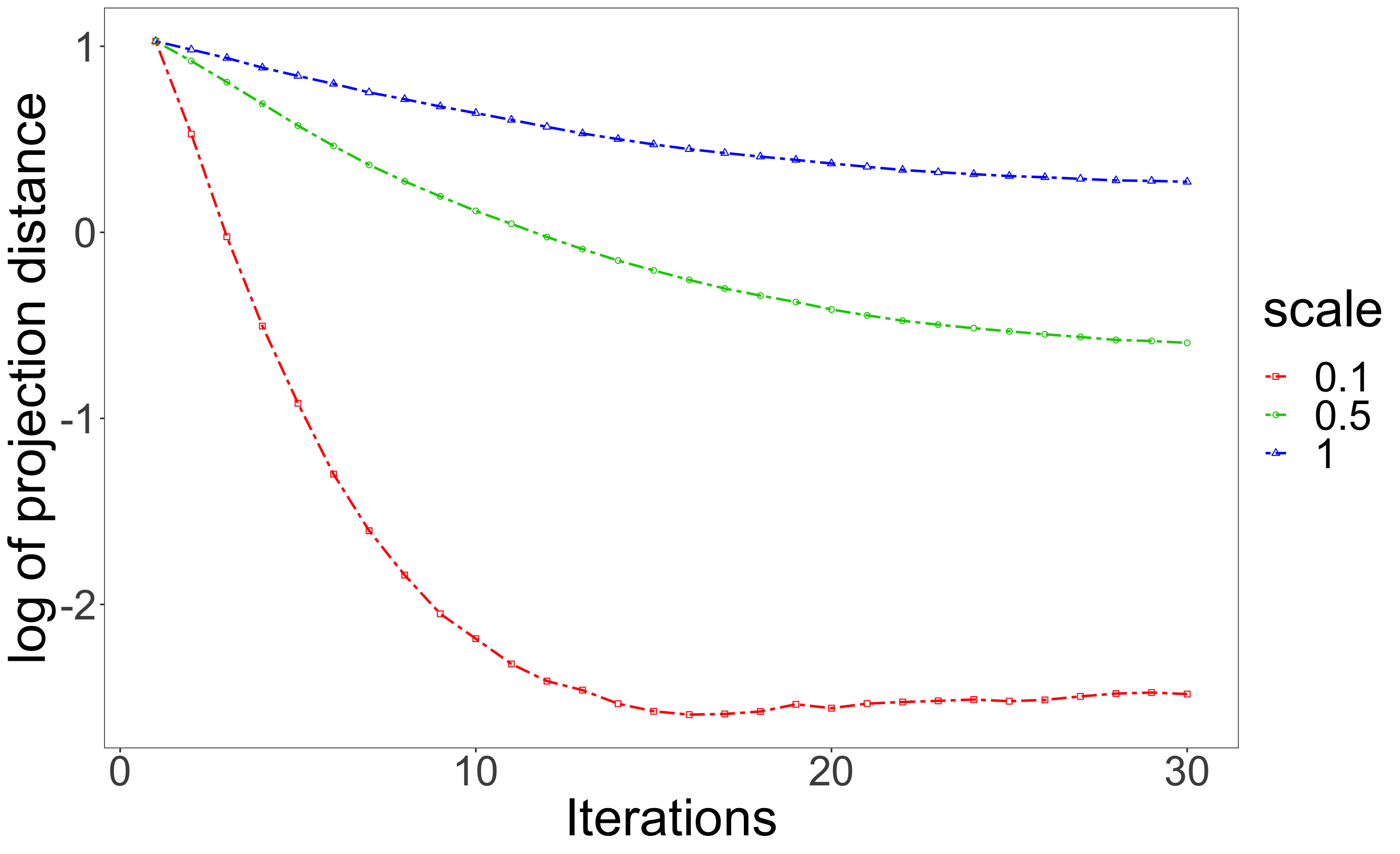}
		\caption{\tiny{Accelerated version of Algorithm \ref{algo:gd}}}
	\end{subfigure}
	\caption{Simulation 1-2 for general hLSM: the convergence of the logarithm of sum of squares of projection error for  $U^{(t)},V^{(t)},W^{(t)}$ under different scales $\sigma\in\{1,0.5,0.1\}$}
	\label{fig:general_logUerror}
\end{figure}

The computation of the projected gradient descent updates $U^{(t)}, V^{(t)}, W^{(t)}$ is fast and memory-efficient.  The main computation burden of Algorithm \ref{algo:gd} comes from the update of the core tensor $\bC^{(t)}$.  
Computing $\bC^{(t)}$ can be recast as essentially estimating a generalized linear model, e.g., logistic/probit regression with logit/probit link function. This step can be computationally demanding when $n_1n_2n_3$ 
is large. Fortunately, the number of parameters we desire to estimate is only $r_1r_2r_3$, comparatively much smaller than $n_1n_2n_3$. To alleviate the computation costs of this step and accelerate our algorithm, 
{ we accelerate by updating $\bC^{(t)}$ using a small sub-sample input: $([A]_{{\cal S}_1, {\cal S}_2, {\cal S}_3}; [U^{(t)}]_{{\cal S}_1,:}, [V^{(t)}]_{{\cal S}_2,:}, [W^{(t)}]_{{\cal S}_3,:}$, where ${\cal S}_k\subset [n_k]$, $|{\cal S}_k|\ll n_k$, instead of the original $(A; U^{(t)}, V^{(t)}, W^{(t)})$,} except the last few iterations. 
{ This random sampling procedure allows to solve $\bC^{(t)}$ via a much smaller scale logistic regression. We regard this method as an accelerated version of Algorithm \ref{algo:gd}. 
This accelerated Algorithm \ref{algo:gd} can greatly improve speed at little cost of estimation accuracy -- Figure \ref{fig:general_projdist} and Figure \ref{fig:general_logUerror} demonstrate that it enjoys almost same convergence and accuracy as the original algorithm. In these simulations, the sampling proportion is $0.1$ and the algorithm runs 5 times faster than the original algorithm.  

}


{
Figure \ref{fig:general_projdist} and Figure \ref{fig:general_logUerror} present the simulation results.
Both figures report that our algorithm converges in around $20$ iterations in terms of the error of latent positions estimates. 
In Figure \ref{fig:general_logUerror}, a smaller $\sigma$ corresponds to the easier dense network setting, and our algorithm converges even faster. Moreover, the linear pattern at the early stages echos the linear convergence of Algorithm \ref{algo:gd} predicted by our theory, see Theorem \ref{thm:main}.
}

\subsection{Simulation 2: mixture multi-layer latent space model}\label{sec:sim-MMLSM}
We consider the mixture multi-layer latent space model and fix $r=3, m=3$. The global latent position matrix $\bar U^*\in \mathbb{R}^{n\times r}$ is generated by the $n^{1/2}$ scaling of the left singular vectors of the $n\times r$ random matrix $\widetilde U$ with its entries $[\widetilde U]_{ij}\overset{{\rm i.i.d.}}{\sim} \mathcal{N}(0.5,1)$. For each $l\in [L]$, we generate the latent network class $s_l$ for the $l$-th layer by the uniform multinomial distribution that $\Prob(s_l=j)=m^{-1}, j\in[m]$. For each $j\in[m]$, we generate the interaction matrix by $C_j=E_jE_j^\top$, where $[E_j]_{ik}\overset{{\rm i.i.d.}}{\sim} \text{Uniform}(-1,1)$. The low-rank parameter tensor  $\bTheta^*=\bC^*\cdot \llbracket\bar U^*,\bar U^*,W^*\rrbracket$, where $\bC^*, W^*$ are defined as that in Section \ref{sec:MMLSM}. For each layer $l\in[L]$, set each individual entry of the adjacency tensor by $[\bA]_{ijl}\overset{{\rm ind.}}{\sim}\text{Bernoulli}(g([\bTheta^*]_{ijl}))$ for $1\le i<j\le n$, and $[\bA]_{ijl}=0$ for $i=j$. The lower-triangular entries in each slice of $\bA$ are set by symmetry. 

We run Algorithm \ref{algo:gd} on $\bA$ and obtain $\widehat U$ and $\widehat W$, where  we initialize $\widehat U$ and $\widehat W$ for Algorithm \ref{algo:gd} by higher-order SVD.
We apply K-means clustering to the rows of $\widehat{W}$ and obtain the estimated network classes $\widehat{\mathbb{S}}$ and measure the performance of latent position estimates by  $\fro{\widehat U\widehat U^\top-n^{-1}\bar U^*\bar U^{*\top}}$ and that of network clustering by the normalized Hamming error $L^{-1}\mathcal{L}(\mathbb{S},\widehat{\mathbb{S}})$. 

\begin{figure}[!h]
	\centering
	\begin{subfigure}[b]{.49\linewidth}
		\includegraphics[width=\linewidth]{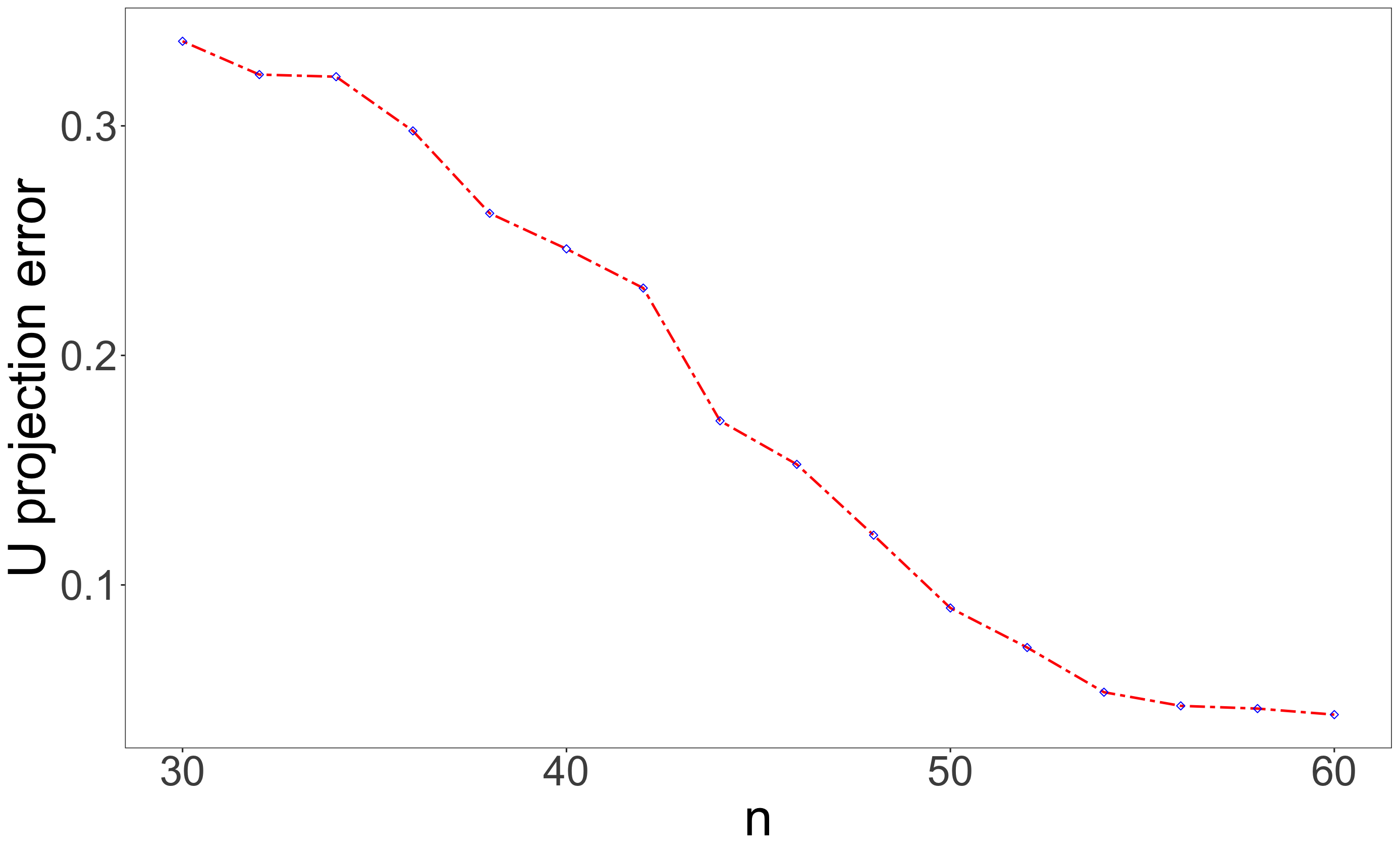}
		\caption{\tiny{Simulation 2-1 for MMLSM: error of latent position estimates with $n$ varying. Here, $L=150, m=3, r=3$.}}
	\end{subfigure}
	\begin{subfigure}[b]{.49\linewidth}
		\includegraphics[width=\linewidth]{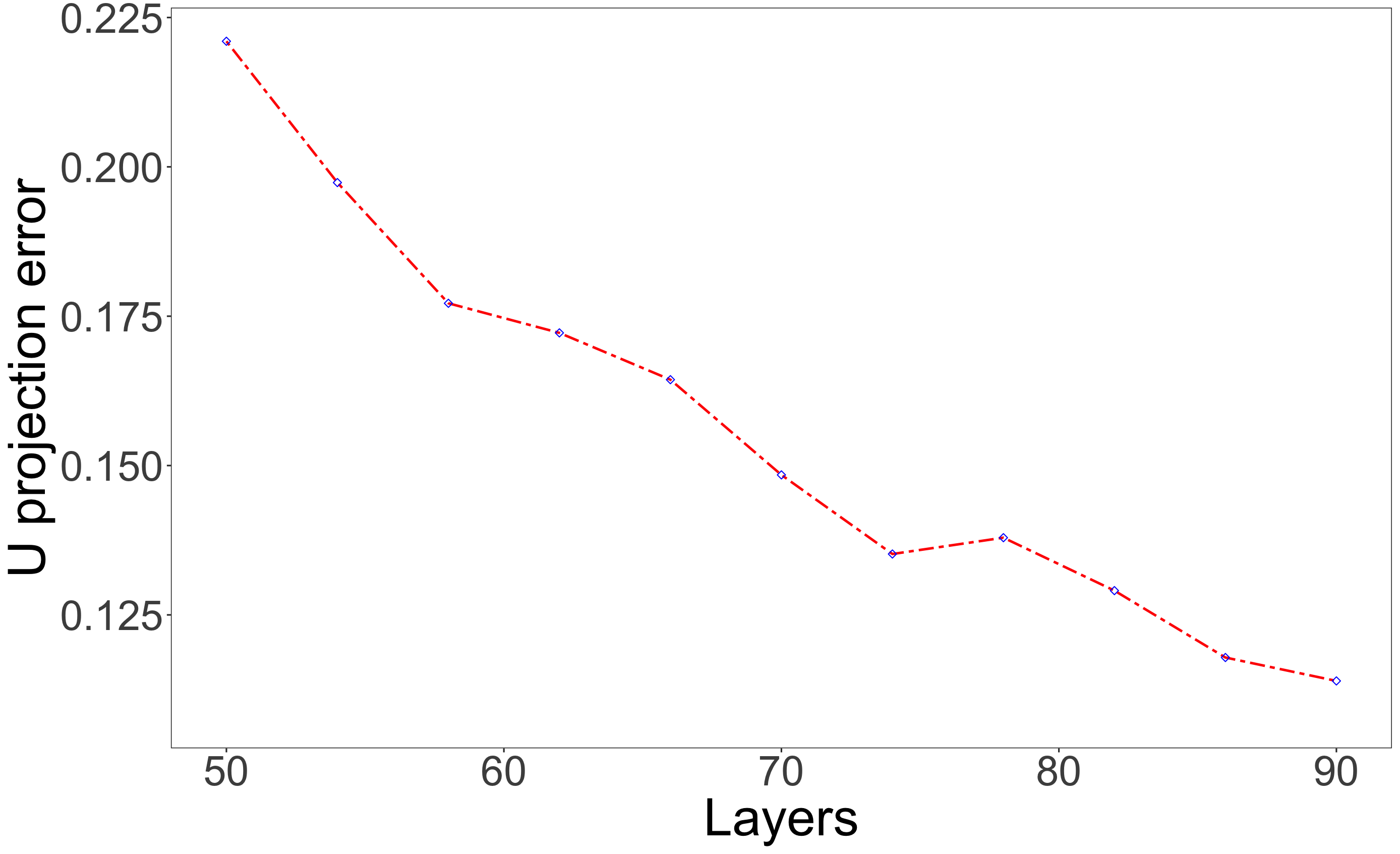}
		\caption{\tiny{Simulation 2-2 for MMLSM: error of latent position estimates with $n$ varying. Here, $L=20, m=3, r=3$.}}
	\end{subfigure}
	\caption{Error of latent position estimates with $n$ varying under two scenarios: $n<L$ and $n>L$}
	\label{fig:multi_Uerror_nvarying}
\end{figure}

\begin{figure}[!h]
	\centering
	\begin{subfigure}[b]{.49\linewidth}
		\includegraphics[width=\linewidth]{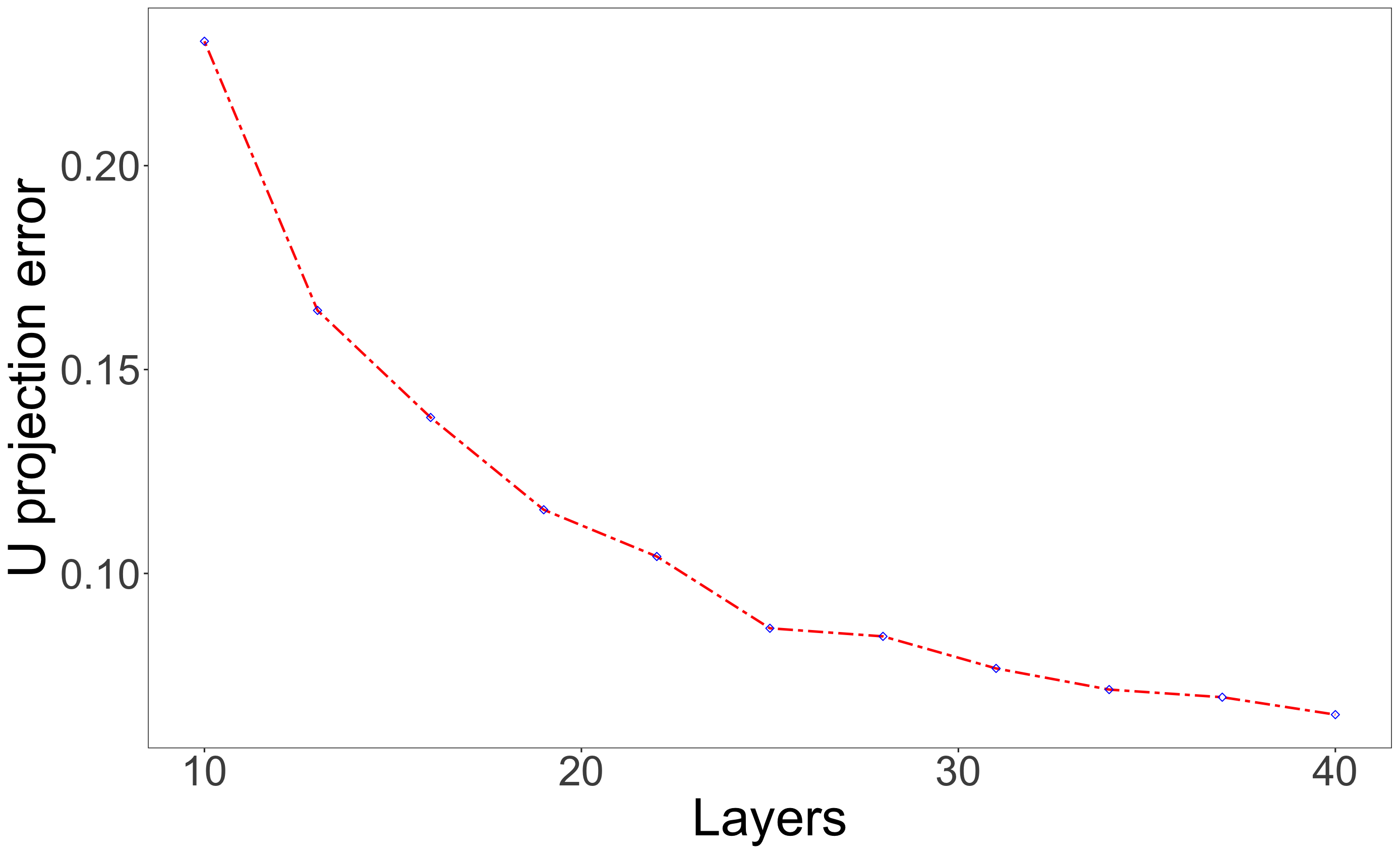}
		\caption{\tiny{Simulation 2-3 for MMLSM: error of latent position estimates with $L$ varying. Here, $n=100, m=3, r=3$.}}
	\end{subfigure}
	\begin{subfigure}[b]{.49\linewidth}
		\includegraphics[width=\linewidth]{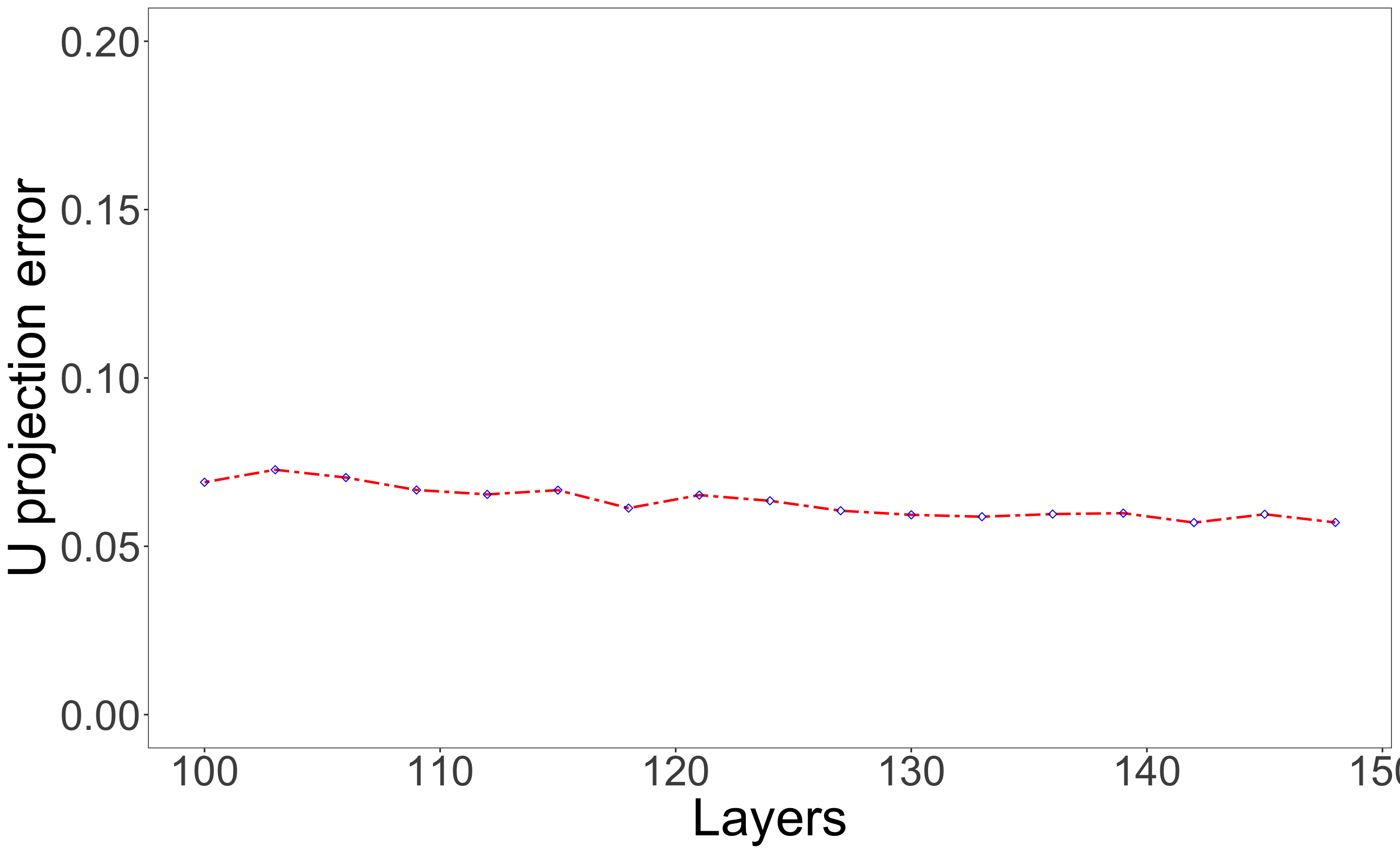}
		\caption{\tiny{Simulation 2-4 for MMLSM: error of latent position estimates with $L$ varying. Here, $n=50, m=3, r=3$.}}
	\end{subfigure}
	\caption{Error of latent position estimates with $L$ varying under two scenarios: $L<n$ and $L>n$}
	\label{fig:multi_Uerror_Lvarying}
\end{figure}

\begin{figure}[!h]
	\centering
	\begin{subfigure}[b]{.49\linewidth}
		\includegraphics[width=\linewidth]{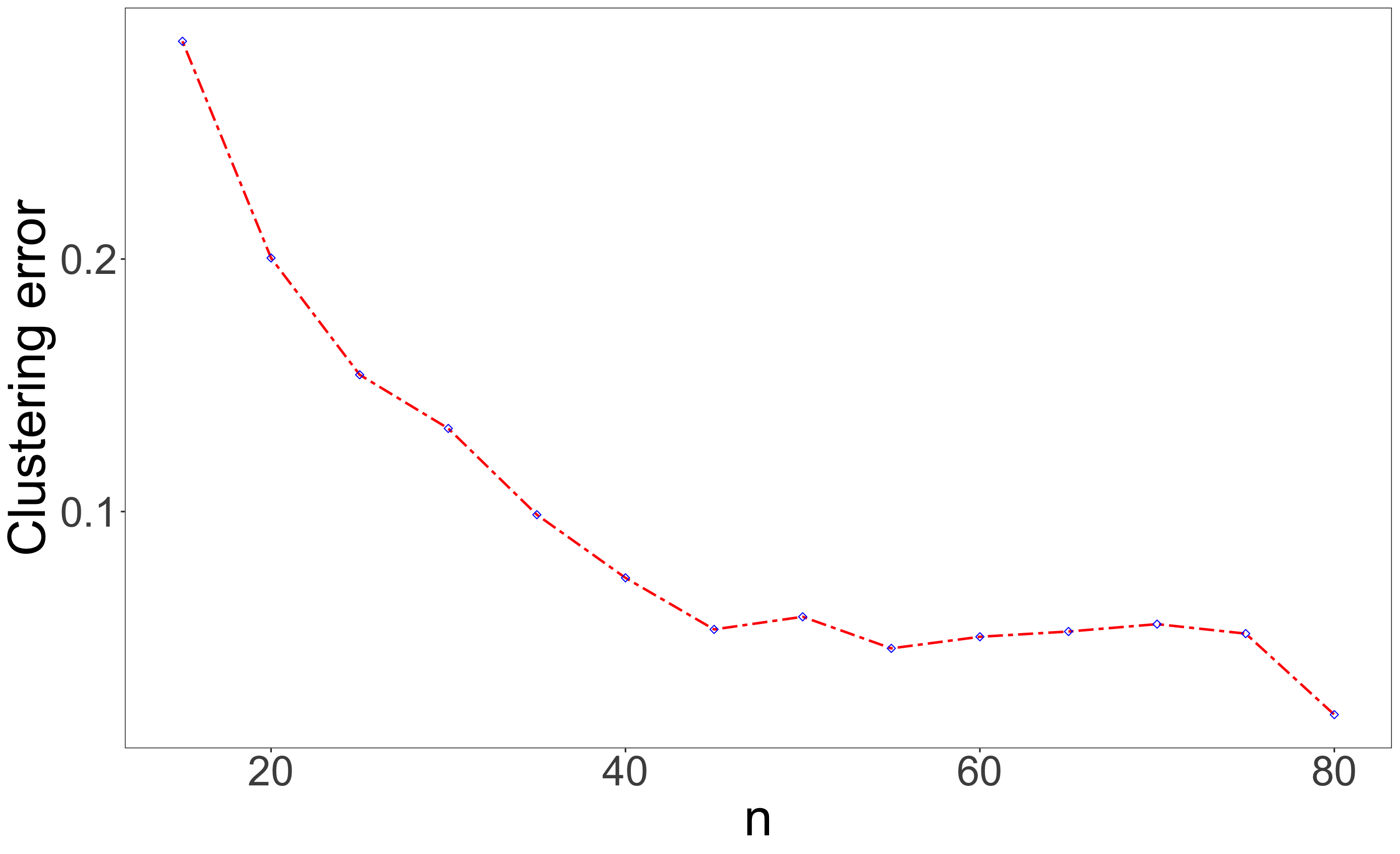}
		\caption{\tiny{Simulation 2-5 for MMLSM: error of network clustering with $n$ varying. Here, $L=80, m=5, r=3$.}}
	\end{subfigure}
	\begin{subfigure}[b]{.49\linewidth}
	\includegraphics[width=\linewidth]{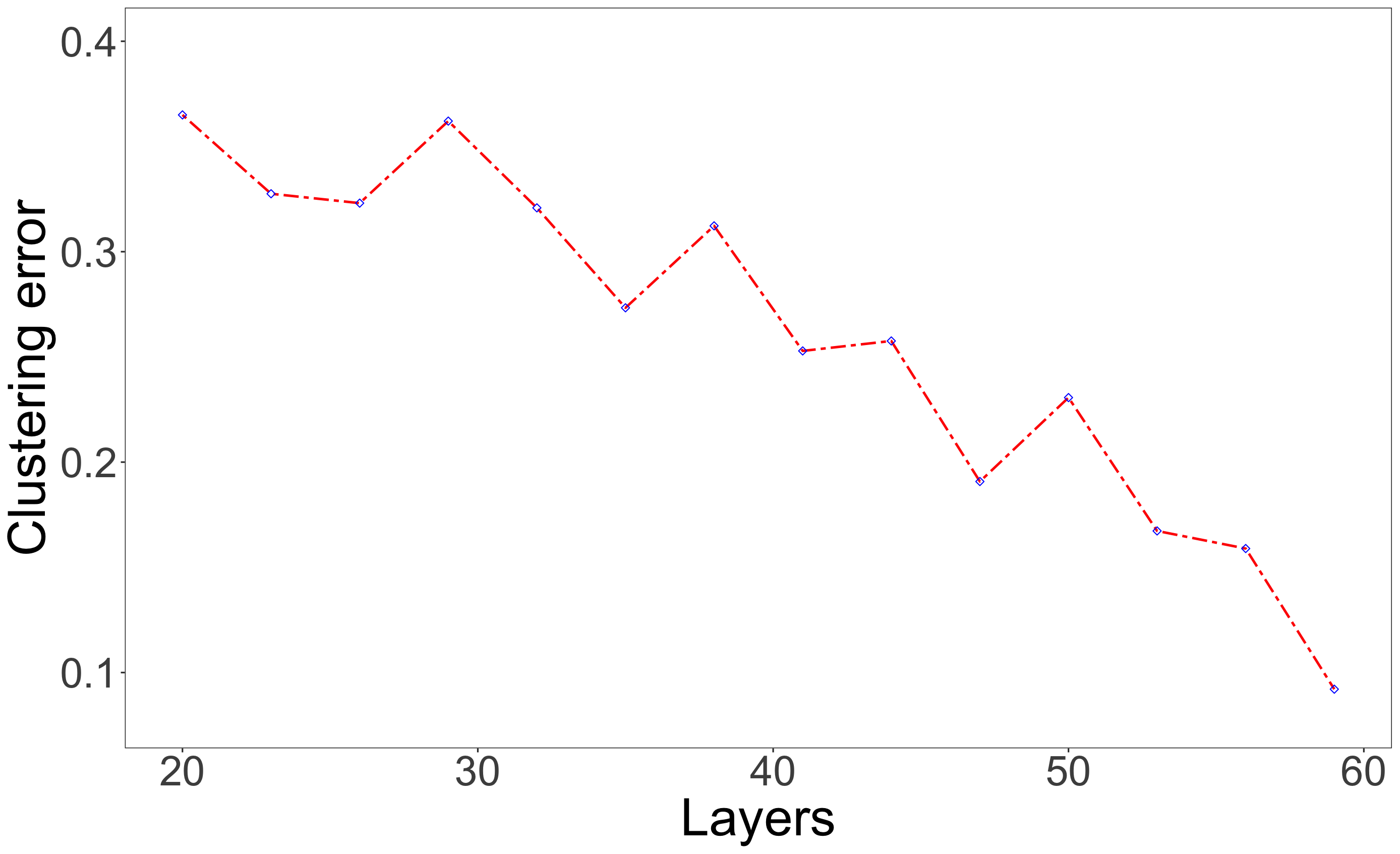}
	\caption{\tiny{Simulation 2-6 for MMLSM: error of network clustering with $L$ varying. Here, $n=50, m=3, r=3$.}}
	\end{subfigure}
	\caption{Error of network clustering with $n$ or $L$ varying}
	\label{fig:multi_clustering}
\end{figure}

Simulation results for various combinations of $n$ and $L$ are shown in Figures \ref{fig:multi_Uerror_nvarying}--\ref{fig:multi_clustering}.
{
The two plots in Figure \ref{fig:multi_Uerror_nvarying} show that the estimation error decreases decently fast as $n$ grows, with large and small $L$, respectively.
Comparing the two panels in Figure \ref{fig:multi_Uerror_Lvarying} echoes our intuitive interpretation of our theoretical analysis (Corollary \ref{col:MMLSM_latent}) that the method's accuracy should improve significantly as $L$ grows for $L\ll n$, and such improvement would become diminishing for $L>n$.  The same observation goes with the accuracy of the downstream clustering, whose result is presented by
Figure \ref{fig:multi_clustering} and consistent with the prediction of our Theorem \ref{MMLSM:network}.
}


\subsection{Simulation 3: hypergraph latent space model} 
We now consider the estimation of latent positions in hyergraphs. Similar to the previous simulations, the dimension of latent space is fixed at $r=3$. 
Here we generate the latent position matrix $U^*$ and interaction tensor $\bC^{\ast}$ similarly to Section \ref{sec:sim-MMLSM}. The low-rank parameter tensor in this simulation is $\bTheta^*=\bC^*\cdot \llbracket U^*, U^*, U^*\rrbracket $.  Each entry of the adjacency tensor is sampled by $[\bA]_{ijk}\overset{{\rm ind.}}{\sim}\text{Bernoulli}(g([\bTheta^*]_{ijk}))$ for $1\le i<j<k\le n$ and $A_{ijk}=0$ if $i,j,k$ are not all distinct.  The lower-triangular entries are also determined by symmetry, slice-wisely. 

{Due to symmetry, it suffices to estimate the singular vectors $U$. }
Using Algorithm \ref{algo:gd} again with the higher-order SVD initialized $U^{(0)}$, we obtain an estimation for $\hat U$.  We define the error measurement for this setting by $\|\widehat U\widehat U^\top-n^{-1}U^*U^{*\top}\|_{\rm F}^2$.

\begin{figure}
	\centering
	\begin{subfigure}[b]{.49\linewidth}
		\includegraphics[width=\linewidth]{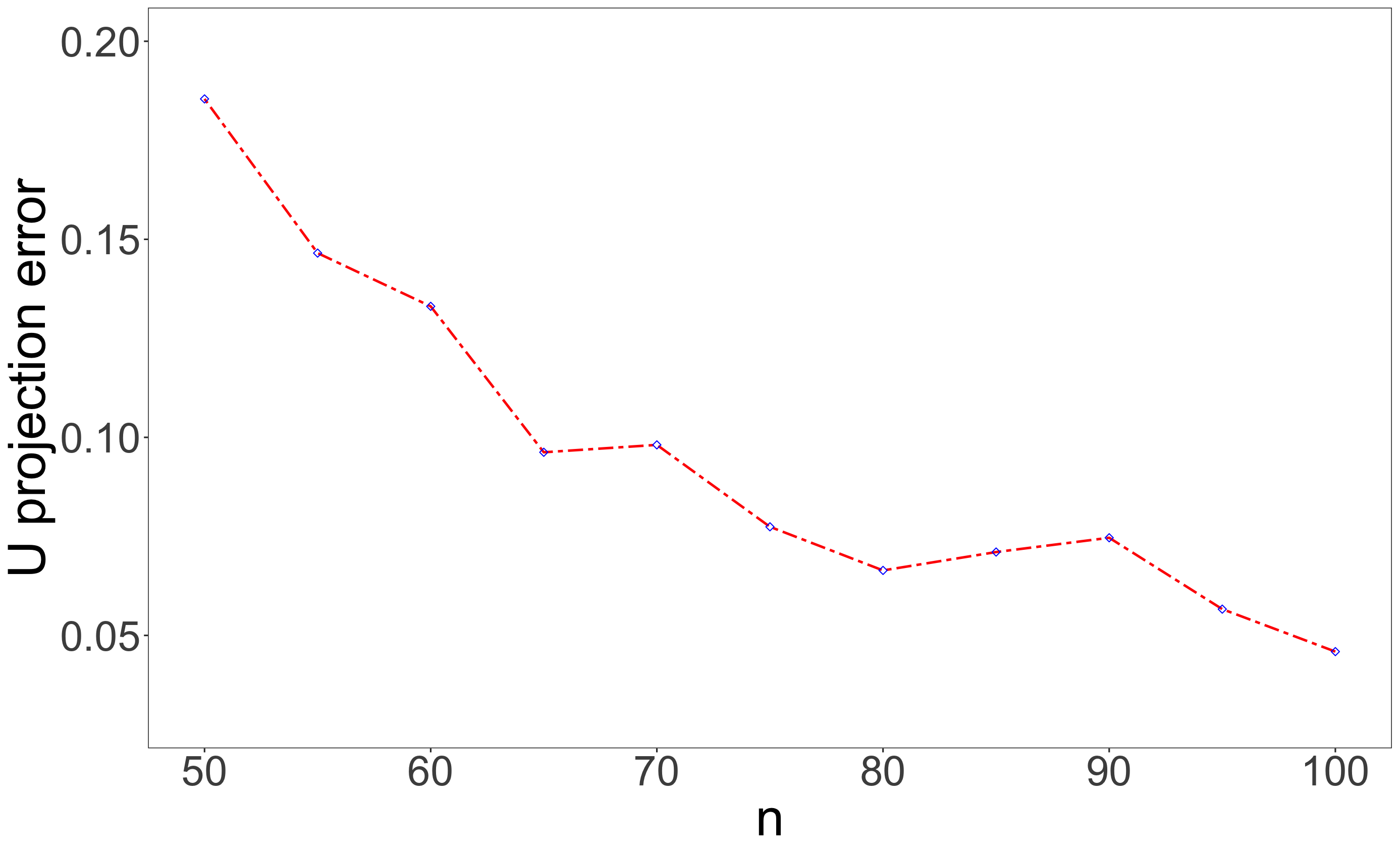}
		\caption{\tiny{Simulation 3-1 for hyper-LSM: error of latent position estimates with $n$ varying. Here, $r=3$.}}
	\end{subfigure}
	\begin{subfigure}[b]{.49\linewidth}
		\includegraphics[width=\linewidth]{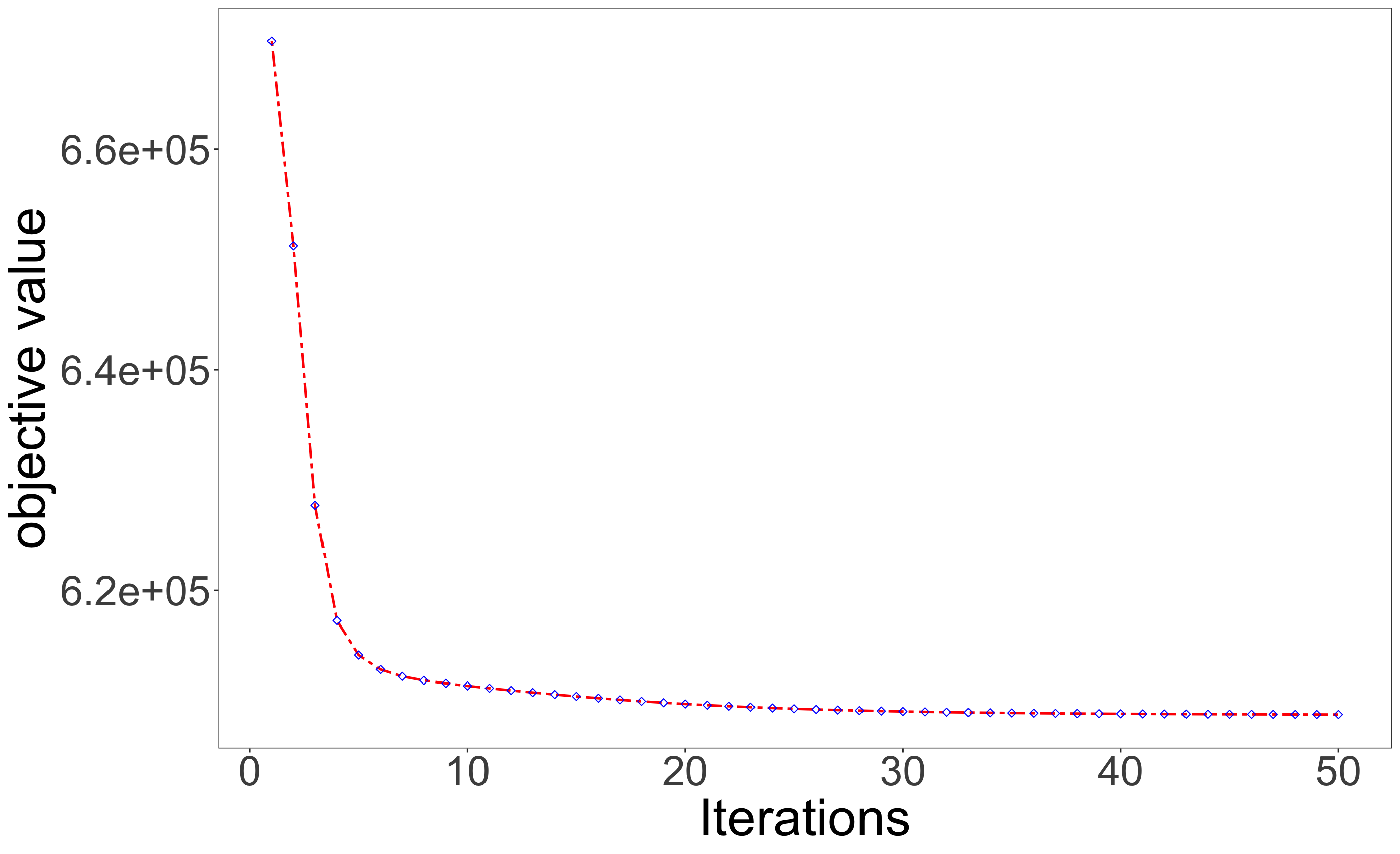}
		\caption{\tiny{Simulation 3-2 for hyper-LSM: value of objective function with respect to the iterations. Here, $n=100, r=3$.}}
	\end{subfigure}
	\caption{Error of latent position estimates with $n$ varying; decrease of objective value with respect to iterations}
	\label{fig:hyper_Uerror_Lvarying}
\end{figure}

{
The two panels of Figure \ref{fig:hyper_Uerror_Lvarying} present the results on accuracy and convergence.  Plot (a) shows that, again, estimation error decreases decently fast in $n$, consistent with our Corollary \ref{col:hLSM_latent}. In plot(b), the objective value shows linear decrement before hitting convergence in just about $5$ iterations.  This demonstrates our method's fast convergence rate and matches our theoretical prediction.
}

\subsection{Simulation 4: dynamic latent space model} 
In the experiment, we set $n=50$, and, for simplicity, $q_j=2$ (the rank of $U_j^*$ in all time intervals) for all $j\in[m]$ as we are interested in change point detection. 
We randomly pick $m=4$ change points $\{t_j\}_{j=2}^m$ uniformly from $\{2,\cdots,T\}$.  For each layer corresponding to the time interval $t\in(t_j,t_{j+1}]$, we generate the latent position matrix $U_j$ and the interaction matrix $C_j$ similarly to that in Section \ref{sec:sim-MMLSM}. 
We apply Algorithm \ref{algo:gd} with { warm initializations attained by HOSVD} and focus on the estimated $\widehat W$. 
We run a row-wise screening procedure {(see Section \ref{sec:thm-dynamic-LSM})} on $\widehat W$ to identify change points. 
{ We measure the performance by the proportion of repeated experiments that correctly identify both the number of change points and their locations.}

\begin{table}[!h]
\centering
\begin{tabular}{|l|c|c|}
\hline
 & Exact detection rate & Accuracy \\ \hline
T=20 & 0.50 & $0.82\pm 0.24$ \\ \hline
T=50 & 0.85 & $0.94\pm 0.14$ \\ \hline
T=80 & 0.96 & $0.99\pm 0.07$ \\ \hline
\end{tabular}
\caption{Simulation 4-1 for dynamic LSM: rate of exact detection and accuracy over $100$ simulations}
\label{table:sim4-1}
\end{table}

Table \ref{table:sim4-1} reports the result over 100 simulations. 
The exact detection rate increases as $T$ grows, which aligns with our Corollary \ref{DLSM:latent}.


\section{Data examples}\label{sec:real-data}
In this section, we demonstrate the merits of our methods in node embedding and link predictions on two real-world datasets. 
\subsection{Trade flow multi-layer network from UN Comtrade}\label{subsec:real_data_trade}
The multi-layer network data are constructed based on the international commodity trade data collected from the \textit{UN Comtrade Database} (\href{url}{https://comtrade.un.org}). The dataset contains annual trade information for countries/regions from different continents in 2019, where, for ease of presentation, we only focus on the top representative $48$ countries/regions ranked by the exports of goods and services in US dollars. Each layer represents a different type of commodities classified into $97$ categories based on the 2-digit HS code (\href{url}{https://www.foreign-trade.com/reference/hscode.htm}).  
{
	For every two nodes $i$ and $j$, we convert the two weighted edges $w_{i\to j}, w_{j\to i}\geq 0$ in the original data into one binary directed edge:  if $w_{i\to j}>w_{j\to i}$ then we set $A_{i\to j}=1, A_{j\to i}=0$, indicating a trade surplus of $i$ in its trade with $j$, and vice versa. 
}
The adjacency tensor $\bA$ is defined in the way such that $[\bA]_{ijl}=1$  if country $i$ exports to country $j$ in terms of commodity type $l$.  We remove empty layers and obtain a binary adjacency tensor $\bA$ of size $48\times 48\times 97$. 

We apply our Algorithm \ref{algo:gd}, initialized by HOOI \citep{zhang2018tensor, ke2019community}, to $\bA$ and obtain an estimated $\widehat W$. 
{ Empirical evidence (the numerical scales of the leading eigenvalues and the plot of $\hat W$ rows projected onto the first two principal components) suggest that $r=5$ and $m=2$ lead to a most interpretable model fit.}
Then we apply K-means clustering on the rows of $\widehat W$ with $m=2$ clusters and report the result in Table \ref{table:clustering}. 
It is interesting to observe that {\it bio-related} daily products including animal \& animal products, vegetable products, over half of foodstuffs fall into cluster $1$, most of which are all products of low durability. On the other hand, most {\it industrial} products including main parts of chemicals \& allied industries, plastic/rubbers, stone/glass, machinery/electrical, aircraft, spacecraft, optical, photographic, etc., and clocks and watches constitute cluster $2$.

\begin{figure}
	\centering
	\includegraphics[scale=0.16]{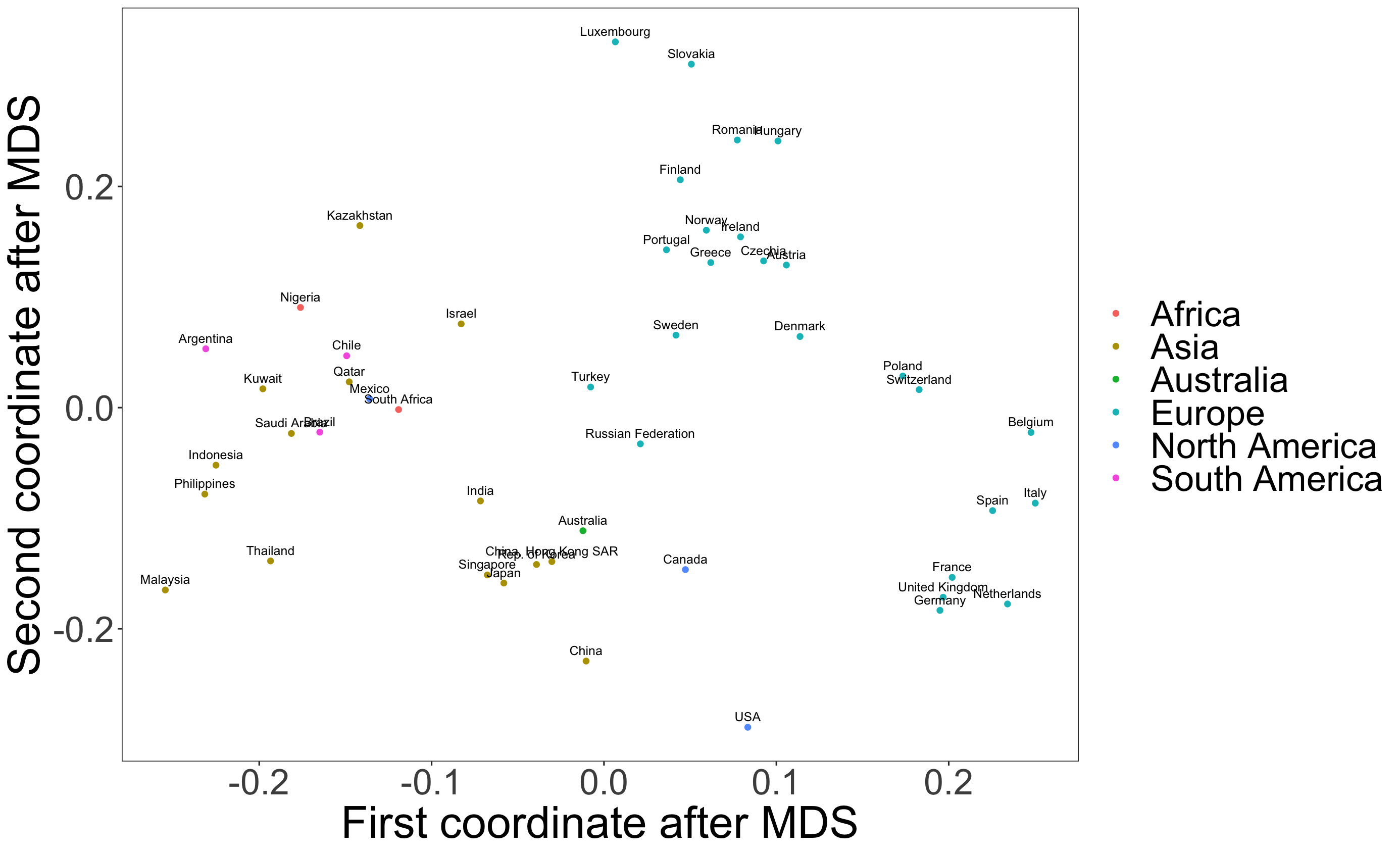}
	\caption{{Latent positions of countries/regions for layers of bio-related daily products}}
	\label{fig:latent_bio}
\end{figure}

\begin{figure}
	\centering
	\includegraphics[scale=0.16]{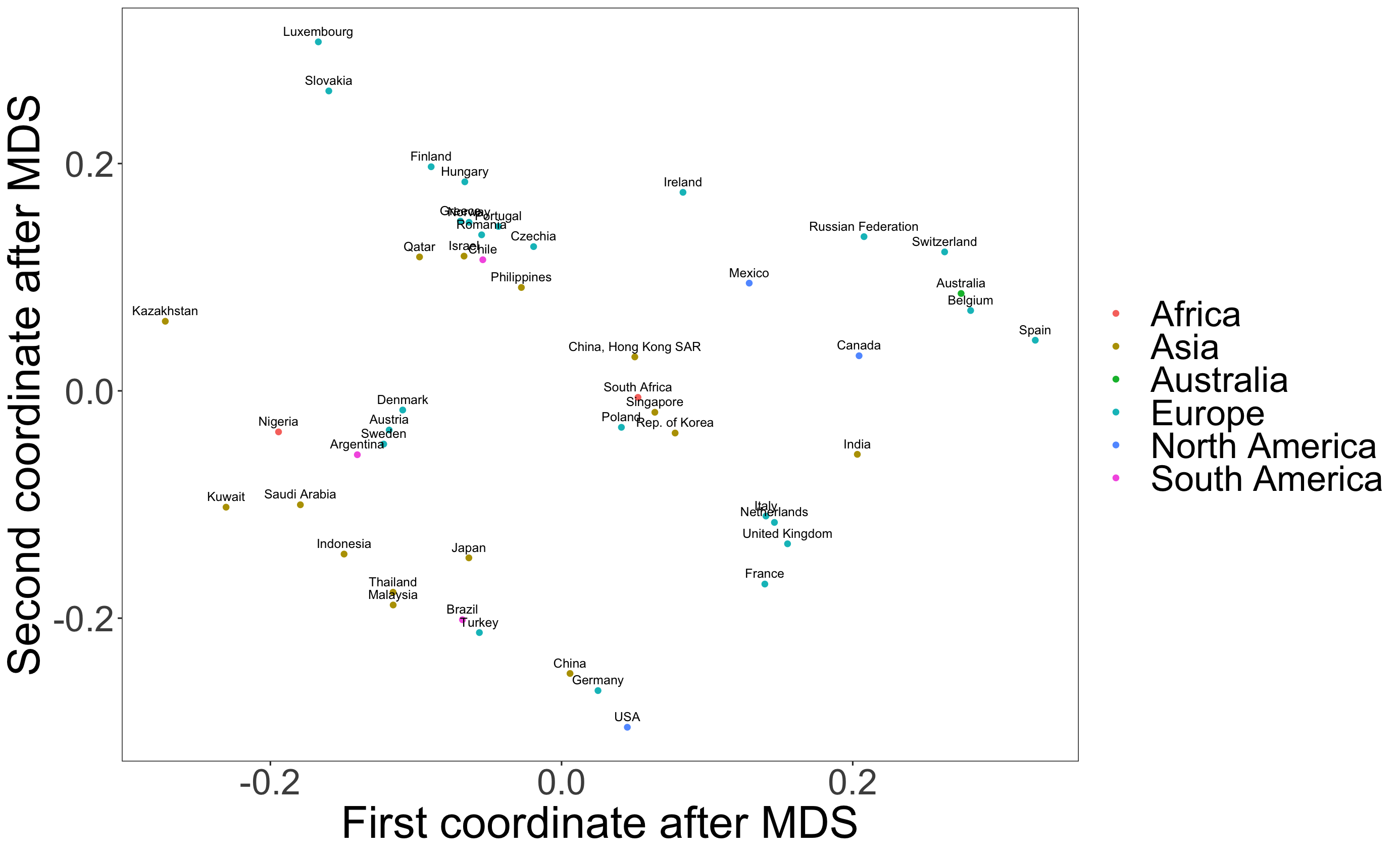}
	\caption{{Latent positions of countries/regions for layers of industrial products}}
	\label{fig:latent_industrial}
\end{figure}

Based on the layers clustering in Table~\ref{table:clustering}, 
we further investigate the shared trade pattern among different countries/regions.  Specifically, we construct a sub-tensor of size $49\times 49\times 20$ from cluster $1$ for {\it bio-related} commodities, and a sub-tensor of size $49\times 49\times 15$ from cluster $2$ for {\it industrial} commodities. 
A scientifically interesting question is to compare the latent position representations in these two groups of layers.
Toward this end, we apply Algorithm \ref{algo:gd} with $r=3$ and $m=1$ on these two sub-tensors. Since the trading flows are directed, the left singular vectors $\widehat U$ and right singular vectors $\widehat V$ are distinct. 
It turns out that the latent position of imports $\widehat V$ provide clear and interpretable results.  We further perform multidimensional scaling (MDS) on the rows of $\widehat V_{\textsf{bio}}$ and $\widehat V_{\textsf{ind}}$, projecting them into $\mathbb{R}^2$ for visualization. In Figure \ref{fig:latent_bio} and Figure \ref{fig:latent_industrial} plot the projected embedding of countries/regions according to their latent positions $\widehat V_{\textsf{bio}}$ and $\widehat V_{\textsf{ind}}$ after MDS, with nodes being colored by corresponding continents.

The latent positions of countries/regions for the two groups of layers exhibit different patterns. In Figure \ref{fig:latent_bio}, latent positions for countries/regions in the same continent in general are close to each other, which to some extent reflects geographic proximity relations. We could observe several ``clusters'' such as European countries in the bottom right and the top middle; Hong Kong SAR, Singapore, South Korea (three out of Four Asian Tigers) and Japan in the bottom middle; Indonesia, Philippines, Thailand and Malaysia (known as Tiger Cub Economies). This is reasonable since for commodities of low durability, regional trade partnerships usually dominate the inter-continental ones. However, it is interesting to notice those outliers. Three large economies China, USA and Canada are relatively close in latent positions even though China is not geographically close to USA and Canada, since they export a large amount of bio-related/daily products to all other countries. Three South America countries (Argentina, Chile and Brazil), two Africa countries (Nigeria and South Africa) and Mexico are embedded closer to the Middle East countries, as these nations import similar products mainly from several largest exporting economies. In Figure \ref{fig:latent_industrial}, the geographical impact, to some extent, is weakened. Germany, originally near United Kingdom, France and Netherlands in Figure~\ref{fig:latent_bio}, is now clustered closer to China and USA, largely due to the fact that they are all big industrial nations with a huge demand of importing industrial raw materials. Denmark, Austria and Sweden (three out of Frugal Four) are mixed with countries/regions from Asia, South America and Africa, indicating that these developed industrial nations are heavily depending on imported industrial products from those developing countries. Australia is relocated nearer to European nations, which can be explained by their similarity of imported goods which outweighs the geographical closeness to Asian nations. Overall, high durability for industrial products means relatively low cost in freight and hence the trade partnerships are less regionally restricted and more related to their resemblance and connection in terms of industrial products.

\begin{figure}
	\centering
	\includegraphics[scale=0.13]{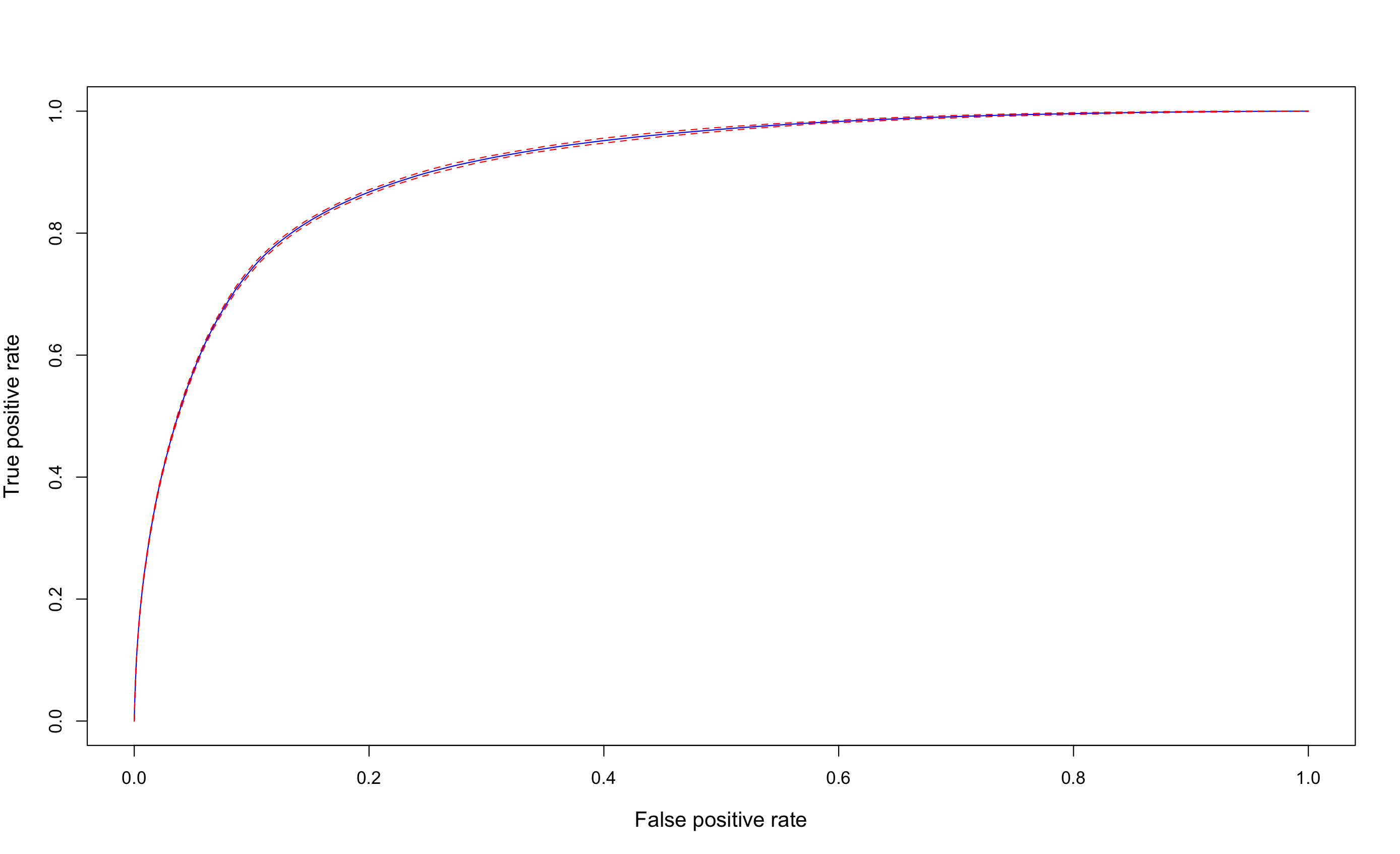}
	\caption{ROC curve for link prediction with $99.9\%$ CI (red dashed line)}
	\label{fig:link_pred}
\end{figure}

To further assess the performance of the estimated latent positions, we apply our method to this dataset for the task of link prediction. We adopt the evaluation metric for link prediction in \cite{zhao2017link}. Specifically, we set $20\%$ of entries of $\bA$ ($10\%$ randomly selected out of non-zero entries and $10\%$ out of zero entries) to be $0$ and construct the test tensor data $\bA_{\textsf{test}}$. Then Algorithm \ref{algo:gd} is applied to $\bA_{\textsf{test}}$ to get the estimated probability tensor $\widehat{\mathbf{P}} =g(\hat \bTheta)$. We evaluate the link prediction performance on those randomly deleted entries by $\text{AUC}$, which is defined to be $\text{the area under the ROC curve}$. By  30 simulations, we observe $\text{AUC}=0.910(\pm 0.001)$.  The $\text{ROC}$ curve with $99.9\%$ confidence interval is displayed in Figure \ref{fig:link_pred}.
\begin{table}[!htbp]
\centering
\begin{tabular}{|l|l|}
\hline
Commodity cluster 1 & \begin{tabular}[c]{@{}l@{}}01-05 Animal \& Animal Products (100\%) 06-15 Vegetable Products (100\%)\\ 16-18,23-24 Foodstuffs (56\%) 26 Mineral Products (33\%) \\ 31,36-37 Chemicals \& Allied Industries (27\%)\\ 41,43 Raw Hides, Skins, Leather \& Furs (66\%)\\ 45-47 Wood \& Wood Products (50\%) 50-55,57-58,60 Textiles (64\%)\\ 66-67 Footwear / Headgear (50\%) 75,78-81 Metals (45\%)\\ 86,89 Transportation (50\%) 92,93,97 Miscellaneous (37.5\%)\end{tabular}                                   \\ \hline
Commodity cluster 2 & \begin{tabular}[c]{@{}l@{}}19-22 Foodstuffs (44\%) 25,27 Mineral Products (67\%)\\ 28-30,32-35,38 Chemicals \& Allied Industries (73\%) \\ 39-40 Plastics / Rubbers (100\%)\\ 42 Raw Hides, Skins, Leather, \& Furs (33\%)\\ 44,48-49 Wood \& Wood Products (50\%) 56,59,61-63 Textiles (36\%)\\ 64,65 Footwear / Headgear (50\%) 68-71 Stone / Glass (100\%)\\ 72-74,76,82-83 Metals (55\%) 84-85 Machinery / Electrical (100\%) \\ 87-88 Transportation (50\%) 90-91,94-96,99 Miscellaneous (62.5\%)\end{tabular} \\ \hline
\end{tabular}
\caption{Network clustering results of 97 commodity layers, $\%$ denote the proportion of number of layers in the same category characterized by HS Code}
\label{table:clustering}
\end{table}


\subsection{Disease hypergraph network from MEDLINE}
In the second data example, we analyze a hypergraph originated from the \textit{MEDLINE Database} (\href{url}{www.nlm.nih.gov/medline}). The database contains more than 27 million papers indexed by Medical Subject Headings (MeSH) concentrated on biomedicine. We focus on 12,637 papers published in 1960 annotated with 318 MeSH terms categorized into two types: Neoplasms (C04) and Nerve System Diseases (C10). In the constructed hypergraph network, the nodes are MeSH terms, and the hyperedges of sizes $1,2,3,4,5$ are formed among the nodes annotated by the same paper. For simplicity, we only deal with {\it triadic} relations. 
We remove the hyperedges of size $1$ and greater than $3$, and add {\it one} additional dummy node for those hyperedges of size $2$. We further abandon nodes of degrees less than $4$ to eliminate those with insignificant information. Finally, we obtain an adjacency tensor $\bA$ sized $ 166\times 166\times 166$ (including {\it one} dummy node) of the hypergraph network with $n=165$ MeSH terms, among which, $115$ fall into class C04 and $50$ are in class C10.

\begin{figure}
	\centering
	\includegraphics[scale=0.13]{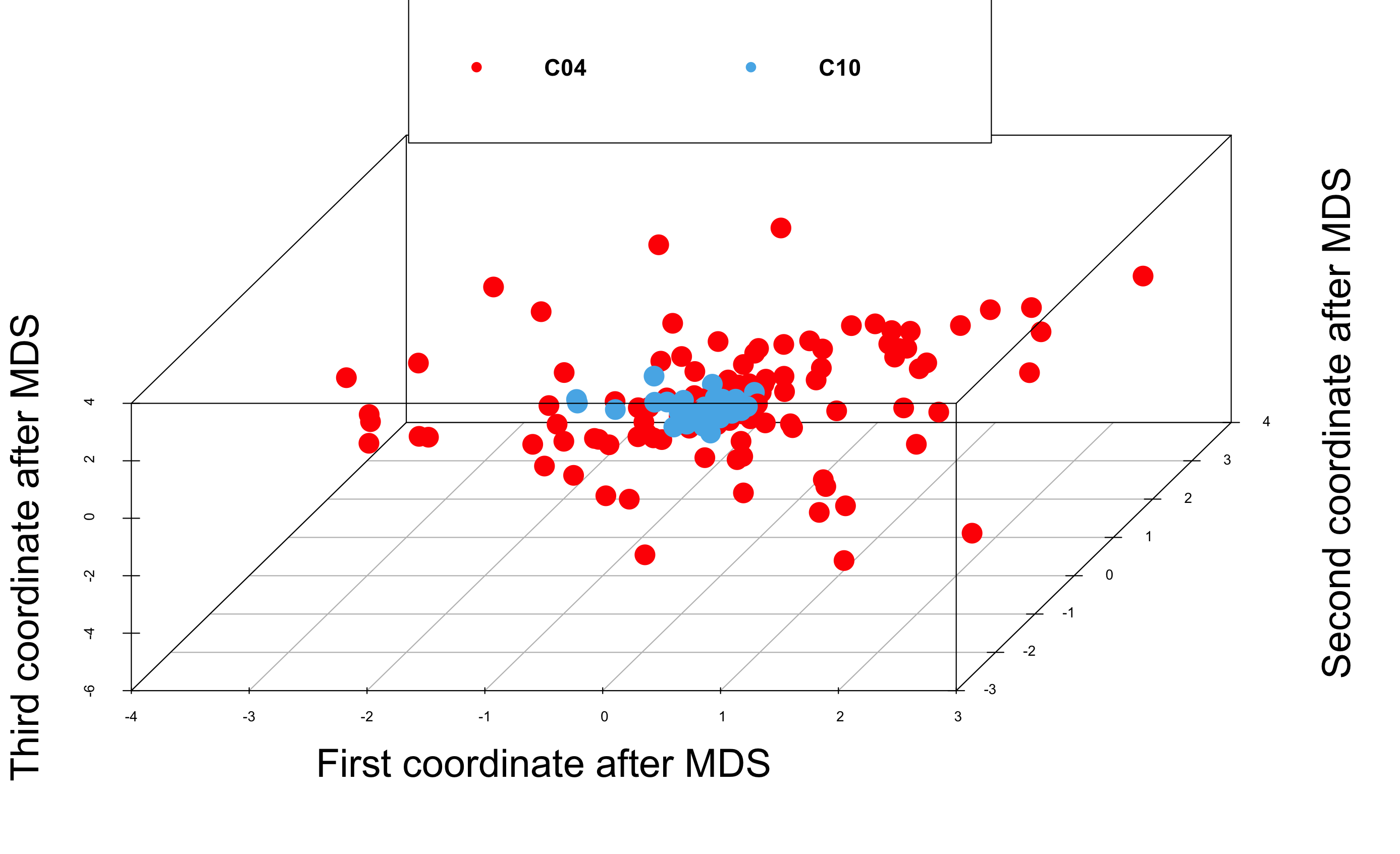}
	\caption{Initialized latent positions $\widehat U^{(0)}$}
	\label{fig:latent_hyper_int}
\end{figure}

\begin{figure}
	\centering
	\includegraphics[scale=0.13]{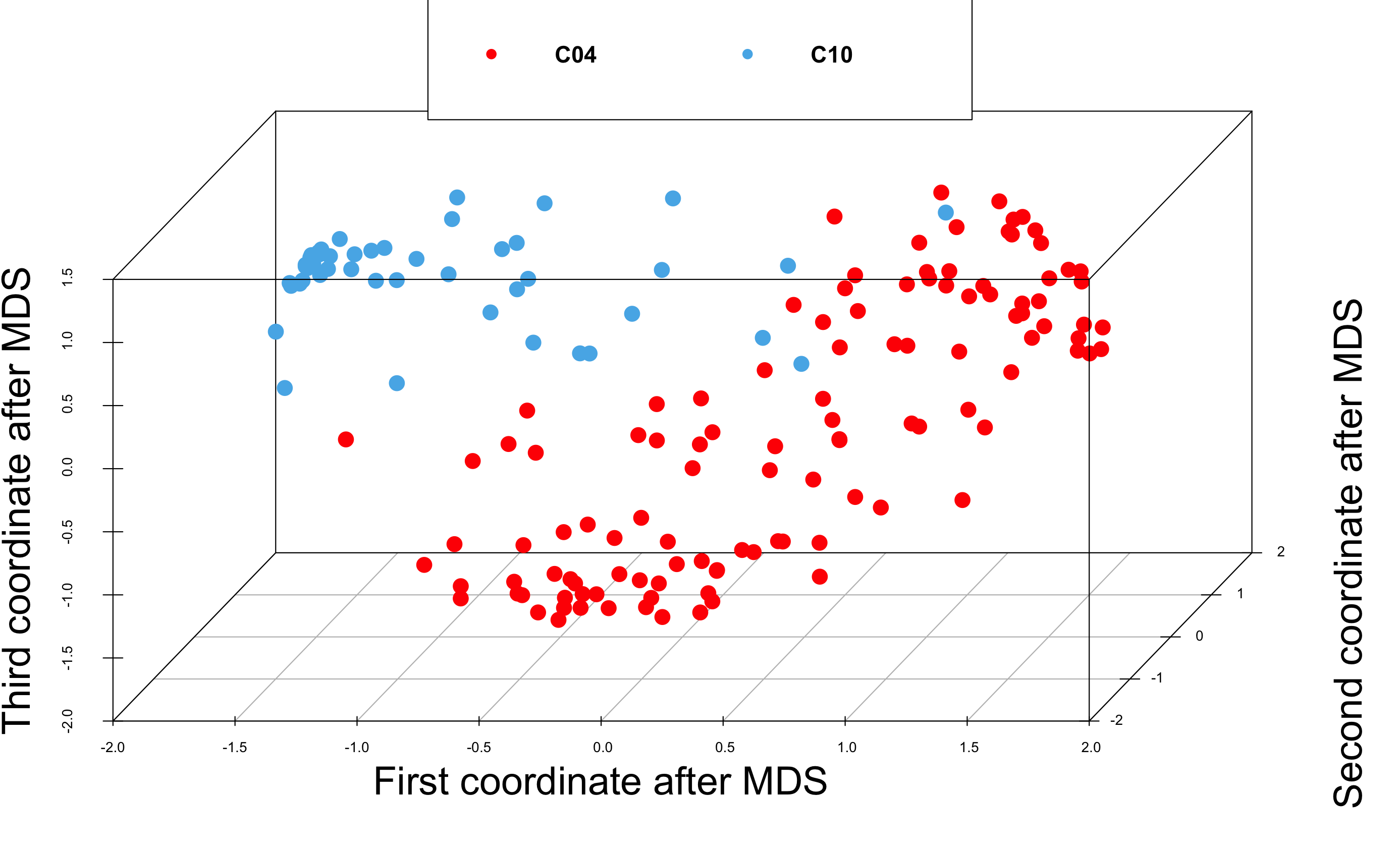}
	\caption{Latent positions $\widehat U$ output by Algorithm~\ref{algo:gd}}
	\label{fig:latent_hyper_final}
\end{figure}

We initiate Algorithm \ref{algo:gd} by $2$ iterations of HOOI, run it on $\bA$, and obtain the estimated $\widehat U$ positions, in which, we set $r=5$. Similarly, we perform MDS on $\widehat U$ for visualization. The result of node embedding is plotted in Figure \ref{fig:latent_hyper_int}-\ref{fig:latent_hyper_final}. 
Started with an initialization $ U^{(0)}$ in Figure \ref{fig:latent_hyper_int}, where the two types of disease are mixed together, the eventual estimation $\widehat U$ in Figure \ref{fig:latent_hyper_final} shows a clear separation between the two clusters. Indeed, K-means clustering on the rows of $U^{(0)}$ and $\widehat U$ with $K=2$ clusters would produce $49.7\%$ and $3.64\%$ misclassification error rates, respectively. This clearly demonstrates the effectiveness and utility of our algorithm.

\begin{figure}
	\centering
	\includegraphics[scale=0.13]{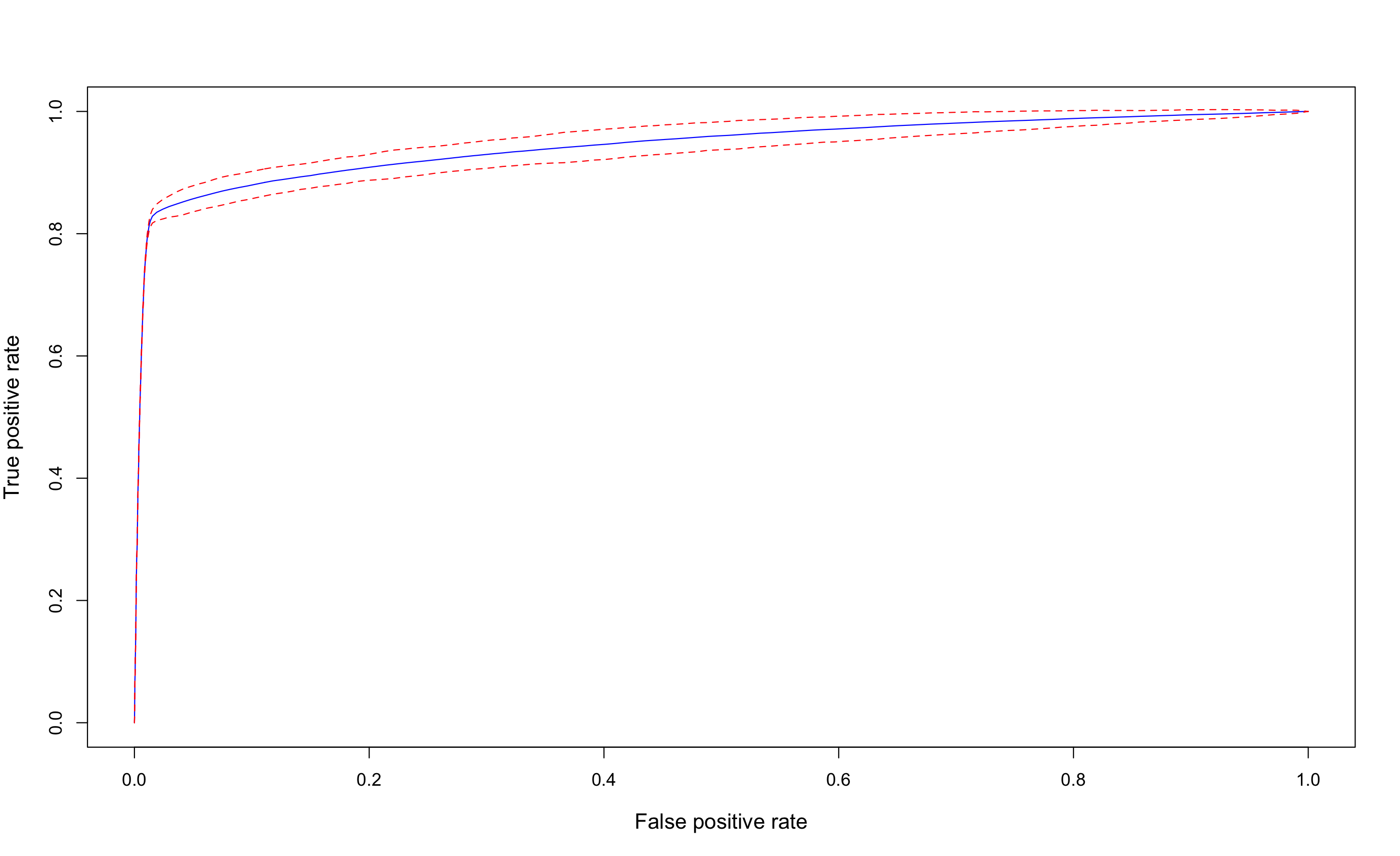}
	\caption{ROC curve for link prediction with $99.9\%$ CI (red dashed line)}
	\label{fig:hyper_link_pred}
\end{figure}

Finally, run a link prediction similar to that described in Section \ref{subsec:real_data_trade}. Since the MeSH network is extremely sparse ($99.9\%$ entries of $\bA$ are $0$'s), we construct the test tensor $\bA_{\textsf{test}}$ by randomly setting half of $1$'s and the same number of $0$'s to be $0$, {on which spots the accuracy of link prediction will be evaluated.  This set up is constructed towards a balanced share between 0/1 values and a numerically stabler evaluation.  Also in light of the observed sparsity, we choose a smaller}  scale parameter $\sigma$ in the link prediction here
We obtain $\text{AUC}=0.944(\pm 0.005)$ over $30$ simulations.  The $\text{ROC}$ curve with $99.9\%$ confidence interval is presented in Figure \ref{fig:hyper_link_pred}.

{
	\section{Concluding remarks}
	
	In this paper, we propose a novel unified method for investigating the higher-order interactions in network data.  Our framework is general in its abstraction of the concept ``layer'', which could be either a third participant in a dyadic relationship/interaction, or it could index the multiple interactions between two nodes, or encode the time stamp in a dynamic network setting.  Our model also allows the data generation scheme to connect to the interaction latent positions via a generalized linear link function.  It covers several popular mainstream higher-order network models, including multilayer networks, hypergraphs and dynamic networks, as special cases.  Our proposed method is therefore versatile and widely applicable.  Further, we developed original theory that rigorously guarantees the good performance of the algorithm and quantitatively understand the finite-sample error bounds over our method's iterations.
	
	There are a number of interesting directions of future work.  In this work, we focused on binary network interactions.  We expect our algorithm and analysis can be expanded to some weighted edge generation schemes, such as exponential distribution and some sub-Gaussian distributions.  But given the volume of work even under the Bernoulli model, we stick to binary edges in this paper and leave the direction for future investigation.  Second, we constrain our data generation scheme to generalized linear link functions.  While this formulation decently caters to the need of many real-life data analysis tasks, it is interesting to expand the methodology to more general link functions.  A third interesting but much more challenging future exploration is to account for dependency between the higher-order interactions.
}

\newpage

\bibliographystyle{plainnat}
\bibliography{reference.bib}
\section{Proofs}
For the ease of presentation, throughout the proofs we use $U_1,U_2,U_3$ to denote $U,V,W$ respectively together with their variants of different superscripts and subscripts (e.g. $U_k^*$, $U_k^{(t)}$ for $k=1,2,3$). 
\subsection{Proof of Lemma \ref{lem:solveC}}\label{sec::proof::lemma::2}
To see the convexity of $\ell_{n}(\bC\cdot\llbracket U_1,U_2,U_3\rrbracket)$, first note that
$$\text{vec}(\mathcal{M}_1(\bC\times_1U_1\times_2U_2\times_3U_3))=(U_3\otimes U_2\otimes U_1)\cdot\text{vec}(\mathcal{M}_1(\bC))$$
Denote $c_v:=\text{vec}(\mathcal{M}_1(\bC))\in \mathbb{R}^{r_1r_2r_3}$, $\mathring{U}:=U_3\otimes U_2\otimes U_1\in\mathbb{R}^{n_1n_2n_3\times r_1r_2r_3}$ and $\mathring{U}_{ijk}:=U_3(k,:)^T\otimes U_2(j,:)^T\otimes U_1(i,:)^T\in \mathbb{R}^{r_1r_2r_3}$, then the decomposition of $\Theta_{ijk}$ can be written as
$$\Theta_{ijk}=\sum_{l_1l_2l_3}C_{l_1l_2l_3}U_1(i,l_1)U_2(j,l_2)U_3(k,l_3)=\langle c_v,\mathring{U}_{ijk}\rangle$$
The objective $\ell_{n}(\bC\cdot\llbracket U_1,U_2,U_3\rrbracket)$ essentially becomes
\begin{align*} 
f(c_v):&=-\sum_{ijk}\left[A_{ijk}\log p(\langle c_v,\mathring{U}_{ijk}\rangle)+(1-A_{ijk})\log (1-p(\langle c_v,\mathring{U}_{ijk}\rangle))\right]
\end{align*}
and calculating the hessian of $f(\cdot)$ gives that 
\begin{align*} 
\nabla^2 f(c_v)&=\sum_{ijk}\Bigg[A_{ijk}\left(\left(\frac{p^\prime(\langle c_v, \mathring{U}_{ijk}\rangle)}{p(\langle c_v,\mathring{U}_{ijk}\rangle)}\right)^2-\frac{p^{\prime\prime}(\langle c_v, \mathring{U}_{ijk}\rangle)}{p(\langle c_v,\mathring{U}_{ijk}\rangle)}\right)\\
&+(1-A_{ijk})\left(\left(\frac{p^\prime(\langle c_v, \mathring{U}_{ijk}\rangle)}{1-p(\langle c_v,\mathring{U}_{ijk}\rangle)}\right)^2+\frac{p^{\prime\prime}(\langle c_v, \mathring{U}_{ijk}\rangle)}{1-p(\langle c_v,\mathring{U}_{ijk}\rangle)}\right)\Bigg]\mathring{U}_{ijk}\mathring{U}_{ijk}^T
\end{align*}
For any $x\in\mathbb{R}^{r_1r_2r_3}$ such that $\|x\|_2= 1$, by Assumption \ref{assump:link} we have
\begin{align*} 
\langle \nabla^2 f(c_v)x, x\rangle&=\sum_{ijk}\Bigg[A_{ijk}\left(\left(\frac{p^\prime(\langle c_v, \mathring{U}_{ijk}\rangle)}{p(\langle c_v,\mathring{U}_{ijk}\rangle)}\right)^2-\frac{p^{\prime\prime}(\langle c_v, \mathring{U}_{ijk}\rangle)}{p(\langle c_v,\mathring{U}_{ijk}\rangle)}\right)\\
&+(1-A_{ijk})\left(\left(\frac{p^\prime(\langle c_v, \mathring{U}_{ijk}\rangle)}{1-p(\langle c_v,\mathring{U}_{ijk}\rangle)}\right)^2+\frac{p^{\prime\prime}(\langle c_v, \mathring{U}_{ijk}\rangle)}{1-p(\langle c_v,\mathring{U}_{ijk}\rangle)}\right)\Bigg]x^T\mathring{U}_{ijk}\mathring{U}_{ijk}^Tx\\
&\ge \gamma_\alpha \sum_{ijk}x^T\mathring{U}_{ijk}\mathring{U}_{ijk}^Tx=\gamma_\alpha x^T\left(\sum_{ijk}\mathring{U}_{ijk}\mathring{U}_{ijk}^T\right)x=\gamma_\alpha
\end{align*}
which implies that $\nabla^2 f(c_v)\succeq \gamma_\alpha \boldsymbol{I}_{r_1r_2r_3}$. 

\subsection{Proof of Lemma \ref{lem:xibound}}
Denote $\bX:=\nabla \ell_{n} (\bTheta^*)-\E\nabla\ell_{n}(\bTheta^*)$, by the definition of $\textsf{Err}_{\br}$ we have
\begin{align*}
	\textsf{Err}_{\br} &=\sup_{\substack{\bTheta\in \mathbb{R}^{n_1\times n_2 \times n_3}, \|\bTheta\|_F\le 1 \\  \text{rank}(\bTheta)=(r_1,r_2,r_3)}}\langle X,\bTheta\rangle=\sup_{\substack{\bC\in \mathbb{R}^{r_1\times r_2 \times r_3}, \|\bC\|_F\le 1 \\  \op{U_k}\le 1,k=1,2,3}}\langle X,\bC\times_1U_1\times_2U_2\times_3U_3\rangle
\end{align*}
Now we let
$$
(\bC^\dagger,U_1^\dagger,U_2^\dagger,U_3^\dagger)=\argmax_{\substack{\bC\in \mathbb{R}^{r_1\times r_2 \times r_3}, \|\bC\|_F\le 1 \\  \op{U_k}\le 1,k=1,2,3}}\langle X,\bC\times_1U_1\times_2U_2\times_3U_3\rangle
$$
and let $\mathcal N_C^\epsilon$ be an $\epsilon$-net of $\{\bC\in\R^{r_1\times r_2\times r_3}:\fro{\bC}\le 1\}$ and $\mathcal N_k^\epsilon$ be an $\epsilon$-net of $\{U\in\R^{n_k\times r_k}:\op{U}\le 1\}$ for $k=1,2,3$. A simple fact is that
$$
\abs {\mathcal N_C^\epsilon}\le \left(\frac{2+\epsilon}{\epsilon}\right)^{r_1r_2r_3},\quad \abs {\mathcal N_k^\epsilon}\le \left(\frac{2+\epsilon}{\epsilon}\right)^{n_kr_k}\text{~for~} k=1,2,3
$$
By the definition of $\epsilon$-net, there exists $\widetilde{\bC}\in \mathcal N_C^\epsilon$ and $\widetilde {U}_k\in \mathcal{N}_k^\epsilon$ such that 
$$
\fro{\widetilde C-C^\dagger}\le \epsilon,\quad \op{\widetilde U_k-U_k^\dagger}\le \epsilon
$$
Hence we have
\begin{align*}
	\textsf{Err}_{\br}&=\langle\bX, \widetilde \bC\times_1\widetilde U_1\times_2\widetilde U_2\times_3\widetilde U_3\rangle+\langle\bX, \bC^\dagger \times_1U_1^\dagger\times_2U_2^\dagger\times_3U_3^\dagger\rangle-\langle\bX, \widetilde \bC\times_1\widetilde U_1\times_2\widetilde U_2\times_3\widetilde U_3\rangle\\
	&\le \langle\bX, \widetilde \bC\times_1\widetilde U_1\times_2\widetilde U_2\times_3\widetilde U_3\rangle+4\epsilon\cdot  \textsf{Err}_{\br} 
\end{align*}
Note that for any $\bC\in \mathcal N_C^\epsilon$ and $U_k\in \mathcal N_k^\epsilon$, the Hoeffding inequality gives
$$
\Prob\Big(\abs{\langle\bX,\bC\times_1U_1\times_2U_2\times_3U_3 \rangle}\ge \frac{t}{2}\Big)\le \exp\left(-{\frac{t^2}{2\zeta_\alpha^2}}\right)
$$
where we use $X_{ijk}\in\left\{-\frac{p^\prime(\Theta_{ijk})}{p(\Theta_{ijk})},\frac{p^\prime(\Theta_{ijk})}{1-p(\Theta_{ijk})}\right\}$ and the definition of $\zeta_\alpha$. Now taking $\epsilon=1/8$ and using a union bound, we conclude that 
\begin{align*}
	\Prob (\textsf{Err}_{\br} \ge t)&\le \Prob \Big(\max_{\substack{\bC\in N_C^\epsilon\\ U_k\in N_k^\epsilon, k=1,2,3}}\langle\bX,\bC\times_1U_1\times_2U_2\times_3U_3 \rangle\ge \frac{t}{2}\Big)\\
	&= \Prob \Big(\bigcup_{\substack{\bC\in N_C^\epsilon\\ U_k\in N_k^\epsilon, k=1,2,3}}\left\{\langle\bX,\bC\times_1U_1\times_2U_2\times_3U_3 \rangle\ge \frac{t}{2}\right\}\Big)\\
	&\le C^{r_1r_2r_3+\sum_{k=1}^3n_kr_k}\exp\left(-\frac{t^2}{2\zeta_\alpha^2}\right)
\end{align*}
The proof is completed by adjusting the constant.
\subsection{Proof of Theorem \ref{thm:main}}
\subsubsection{Notations and conditions}
For the notational simplicity, we interchangeably write $\bC^*\cdot\llbracket U_1^*,U_2^*,U_3^*\rrbracket$ and $\bC^*\times_1U_1^*\times_2U_2^*\times_3U_3^*$ to denote the multilinear product throughout the proof. We denote $O_k^{(t)}:=\text{argmin}_O\fro{U^{(t)}-n_k^{-1/2}U_k^*O}$, $d_k^{(t)}:=d_\textsf{f}(U^{(t)},n_k^{-1/2}U_k^*)$, and
$d_{\textsf{C}}^{(t)}:=\fro{\bC^{(t)}-\bC^*\cdot\llbracket O_1^{(t)\top},O_2^{(t)\top},O_3^{(t)\top}\rrbracket}$. We also denote $\bar{r}=\max_k r_k$ and
$$\widetilde U_k^{(t)}:=U^{(t)}-\nabla_{U_k} \ell_n(\bTheta^{(t)})$$
$$\widetilde{\bTheta}^{(t)}:=(\bC^*\cdot\llbracket O_1^{(t)\top},O_2^{(t)\top},O_3^{(t)\top}\rrbracket)\times_1U_1^{(t)}\times_2U_2^{(t)}\times_3U_3^{(t)}$$ 
As we noted before, $U_k^{(t)}$ is an estimate of $n_k^{-1/2}U_k^*$. Without loss of generality, throughout the proof we assume $U_k^*$ is multiplied by the scale factor $n_k^{-1/2}$ and the core tensor $\bC^\ast$ is multiplied by the scale factor $(n_1n_2n_3)^{1/2}$. Now we state the conditions in the theorem explicitly. Here $c_0$ is some constant to be determined later.
\begin{enumerate}
	\item[{\crtcrossreflabel{(a)}[thmcond2]}]
\begin{equation*}
\textsf{D}_0^2=\sum_{k=1}^3(d_k^{(0)})^2\le\frac{c_1}{\kappa_0^8\cdot \bar{r}}
\end{equation*}
where $c_1=\min\left\{c_1^{(k)},k=0,\cdots,4\right\}$
$$c_1^{(0)}=\frac{c_0^2\gamma_\alpha^2\bar{r}}{12\beta_\alpha^2},\quad c_1^{(1)}=\frac{\bar{r}}{8(2+4\sqrt{\bar{r}})}$$
$$\quad c_1^{(2)}=\frac{\gamma_\alpha^3\bar{r}}{8\Big[\gamma_\alpha(1+c_0)+3(\beta_\alpha+1)\Big](\beta_\alpha+\gamma_\alpha)},\quad c_1^{(3)}=\frac{c_2^{2}\gamma_\alpha^2}{513^2\cdot 128}$$
$$c_1^{(4)}=\frac{\gamma_\alpha^2\sqrt{\bar{r}}}{16(\beta_\alpha+3\gamma_\alpha)(1+c_0)^2}$$
	\item[{\crtcrossreflabel{(b)}[thmcond1]}]
\begin{equation*}
\frac{\underline{\Lambda}(\bC^*)}{\textsf{Err}_{\br}} \ge  \max\left\{\frac{2}{c_0\gamma_\alpha},\sqrt{\frac{{256C_0^{\prime\prime} c_3}}{{\bar{r}}}},\frac{513\sqrt{128C_0^{\prime\prime}\bar{r}}}{c_2\gamma_\alpha}\kappa_0^2,\sqrt{\frac{8{C_0}\bar{r}}{c_1\gamma_\alpha}}\kappa_0^4\right\}
\end{equation*}
\end{enumerate}
and also 
$$\quad c_3=\frac{\gamma_\alpha^3}{384\beta_\alpha^2(\beta_\alpha+\gamma_\alpha)^2(1+c_0)^2}$$


\subsubsection{Error of the core tensor $C^{(t)}$}
We first focus on iteration $t=1$ and will finalize our proof by induction in the last part (for generality we keep the superscript $t$ here). Notice that $\E \nabla\ell_{n}(\bTheta^{*})=0$.
At iteration $t$, $\bC^{(t-1)}=\text{argmin}_{\fro{\bC}\le \textsf{Err}_{\br}}\ell_{n}(\bC\times_1U_{1}^{(t-1)}\times_2U_{2}^{(t-1)}\times_3U_{3}^{(t-1)})$. By Lemma \ref{lem:solveC} we have for any $\bC\in \R^{r_1\times r_2\times r_3}$ 
\begin{align}\label{optim_cond}
\langle\nabla_{\bC}\ell_{n}(\bTheta^{(t-1)}),\bC^{(t-1)}-\bC\rangle\le 0
\end{align}
where $\nabla_{\bC}\ell_{n}(\bTheta^{(t-1)})=\nabla\ell_{n}(\bTheta^{(t-1)})\cdot\llbracket U_1^{(t-1)\top},U_2^{(t-1)\top},U_3^{(t-1)\top}\rrbracket$. Since $\fro{\bC^{(t-1)}}\le \textsf{Err}_{\br}$ and $U_k^{(t-1)}$'s are incoherent, we can guarantee that $\|\widetilde{\bTheta}^{(t-1)}\|_\infty\le \alpha$ and $\|\bTheta^{(t-1)}\|_\infty\le \alpha$. Then by Lemma \ref{lem:lnproperty} we have 
\begin{align*}
\langle\nabla\ell_{n}(\bTheta^{(t-1)})-\nabla\ell_{n}(\widetilde{\bTheta}^{(t-1)}), \bTheta^{(t-1)}-\widetilde{\bTheta}^{(t-1)}\rangle\ge\gamma_\alpha\|\bTheta^{(t-1)}-\widetilde{\bTheta}^{(t-1)}\|_F^2=\gamma_\alpha \cdot (d_\textsf{C}^{(t-1)})^2 \numberthis \label{eqn0}
\end{align*}
On the other hand by \eqref{optim_cond},
\begin{align*}
\langle\nabla\ell_{n}(\bTheta^{(t)})&-\nabla\ell_{n}(\widetilde{\bTheta}^{(t-1)}), \bTheta^{(t-1)}-\widetilde{\bTheta}^{(t-1)}\rangle\le \langle-	\nabla\ell_{n}(\widetilde{\bTheta}^{(t-1)}), \bTheta^{(t-1)}-\widetilde{\bTheta}^{(t-1)}\rangle\\
&=\langle\nabla\ell_{n}(\bTheta^{*})-\nabla\ell_{n}(\widetilde{\bTheta}^{(t-1)}), \bTheta^{(t-1)}-\widetilde{\bTheta}^{(t-1)}\rangle+\langle\nabla\ell_{n}(\bTheta^{*}), \widetilde{\bTheta}^{(t-1)}-\bTheta^{(t-1)}\rangle
\end{align*}
The first term above can be bounded as follows:
\begin{align*}
&\langle\nabla\ell_{n}(\bTheta^{*})-\nabla\ell_{n}(\widetilde{\bTheta}^{(t-1)}), \bTheta^{(t-1)}-\widetilde{\bTheta}^{(t-1)}\rangle\le \|\nabla\ell_{n}(\bTheta^{*})-\nabla\ell_{n}(\widetilde{\bTheta})\|_F\|\bTheta^{(t-1)}-\widetilde{\bTheta}\|_F\\
&\le \beta_\alpha\|\bTheta^{*}-\widetilde{\bTheta}^{(t-1)}\|_F\|\bTheta^{(t-1)}-\widetilde{\bTheta}^{(t-1)}\|_F\le\beta_\alpha\|\bTheta^{*}-\widetilde{\bTheta}^{(t-1)}\|_F\cdot d_\textsf{C}^{(t-1)}\\
&\le\beta_\alpha\|\bC^*\cdot \llbracket U_{1}^*,U_{2}^*,U_{3}^*\rrbracket-\bC^*\cdot \llbracket U_{1}^{(t-1)}O_1^{(t-1)\top},U_{2}^{(t-1)}O_2^{(t-1)\top},U_{3}^{(t-1)}O_3^{(t-1)\top}\rrbracket\|_\text{F}\cdot d_\textsf{C}^{(t-1)}\\
&\le \beta_\alpha\oLambda(\bC^*)\cdot d_\textsf{C}^{(t-1)}\cdot \sum_{k=1}^3d_k^{(t-1)} \numberthis \label{eqn1}
\end{align*}
The second term can bounded as follows:
\begin{align*}
&\langle\nabla\ell_{n}(\bTheta^{*}), \widetilde{\bTheta}^{(t-1)}-\bTheta^{(t-1)}\rangle=\langle\nabla\ell_{n}(\bTheta^{*})-\nabla L(\bTheta^{*}), \widetilde{\bTheta}^{(t-1)}-\bTheta^{(t-1)}\rangle\\
&\le \|\left(\nabla\ell_{n}(\bTheta^{*})-\nabla L(\bTheta^{*})\right)\times_1(U_{1}^{(t-1)})^\top\times_2(U_{2}^{(t-1)})^\top\times_3(U_{3}^{(t-1)})^\top\|_F\cdot d_\textsf{C}^{(t-1)}\\
&=\sup_{\substack{\bC\in \mathbb{R}^{r_1\times r_2 \times r_3}, \|\bC\|_F\le 1}}\langle\left(\nabla\ell_{n}(\bTheta^{*})-\nabla L(\bTheta^{*})\right),\bC\times_1U_{1}^{(t-1)}\times_2U_{2}^{(t-1)}\times_3U_{3}^{(t-1)}\rangle  \cdot d_\textsf{C}^{(t-1)}\\
&\le \textsf{Err}_{\br} \cdot d_\textsf{C}^{(t-1)} \numberthis \label{eqn2}
\end{align*}
Combining \eqref{eqn0}, \eqref{eqn1} and \eqref{eqn2} we have 
\begin{align*}
d_\textsf{C}^{(t-1)}\le \frac{\beta_\alpha}{\gamma_\alpha}\cdot \oLambda(\bC^*)\cdot \sum_{k=1}^3d_k^{(t-1)}+\frac{\textsf{Err}_{\br}}{\gamma_\alpha}  \numberthis \label{Cerror}
\end{align*}
which hold for $t=1$. 
\subsubsection{Error of $\widetilde{U}_k^{(t)}$ (Gradient descent step)} By \eqref{Cerror} and the condition \ref{thmcond2} and \ref{thmcond1}, we have
\begin{align*}
\sum_{k=1}^3d_k^{(t-1)}\le\sqrt{3\sum_{k=1}^3(d_k^{(t-1)})^2}\le \frac{c_0\gamma_\alpha}{2\beta_\alpha\kappa_0^2}, \quad \underline{\Lambda}(\bC^*)\ge \frac{2\textsf{Err}_{\br}}{\gamma_\alpha c_0} \numberthis \label{dklambdacond}
\end{align*}
and the following inequality which will be used throughout this section:
\begin{align*}
\oLambda(\bC^{(t)})\le(1+c_0)\oLambda(\bC^*), \quad \underline{\Lambda}(\bC^{(t)})\ge(1+c_0)\underline{\Lambda}(\bC^*)  \numberthis \label{CtCstarrel}
\end{align*}
WLOG we consider the case $k=1$. Note that at iteration $t$, we have 
\begin{align*}
\|\widetilde{U}_1^{(t)}-{U}_1^*O_1^{(t-1)}\|_F^2&=\|{U}_1^{(t)}-\eta \cM_1(\nabla\ell_{n}(\bTheta^{(t-1)}))(U_3^{(t-1)}\otimes U_2^{(t-1)})\cM_1^T(\bC^{(t-1)})-{U}_k^*O_1^{(t-1)}\|_F^2\\
&=(d_1^{(t)})^2+\eta^2\|\cM_1(\nabla\ell_{n}(\bTheta^{(t-1)}))(U_3^{(t-1)}\otimes U_2^{(t-1)})\cM_1^T(\bC^{(t-1)})\|_F^2\\
&-2\eta\langle{U}_1^{(t)}-{U}_k^*O_1^{(t-1)},\cM_1(\nabla\ell_{n}(\bTheta^{(t-1)}))(U_3^{(t-1)}\otimes U_2^{(t-1)})\cM_1^T(\bC^{(t-1)})\rangle
\end{align*}
Then we bound the last two terms separately. Note that
\begin{align*}
&\eta^2\|\cM_1(\nabla\ell_{n}(\bTheta^{(t-1)}))(U_3^{(t-1)}\otimes U_2^{(t-1)})\cM_1^T(\bC^{(t-1)})\|_F^2\\
&\le 2\eta^2 \|\cM_1(\nabla\ell_{n}(\bTheta^{*}))(U_3^{(t-1)}\otimes U_2^{(t-1)})\cM_1^T(\bC^{(t-1)})\|_F^2\\
&+2\eta^2 \|\cM_1(\nabla\ell_{n}(\bTheta^{(t-1)})-\nabla\ell_{n}(\bTheta^{*}))(U_3^{(t-1)}\otimes U_2^{(t-1)})\cM_1^T(\bC^{(t-1)})\|_F^2\\
&\le2\eta^2\|\cM_1(\bC^{(t-1)})\|^2\textsf{Err}_{\br}^2+2\eta^2 \beta_\alpha^2\|\bTheta^{(t-1)}-\bTheta^{*}\|_F^2\|\cM_1(\bC^{(t-1)})\|^2\\
&\le{2\eta^2\oLambda^2(\bC^{(t-1)})\left[\textsf{Err}_{\br}^2+ \beta_\alpha^2\left(d_\textsf{C}^{(t-1)}+\oLambda(\bC^*)\sum_{k=1}^3d_k^{(t-1)}\right)^2\right]}\\
&\le\frac{8\beta_\alpha^2(\beta_\alpha+\gamma_\alpha)^2(1+c_0)^2\oLambda^4(\bC^*)}{\gamma_\alpha^2}\eta^2\cdot \sum_{k=1}^3(d_k^{(t-1)})^2+\frac{(8\beta_\alpha^2+2\gamma_\alpha^2)(1+c_0)^2\oLambda^2(\bC^*)}{\gamma_\alpha^2}\eta^2\textsf{Err}_{\br}^2 =:A
\end{align*}
Similarly we could derive the bound for $k=2$ and $k=3$. Therefore, we can write 
\begin{align*}
\sum_{k=1}^3\|\widetilde{U}_k^{(t)}-{U}_k^*O_k^{(t-1)}\|_F^2&\le \sum_{k=1}^3(d_{k}^{(t-1)})^2+3A-2\eta\langle\nabla\ell_{n}(\bTheta^{(t-1)}),B_0+B_1+B_2+B_3+B_4\rangle \numberthis \label{UB01234}
\end{align*}
where $$B_0=\bC^{(t-1)}\cdot\llbracket U_1^{(t-1)},U_2^{(t-1)},U_3^{(t-1)}\rrbracket-\bC^{*}\cdot\llbracket U_1^{*},U_2^{*},U_3^{*}\rrbracket$$
$$B_1=\bC^{(t-1)}\cdot\llbracket U_1^{(t-1)}-U_1^*O_1^{(t-1)},U_2^{(t-1)}-U_2^*O_2^{(t-1)},U_3^{(t-1)}\rrbracket$$
$$B_2=\bC^{(t-1)}\cdot\llbracket U_1^{(t-1)}-U_1^*O_1^{(t-1)},U_2^{(t-1)},U_3^{(t-1)}-U_3^*O_3^{(t-1)}\rrbracket$$
$$B_3=\bC^{(t-1)}\cdot\llbracket U_1^{(t-1)},U_2^{(t-1)}-U_2^*O_2^{(t-1)},U_3^{(t-1)}-U_3^*O_3^{(t-1)}\rrbracket$$
$$B_4=\left[\bC^*-\bC^{(t-1)}\cdot\llbracket O_1^{(t-1)},O_2^{(t-1)},O_3^{(t-1)}\rrbracket\right]\cdot\llbracket U_1^*,U_2^*,U_2^*\rrbracket$$
We are going to bound $\langle\nabla\ell_{n}(\bTheta^{(t-1)}),B_0+B_1+B_2+B_3+B_4\rangle$ separately. First note that
\begin{align*}
\langle\nabla\ell_{n}(\bTheta^{(t-1)}), B_0\rangle&=\langle\nabla\ell_{n}(\bTheta^{(t-1)})-\nabla\ell_{n}(\bTheta^{*}), \bTheta^{(t-1)}-\bTheta^*\rangle+\langle\nabla\ell_{n}(\bTheta^{*})-\nabla L(\bTheta^{*}), \bTheta^{(t-1)}-\bTheta^*\rangle\\
&\ge \gamma_\alpha\|\bTheta^{(t-1)}-\bTheta^*\|_F^2-|\langle\nabla\ell_{n}(\bTheta^{*})-\nabla L(\bTheta^{*}), \bTheta^{(t-1)}-\bTheta^*\rangle|\numberthis \label{eqn3}
\end{align*}
where we need to expand the $\|\bTheta^{(t-1)}-\bTheta^*\|_F^2$ to find a lower bound of it. Denote $\Delta_k:=U_k^{(t-1)}O_k^{(t-1)T}-U_k^*$ for $k=1,2,3$ and $\Delta_C:=\bC^{(t-1)}-\bC^*\cdot\llbracket O_1^{(t-1)\top},O_2^{(t-1)\top},O_3^{(t-1)\top}\rrbracket $ then we have
\begin{align*}
&\|\bTheta^{(t-1)}-\bTheta^*\|_F^2=\|\bTheta^{(t-1)}-\widetilde{\bTheta}^{(t-1)}+\widetilde{\bTheta}^{(t-1)}-\bTheta^*\|_F^2\\
&=\big\|\bTheta^{(t-1)}-\widetilde{\bTheta}^{(t-1)}+\underbrace{\bC^*\cdot\llbracket {\Delta_1},U_2^{(t-1)}O_2^{(t-1)\top},U_3^{(t-1)}O_3^{(t-1)\top}\rrbracket }_{B_{01}}\\
&+\underbrace{\bC^*\cdot\llbracket U_1^*,{\Delta_2},U_3^{(t-1)}O_3^{(t-1)\top}\rrbracket }_{B_{02}}+\underbrace{\bC^*\cdot\llbracket U_1^*,U_2^*,{\Delta_3}\rrbracket }_{B_{03}}\big\|_F^2\\
&=\big\|\bTheta^{(t-1)}-\widetilde{\bTheta}^{(t-1)}\big\|_F^2+\sum_{k=1}^3\|B_{0k}\|_F^2+2\sum_{k=1}^3\langle\bTheta^{(t-1)}-\widetilde{\bTheta}^{(t-1)},B_{0k}\rangle+2\sum_{k,l\in [3],k\ne l}\langle B_{0k}, B_{0l}\rangle \numberthis \label{eqn4}
\end{align*}
Note that the first term on the RHS of \eqref{eqn4} is nothing but $(d_\textsf{C}^{(t-1)})^2$ and  moreover,
\begin{align*}
\|B_{01}\|_F^2=\|\Delta_1\mathcal{M}(\bC^*)(U_3^{(t-1)}O_3^{(t-1)}\otimes U_2^{(t-1)}O_2^{(t-1)})^\top\|_F^2=\|\Delta_1\mathcal{M}(\bC^*)\|_F^2\ge \underline{\Lambda}^2(\bC^*)(d_1^{(t-1)})^2
\end{align*}
Similar bounds holds for $\|B_{02}\|_F^2$ and $\|B_{03}\|_F^2$, then we have the lower bound for the second term
\begin{align*}
\sum_{k=1}^3\|B_{0k}\|_F^2\ge \underline{\Lambda}^2(\bC^*)\sum_{k=1}^3(d_k^{(t-1)})^2 \numberthis \label{eqn5}
\end{align*}
For the third term of \eqref{eqn4}, note that 
\begin{align*}
&|\langle\bTheta^{(t-1)}-\widetilde{\bTheta}^{(t-1)},B_{01}\rangle|\\
&=|\langle\mathcal{M}_1(\Delta_C)(U_3^{(t-1)}\otimes U_2^{(t-1)}),U_1^{(t-1)T}\Delta_1\mathcal{M}_1(\bC^*)(U_3^{(t-1)}O_3^{(t-1)}\otimes U_2^{(t-1)}O_2^{(t-1)})\rangle|\\
&\le {\oLambda}(\bC^*)d_\textsf{C}^{(t-1)}\|U_1^{(t-1)T}\Delta_1\|_F\le {\oLambda}(\bC^*)d_\textsf{C}^{(t-1)}(d_1^{(t-1)})^2
\end{align*}
where the last inequality can be found in, e.g., \cite{xia2017polynomial}. Therefore, we have
\begin{align*}
2\sum_{k=1}^3|\langle\bTheta^{(t-1)}-\widetilde{\bTheta}^{(t-1)},B_{0k}\rangle|\le 2{\oLambda}(\bC^*)d_\textsf{C}^{(t-1)}\sum_{k=1}^3(d_k^{(t-1)})^2 \numberthis \label{eqn6}
\end{align*}
For the last term of \eqref{eqn4}, note that 
\begin{align*}
|\langle B_{01}, B_{02}\rangle|&=|\langle\Delta_1\mathcal{M}_1(\bC^*)(U_3^{(t-1)}O_3^{(t-1)T}\otimes U_2^{(t-1)}O_2^{(t-1)T}), U_1^*\mathcal{M}_1(\bC^*)(U_3^{(t-1)}O_3^{(t-1)T}\otimes \Delta_2)\rangle|\\
&=|\langle \mathcal{M}_1(\bC^*)(I_{r_3}\otimes U_2^{(t-1)T}O_2^{(t-1)}\Delta_2),\Delta_1^T U_1^*\mathcal{M}_1(\bC^*)\rangle|\\
&\le \sqrt{r_3}\bar{\Lambda}^2(\bC^*)(d_1^{(t-1)})^2(d_2^{(t-1)})^2
\end{align*}
Therefore we have
\begin{align*}
2\sum_{k,l\in [3],k\ne l}|\langle B_{0k}, B_{0l}\rangle|&\le 2{\oLambda}^2(\bC^*)\sqrt{\bar{r}}\left((d_1^{(t-1)}d_2^{(t-1)})^2+(d_1^{(t-1)}d_3^{(t-1)})^2+(d_2^{(t-1)}d_3^{(t-1)})^2\right) \\
&\le 2{\oLambda}^2(\bC^*)\sqrt{\bar{r}} \sum_{k=1}^3(d_k^{(t-1)})^4\numberthis \label{eqn7}
\end{align*}
In addition, similar to \eqref{eqn2}, the second term of \eqref{eqn3} can be bounded as
\begin{align*}
|\langle\nabla\ell_{n}(\bTheta^{*})-\nabla L(\bTheta^{*}), \bTheta^{(t-1)}-\bTheta^*\rangle|\le \left(d_\textsf{C}^{(t-1)}+{\oLambda}(\bC^*)\sum_{k=1}^3 d_k^{(t-1)}\right)\cdot \textsf{Err}_{\br} \numberthis \label{eqn8}
\end{align*}
Combining \eqref{eqn3} to \eqref{eqn8}, we have
\begin{align*}
\langle\nabla\ell_{n}(\bTheta^{(t-1)}), B_0\rangle&\ge \gamma_\alpha\Big[(d_\textsf{C}^{(t-1)})^2+\underline{\Lambda}^2(\bC^*)\sum_{k=1}^3(d_k^{(t-1)})^2-2{\oLambda}(\bC^*)d_\textsf{C}^{(t-1)}\sum_{k=1}^3(d_k^{(t-1)})^2\\
&-2{\oLambda}^2(\bC^*)\sqrt{\bar{r}} \sum_{k=1}^3(d_k^{(t-1)})^4\Big]-\left(d_\textsf{C}^{(t-1)}+{\oLambda}(\bC^*)\sum_{k=1}^3 d_k^{(t-1)}\right)\cdot \textsf{Err}_{\br} \numberthis \label{B0}
\end{align*}
Next, notice that 
\begin{align*}
|\langle\nabla\ell_{n}(\bTheta^{(t-1)}), \sum_{k=1}^3B_k\rangle|&\le |\langle\nabla\ell_{n}(\bTheta^{(t-1)})-\nabla\ell_{n}(\bTheta^{*}), \sum_{k=1}^3B_k\rangle|+|\langle\nabla\ell_{n}(\bTheta^{*}), \sum_{k=1}^3B_k\rangle|\\
&\le \beta_\alpha\|\bTheta^{(t-1)}-\bTheta^{*}\|_F\big\|\sum_{k=1}^3B_k\big\|_F+|\langle\nabla\ell_{n}(\bTheta^{*}), \sum_{k=1}^3B_k\rangle|\\
&\le \beta_\alpha\left(d_\textsf{C}^{(t-1)}+{\oLambda}(\bC^*)\sum_{k=1}^3 d_k^{(t-1)}\right)\sum_{k=1}^3\|B_k\|_F+|\langle\nabla\ell_{n}(\bTheta^{*}), \sum_{k=1}^3B_k\rangle|
\end{align*}
By the definition of $\{B_k\}_{k=1}^3$, we have
\begin{align*}
\sum_{k=1}^3\|B_k\|_F&\le \oLambda(\bC^{(t-1)})\left(d_1^{(t-1)}d_2^{(t-1)}+d_1^{(t-1)}d_3^{(t-1)}+d_2^{(t-1)}d_2^{(t-1)}\right)\le \oLambda(\bC^{(t-1)})\sum_{k=1}^3(d_k^{(t-1)})^2
\end{align*}
And a similar argument to \eqref{eqn2} gives that
\begin{align*}
|\langle\nabla\ell_{n}(\bTheta^{*}), \sum_{k=1}^3B_k\rangle|&\le \oLambda(\bC^{(t-1)})\cdot \textsf{Err}_{\br} \cdot \left(d_1^{(t-1)}d_2^{(t-1)}+d_1^{(t-1)}d_3^{(t-1)}+d_2^{(t-1)}d_2^{(t-1)}\right)\le \oLambda(\bC^{(t-1)})\cdot \textsf{Err}_{\br} \cdot \sum_{k=1}^3(d_k^{(t-1)})^2
\end{align*}
Therefore, we have
\begin{align*}
|\langle\nabla\ell_{n}(\bTheta^{(t-1)}), \sum_{k=1}^3B_k\rangle|&\le \oLambda(\bC^{(t-1)})\left[\beta_\alpha d_\textsf{C}^{(t-1)}+\beta_\alpha{\oLambda}(\bC^*)\sum_{k=1}^3 d_k^{(t-1)}+\textsf{Err}_{\br}\right]\cdot \sum_{k=1}^3(d_k^{(t-1)})^2 \numberthis \label{B123}
\end{align*}
It remains to upper bound $-\langle\nabla\ell_{n}(\bTheta^{(t-1)}),B_4\rangle$. Observe that 
\begin{align*}
&-\langle\nabla\ell_{n}(\bTheta^{(t-1)}),B_4\rangle= \langle\nabla\ell_{n}(\bTheta^{(t-1)}),\left[\bC^{(t-1)}\cdot\llbracket O_1^{(t-1)},O_2^{(t-1)},O_3^{(t-1)}\rrbracket-\bC^*\right]\cdot\llbracket U_1^*,U_2^*,U_2^*\rrbracket\rangle\\
&= \langle\nabla\ell_{n}(\bTheta^{(t-1)}),\underbrace{\Delta_C\cdot \llbracket U_1^*O_1^{(t-1)},U_2^*O_2^{(t-1)},U_3^*O_3^{(t-1)}\rrbracket -\Delta_C\cdot \llbracket U_1^{(t-1)},U_2^{(t-1)},U_3^{(t-1)}\rrbracket }_{B_{40}}\rangle\\
&+\langle\nabla\ell_{n}(\bTheta^{(t-1)}),\Delta_C\cdot \llbracket U_1^{(t-1)},U_2^{(t-1)},U_3^{(t-1)}\rrbracket \rangle\\
&\le \langle\nabla\ell_{n}(\bTheta^{(t-1)}),B_{40}\rangle=\langle\nabla\ell_{n}(\bTheta^{(t-1)})-\nabla\ell_{n}(\bTheta^{*}),{B_{40}}\rangle+\langle\nabla\ell_{n}(\bTheta^{*}),{B_{40}}\rangle \\
&\le \beta_\alpha\|\bTheta^{(t-1)}-\bTheta^{*}\|_F\cdot d_\textsf{C}^{(t-1)}\sum_{k=1}^3 d_k^{(t-1)}+\textsf{Err}_{\br}\cdot d_\textsf{C}^{(t-1)}\sum_{k=1}^3d_k^{(t-1)}\\
&\le \left[\beta_\alpha d_\textsf{C}^{(t-1)}+\beta_\alpha{\oLambda}(\bC^*)\sum_{k=1}^3 d_k^{(t-1)}+\textsf{Err}_{\br}\right]d_\textsf{C}^{(t-1)}\cdot\sum_{k=1}^3d_k^{(t-1)} \numberthis \label{B4}
\end{align*}
where the first inequality is due to the optimality condition \eqref{optim_cond}, and the third inequality follows from the following decomposition
\begin{align*}
B_{40}=&\Delta_C\cdot\Big[\llbracket -\Delta_1O_1^{(t-1)},U_2^*O_2^{(t-1)},U_3^*O_3^{(t-1)}\rrbracket +\llbracket U_1^{(t-1)},-\Delta_2O_2^{(t-1)},U_3^*O_3^{(t-1)}\rrbracket \\
&+\llbracket U_1^{(t-1)},U_2^{(t-1)},-\Delta_3O_3^{(t-1)}\rrbracket \Big]\\
\end{align*}
Combining \eqref{Cerror}, \eqref{B0}, \eqref{B123} and \eqref{B4} we have
\begin{align*}
&-2\eta\langle\nabla\ell_{n}(\bTheta^{(t-1)}),B_0+B_1+B_2+B_3+B_4\rangle\le -2\eta\gamma_\alpha\Big[(d_\textsf{C}^{(t-1)})^2+\underline{\Lambda}^2(\bC^*)\sum_{k=1}^3(d_k^{(t-1)})^2\\
&-2{\oLambda}(\bC^*)d_\textsf{C}^{(t-1)}\sum_{k=1}^3(d_k^{(t-1)})^2-2{\oLambda}^2(\bC^*)\sqrt{\bar{r}} \sum_{k=1}^3(d_k^{(t-1)})^4\Big]+2\eta\left(d_\textsf{C}^{(t-1)}+{\oLambda}(\bC^*)\sum_{k=1}^3 d_k^{(t-1)}\right)\cdot \textsf{Err}_{\br}\\
&+2\eta\oLambda(\bC^{(t-1)})\left[\beta_\alpha d_\textsf{C}^{(t-1)}+\beta_\alpha{\oLambda}(\bC^*)\sum_{k=1}^3 d_k^{(t-1)}+\textsf{Err}_{\br}\right]\cdot \sum_{k=1}^3(d_k^{(t-1)})^2\\
&+2\eta\left[\beta_\alpha d_\textsf{C}^{(t-1)}+\beta_\alpha{\oLambda}(\bC^*)\sum_{k=1}^3 d_k^{(t-1)}+\textsf{Err}_{\br}\right]d_\textsf{C}^{(t-1)}\cdot\sum_{k=1}^3d_k^{(t-1)}\\
&\le -2\eta\gamma_\alpha\underline{\Lambda}^2(\bC^*)\sum_{k=1}^3(d_k^{(t-1)})^2+(2+4\sqrt{\bar{r}})\gamma_\alpha{\oLambda}^2(\bC^*)\eta\cdot\sum_{k=1}^3(d_k^{(t-1)})^2\cdot \sum_{k=1}^3(d_k^{(t-1)})^2\\
&+ \frac{2(\beta_\alpha+\gamma_\alpha){\oLambda}(\bC^*)}{\gamma_\alpha}\eta\textsf{Err}_{\br}\cdot\sum_{k=1}^3 d_k^{(t-1)}+\frac{2}{\gamma_\alpha}\eta\textsf{Err}_{\br}^2+\frac{2\beta_\alpha(\beta_\alpha+\gamma_\alpha)(1+c_0){\oLambda}^2(\bC^*)}{\gamma_\alpha}\eta\cdot \sum_{k=1}^3 d_k^{(t-1)}\cdot\sum_{k=1}^3(d_k^{(t-1)})^2\\
&+\frac{2(\beta_\alpha+\gamma_\alpha)(1+c_0)\oLambda(\bC^*)}{\gamma_\alpha}\eta\textsf{Err}_{\br}\cdot\sum_{k=1}^3(d_k^{(t-1)})^2 + \frac{6\beta_\alpha(\beta_\alpha+\gamma_\alpha){\oLambda}^2(\bC^*)}{\gamma_\alpha^2}\eta\cdot\sum_{k=1}^3 d_k^{(t-1)}\cdot \sum_{k=1}^3 (d_k^{(t-1)})^2\\
&+\frac{6\beta_\alpha(\beta_\alpha+\gamma_\alpha){\oLambda}(\bC^*)}{\gamma_\alpha^2}\eta\textsf{Err}_{\br}\cdot\sum_{k=1}^3 (d_k^{(t-1)})^2+\frac{6(\beta_\alpha+\gamma_\alpha){\oLambda}(\bC^*)}{\gamma_\alpha^2}\eta\textsf{Err}_{\br}\cdot \sum_{k=1}^3 (d_k^{(t-1)})^2+\frac{2(\beta_\alpha+\gamma_\alpha)}{\gamma_\alpha^2}\eta\textsf{Err}_{\br}^2\cdot \sum_{k=1}^3 d_k^{(t-1)} \\
&\le -2\eta\gamma_\alpha\underline{\Lambda}^2(\bC^*)\sum_{k=1}^3(d_k^{(t-1)})^2+\frac{\gamma_\alpha{\oLambda}^2(\bC^*)}{\kappa_0^2}\eta\sum_{k=1}^3 (d_k^{(t-1)})^2\\
&+\Big[(2+4\sqrt{\bar{r}})\gamma_\alpha+\frac{\Big[\gamma_\alpha(1+c_0)+3(\beta_\alpha+1)\Big](\beta_\alpha+\gamma_\alpha)}{\gamma_\alpha^2}\Big]{\oLambda}^2(\bC^*)\eta\cdot\sum_{k=1}^3(d_k^{(t-1)})^2\cdot\sum_{k=1}^3(d_k^{(t-1)})^2\\
&+\frac{\Big[2\gamma_\alpha(1+c_0)+6\Big]\beta_\alpha(\beta_\alpha+\gamma_\alpha){\oLambda}^2(\bC^*)}{\gamma_\alpha^2}\eta\cdot \sum_{k=1}^3 d_k^{(t-1)}\cdot\sum_{k=1}^3(d_k^{(t-1)})^2\\
&+\Big[\frac{\Big[\gamma_\alpha(1+c_0)+3(\beta_\alpha+1)\Big](\beta_\alpha+\gamma_\alpha)+2\gamma_\alpha}{\gamma_\alpha^2}+\frac{3(\beta_\alpha+\gamma_\alpha)^2\kappa_0^2}{\gamma_\alpha^3}+\frac{2(\beta_\alpha+\gamma_\alpha)}{\gamma_\alpha^2}\cdot \sum_{k=1}^3 d_k^{(t-1)}\Big]\eta\textsf{Err}_{\br}^2\\ \numberthis \label{B01234}
\end{align*}
By \ref{thmcond2} we have
\begin{align*}
\sum_{k=1}^3 d_k^{(t-1)}\le \sqrt{3\sum_{k=1}^3 (d_k^{(t-1)})^2}\le \frac{\gamma_\alpha^3}{8\Big[2\gamma_\alpha(1+c_0)+6\Big]\beta_\alpha(\beta_\alpha+\gamma_\alpha)\kappa_0^2} \numberthis \label{dk1cond}
\end{align*}
\begin{align*}
\sum_{k=1}^3 (d_k^{(t-1)})^2\le \frac{1}{\kappa_0^4}\min\left\{\frac{1}{8(2+4\sqrt{\bar{r}})},\frac{\gamma_\alpha^3}{8\Big[\gamma_\alpha(1+c_0)+3(\beta_\alpha+1)\Big](\beta_\alpha+\gamma_\alpha)}\right\} \numberthis \label{dk2cond}
\end{align*}
and also
\begin{align*}
\eta\le \frac{\gamma_\alpha^3}{192\beta_\alpha^2(\beta_\alpha+\gamma_\alpha)^2(1+c_0)^2\kappa_0^4}\cdot\frac{1}{\underline{\Lambda}^2(\bC^*)}\numberthis \label{etacond}
\end{align*}
Using the relation ${\gamma_\alpha{\oLambda}^2(\bC^*)}/{\kappa_0^2}=\gamma_\alpha\underline{\Lambda}^2(\bC^*)$, \eqref{dk1cond} and \eqref{dk2cond} , the concentration for $\widetilde{U}^{(t)}_k$'s becomes
\begin{align*}
\sum_{k=1}^3\|\widetilde{U}_k^{(t)}-{U}_k^*O_k^{(t-1)}\|_F^2\le \left(1-\frac{1}{2}\eta\gamma_\alpha\underline{\Lambda}^2(\bC^*)\right)\sum_{k=1}^3(d_k^{(t-1)})^2+C_0^\prime\eta\textsf{Err}_{\br}^2 \numberthis \label{Utildeerror}
\end{align*}
where
\begin{align*}
C_0^\prime:&=\frac{\Big[\gamma_\alpha(1+c_0)+3(\beta_\alpha+1)\Big](\beta_\alpha+\gamma_\alpha)+2\gamma_\alpha}{\gamma_\alpha^2}+\frac{3(\beta_\alpha+\gamma_\alpha)^2\kappa_0^2}{\gamma_\alpha^3}\\
&+\frac{\gamma_\alpha}{3\Big[2\gamma_\alpha(1+c_0)+6\Big]\beta_\alpha\kappa_0^2}+\frac{(4\beta_\alpha^2+\gamma_\alpha^2)\gamma_\alpha}{96\beta_\alpha^2(\beta_\alpha+\gamma_\alpha)^2\kappa_0^2}
\end{align*}

\subsubsection{Error of $\check{U}_k^{(t)}$ (SVD step)} For $k\in[3]$, denote the compact SVD of $\widetilde U_k^{(t)}$ as $\check U_k^{(t)}\check \Sigma_k^{(t)}\check R_k^{(t)\top}$, then 
$$\check U_k^{(t)}\check \Sigma_k^{(t)}\check R_k^{(t)\top}=\widetilde{U}_k^{(t)}=U_k^*O_k^{(t-1)}+(\widetilde{U}_k^{(t)}-U_k^*O_k^{(t-1)})$$
Hence
\begin{align*}
\check{U}_k^{(t)}-U_k^*O_k^{(t)}&=U_k^*O_k^{(t-1)}\check R_k^{(t)}(\check \Sigma_k^{(t)})^{-1}-U_k^*O_k^{(t)}+(\widetilde{U}_k^{(t)}-U_k^*O_k^{(t-1)})\check R_k^{(t)}(\check \Sigma_k^{(t)})^{-1}\\
&=U_k^*O_k^{(t)}\Big[(\check \Sigma_k^{(t)})^{-1}-I\Big]+(\widetilde{U}_k^{(t)}-U_k^*O_k^{(t-1)})\check R_k^{(t)}(\check \Sigma_k^{(t)})^{-1}
\end{align*}
Therefore, we have
 \begin{align*}
 \|\check {U}_k^{(t)}-U_k^*O_k^{(t)}\|_F^2&\le \|\widetilde {U}_k^{(t)}-U_k^*O_k^{(t-1)}\|_F^2\|(\check \Sigma_k^{(t)})^{-1}\|+\|U_k^*O_k^{(t)}\Big[(\check \Sigma_k^{(t)})^{-1}-I\Big]\|_F^2\\
 &+2\langle U_k^*O_k^{(t)}\Big[(\check \Sigma_k^{(t)})^{-1}-I\Big],(\widetilde {U}_k^{(t)}-U_k^*O_k^{(t-1)})\check R_k^{(t)}(\check \Sigma_k^{(t)})^{-1}\rangle \numberthis \label{3terms}
 \end{align*}
 We are going to bound each term on the RHS of \eqref{3terms} seperately. Note that
 $$\sigma_{\min}(\widetilde{U}_k^{(t)})\ge\sigma_{\min}(U_k^*O_k^{(t-1)})-\|\widetilde{U}_k^{(t)}-U_k^*O_k^{(t-1)}\|\ge 1-\frac{c_0}{2\kappa_0^4}$$
where we used the condition  \ref{thmcond1}. Thus we have
\begin{align*}
\|(\check \Sigma_k^{(t)})^{-1}\|=(\sigma_{\min}(\widetilde U_k^{(t)}))^{-1}\le \frac{1}{1-\frac{c_0}{2\kappa_0^2}}\le 1+\frac{c_0}{\kappa_0^4} \numberthis \label{term1}
\end{align*}
To bound the second term of \eqref{3terms}, let $k_1,k_2\in[3],k_1>k_2$, observe that
\begin{align*}
\widetilde{U}_k^{(t)}&=\underbrace{U_k^{(t-1)}-\eta\mathcal{M}_k(\nabla\ell(\bTheta^*))(U_{k_1}^{(t-1)}\otimes U_{k_2}^{(t-1)})\mathcal{M}_k^T(\bC^{(t-1)})}_{V_k}\\
&+\underbrace{\eta\mathcal{M}_k(\nabla\ell(\bTheta^*)-\nabla\ell(\bTheta^{(t-1)}))(U_{k_1}^{(t-1)}\otimes U_{k_2}^{(t-1)})\mathcal{M}_k^T(\bC^{(t-1)})}_{\Delta V_k}
\end{align*}
with
\begin{align*}
\|{\Delta V_k}\|&\le \|{\Delta V_k}\|_F\le \eta\oLambda(\bC^{(t-1)})\cdot\|\mathcal{M}_k(\nabla\ell(\bTheta^*)-\nabla\ell(\bTheta^{(t-1)}))(U_{k_1}^{(t-1)}\otimes U_{k_2}^{(t-1)})\|_F\\
&\le\eta\beta(1+c_0)\oLambda(\bC^*)\cdot\|\bTheta^*-\bTheta^{(t-1)}\|_F\le\eta\beta(1+c_0)\oLambda(\bC^*)\Big(d_C^{(t-1)}+\oLambda(\bC^*)\sum_{k=1}^3d_k^{(t-1)}\Big)\\
&\le\eta\beta(1+c_0)\oLambda(\bC^*)\Big(\frac{\beta+\gamma}{\gamma}\oLambda(\bC^*)\sum_{k=1}^3d_k^{(t-1)}+\frac{\textsf{Err}_{\br}}{\gamma}\Big)\\
\end{align*}
where we've used the relationship between $d_C^{(t-1)}$ and $\sum_{k=1}^3d_k^{(t-1)}$ in \eqref{Cerror}. Also note that
\begin{align*}
&V_k^TV_k=I_{r_k}-\eta\mathcal{M}_k(\bC^{(t-1)})(U_{k_1}^{(t-1)}\otimes U_{k_2}^{(t-1)})^T\mathcal{M}_k^T(\nabla\ell(\bTheta^*))U_k^{(t-1)}\\
&-\eta (U_k^{(t-1)})^T\mathcal{M}_k(\nabla\ell(\bTheta^*))(U_{k_1}^{(t-1)}\otimes U_{k_2}^{(t-1)})\mathcal{M}_k^T(\bC^{(t-1)})\\
&+\eta^2\mathcal{M}_k(\bC^{(t-1)})(U_{k_1}^{(t-1)}\otimes U_{k_2}^{(t-1)})^T\mathcal{M}_k^T(\nabla\ell(\bTheta^*))\mathcal{M}_k(\nabla\ell(\bTheta^*))(U_{k_1}^{(t-1)}\otimes U_{k_2}^{(t-1)})\mathcal{M}_k^T(\bC^{(t-1)})
\end{align*}
Hence we have
\begin{align*}
\|V_k^TV_k-I_{r_k}\|&\le\|V_k^TV_k-I_{r_k}\|_F\le \eta^2\|\mathcal{M}_k(\nabla\ell(\bTheta^*))(U_{k_1}^{(t-1)}\otimes U_{k_2}^{(t-1)})\mathcal{M}_k^T(\bC^{(t-1)})\|_F^2 \\
&+2\eta\|\mathcal{M}_k(\bC^{(t-1)})(U_{k_1}^{(t-1)}\otimes U_{k_2}^{(t-1)})^T\mathcal{M}_k^T(\nabla\ell(\bTheta^*))U_k^{(t-1)}\|_F\\
&\le \eta^2\Big[\sup_{U\in \mathbb{R}^{n_k\times r_k}, \|U\|_F\le 1}\langle\mathcal{M}_k(\nabla\ell(\bTheta^*)),U \mathcal{M}_k(\bC^{(t-1)})(U_{k_1}^{(t-1)}\otimes U_{k_2}^{(t-1)})^T\rangle\Big]^2\\
&+2\eta\sup_{U\in \mathbb{R}^{n_k\times r_k}, \|U\|_F\le 1}\langle\mathcal{M}_k(\nabla\ell(\bTheta^*)),U \mathcal{M}_k(\bC^{(t-1)})(U_{k_1}^{(t-1)}\otimes U_{k_2}^{(t-1)})^T\rangle\\
&\le (1+c_0)^2\oLambda^2(\bC^*)\eta^2\textsf{Err}_{\br}^2+2(1+c_0)\oLambda(\bC^*)\eta\textsf{Err}_{\br}\\
&\le 3(1+c_0)\oLambda(\bC^*)\eta\textsf{Err}_{\br}:=\omega
\end{align*}
where the last inequality is due to the assumption \eqref{dklambdacond} and \eqref{etacond}. It follows that $1-\omega\le \sigma_{r_k}(V_k)\le \sigma_1(V_k)\le 1+\omega$. Then we have
\begin{align*}
\|\widetilde{U}_k^{(t)}\|-1= \|V_k+\Delta V_k\|-1 \le \omega + \|\Delta V_k\|, \quad 1-\|\widetilde{U}_k^{(t)}\|\le 1-\sigma_{r_k}(V_k+\Delta V_k)\le \omega + \|\Delta V_k\|
\end{align*}
Similarly, we have
\begin{align*}
1-\sigma_{r_k}(\widetilde{U}_k^{(t)})\le \omega + \|\Delta V_k\|, \quad \sigma_{r_k}(\widetilde{U}_k^{(t)})-1\le \|\widetilde{U}_k^{(t)}\|-1\le \omega + \|\Delta V_k\|
\end{align*}
Therefore, we get
\begin{align*}
\|(\check \Sigma_k^{(t)})^{-1}-I\|=\max\Big\{\Big|\|\check \Sigma_k^{(t)}\|-1\Big|,\Big|1-\sigma_{r_k}(\check \Sigma_k^{(t)})\Big|\Big\}\le \omega + \|\Delta V_k\|
\end{align*}
Then the second term of \eqref{3terms} can bounded as
\begin{align*}
&\|U_k^*O_k^{(t)}\Big[(\check \Sigma_k^{(t)})^{-1}-I\Big]\|_F^2\le r_k\|(\Sigma_k^{(t)})^{-1}-I\|^2\le 2\bar{r}(\omega^2+\|\Delta V_k\|^2)\\
&\le \bar{r}\Big[18(1+c_0)^2\oLambda^2(\bC^*)\eta^2\textsf{Err}_{\br}^2+\frac{8\beta_\alpha^2(\beta_\alpha+{\gamma_\alpha})^2(1+c_0)^2\oLambda^4(\bC^*)}{{\gamma_\alpha}^2}\eta^2\cdot \sum_{k=1}^3(d_k^{(t-1)})^2\\
&+\frac{8\beta_\alpha^2(1+c_0)^2\oLambda^2(\bC^*)}{{\gamma_\alpha}^2}\eta^2\textsf{Err}_{\br}^2\Big]\\
&= \frac{8\bar{r}\beta_\alpha^2(\beta_\alpha+{\gamma_\alpha})^2(1+c_0)^2\oLambda^4(\bC^*)}{{\gamma_\alpha}^2}\eta^2\cdot \sum_{k=1}^3(d_k^{(t-1)})^2 + \frac{\bar{r}(8\beta_\alpha^2+18\gamma_\alpha^2)(1+c_0)^2\oLambda^2(\bC^*)}{{\gamma_\alpha}^2}\eta^2\textsf{Err}_{\br}^2
\end{align*}
It remains to bound the third term of \eqref{3terms}, notice that
\begin{align*}
&2\langle U_k^*O_k^{(t)}\Big[(\check \Sigma_k^{(t)})^{-1}-I\Big],(\widetilde{U}_k^{(t)}-U_k^*O_k^{(t-1)})\check R_k^{(t)}(\check \Sigma_k^{(t)})^{-1}\rangle\\
&= 2\langle(\check \Sigma_k^{(t)})^{-1}-I,(U_k^*O_k^{(t)})^\top (\widetilde{U}_k^{(t)}-U_k^*O_k^{(t-1)})\check R_k^{(t)}(\check \Sigma_k^{(t)})^{-1}\rangle\\
&\le 2\|(\Sigma_k^{(t)})^{-1}-I\|_F\|(U_k^*O_k^{(t)})^T(\widetilde{U}_k^{(t)}-U_k^*O_k^{(t-1)})\|_F\|(\check \Sigma_k^{(t)})^{-1}\|\\
&\le 2\sqrt{\bar{r}}(1+\frac{c_0}{\kappa_0^2})\Big[\omega + \|\Delta V_k\|\Big]\|\widetilde{U}_k^{(t)}-{U}_k^*O_k^{(t-1)}\|_F^2\\
&\le 2\sqrt{\bar{r}}(1+\frac{c_0}{\kappa_0^2})\Big[\frac{\beta_\alpha(\beta_\alpha+\gamma_\alpha)(1+c_0)\oLambda^2(\bC^*)}{\gamma_\alpha}\eta\sum_{k=1}^3d_k^{(t-1)}+\frac{(\beta_\alpha+3\gamma_\alpha)(1+c_0)\oLambda(\bC^*)}{\gamma_\alpha}\eta\textsf{Err}_{\br}\Big]\|\widetilde{U}_k^{(t)}-{U}_k^*O_k^{(t-1)}\|_F^2
\end{align*}
 Therefore, combining bounds for three terms of \eqref{3terms} and the relationship \eqref{Utildeerror}, we have
 \begin{align*}
 &\sum_{k=1}^3\|\check {U}_k^{(t)}-U_k^*O_k^{(t)}\|_F^2\le \left(1+\frac{c_0}{\kappa_0^4}\right)^2\sum_{k=1}^3\|\widetilde{U}_k^{(t)}-U_k^*O_k^{(t-1)}\|_F^2\\
 &+\frac{24\bar{r}\beta_\alpha^2(\beta_\alpha+\gamma_\alpha)^2(1+c_0)^2\oLambda^4(\bC^*)}{\gamma_\alpha^2}\eta^2\cdot \sum_{k=1}^3(d_k^{(t-1)})^2 + \frac{6\bar{r}(4\beta_\alpha^2+9\gamma_\alpha^2)(1+c_0)^2\oLambda^2(\bC^*)}{\gamma_\alpha^2}\eta^2\textsf{Err}_{\br}^2\\
 &+2\sqrt{\bar{r}}(1+\frac{c_0}{\kappa_0^2})\Big[\frac{\beta_\alpha(\beta_\alpha+\gamma_\alpha)(1+c_0)\oLambda^2(\bC^*)}{\gamma_\alpha}\eta\sum_{k=1}^3d_k^{(t-1)}+\frac{(\beta_\alpha+3\gamma_\alpha)(1+c_0)\oLambda(\bC^*)}{\gamma_\alpha}\eta\textsf{Err}_{\br}\Big]\sum_{k=1}^3\|\widetilde{U}_k^{(t)}-{U}_k^*O_k^{(t-1)}\|_F^2\\
 &\le \left(1+\frac{c_0}{\kappa_0^4}\right)^2\Big[\left(1-\frac{1}{2}\eta\gamma_\alpha\underline{\Lambda}^2(\bC^*)\right)\sum_{k=1}^3(d_k^{(t-1)})^2+C_0^\prime\eta\textsf{Err}_{\br}^2\Big]\\
 &+\frac{24\bar{r}\beta_\alpha^2(\beta_\alpha+\gamma_\alpha)^2(1+c_0)^2\oLambda^4(\bC^*)}{\gamma_\alpha^2}\eta^2\cdot \sum_{k=1}^3(d_k^{(t-1)})^2 + \frac{6\bar{r}(4\beta_\alpha^2+9\gamma_\alpha^2)(1+c_0)^2\oLambda^2(\bC^*)}{\gamma_\alpha^2}\eta^2\textsf{Err}_{\br}^2\\
 &+\frac{2\sqrt{\bar{r}}\beta_\alpha(\beta_\alpha+\gamma_\alpha)(1+c_0)^2\oLambda^2(\bC^*)}{\gamma_\alpha}\eta\sum_{k=1}^3d_k^{(t-1)}\Big[\left(1-\frac{1}{2}\eta\gamma_\alpha\underline{\Lambda}^2(\bC^*)\right)\sum_{k=1}^3(d_k^{(t-1)})^2+C_0^\prime\eta\textsf{Err}_{\br}^2\Big]\\
 &+\frac{2\sqrt{\bar{r}}(\beta_\alpha+3\gamma_\alpha)(1+c_0)^2\oLambda(\bC^*)}{\gamma_\alpha}\eta\textsf{Err}_{\br}\Big[\left(1-\frac{1}{2}\eta\gamma_\alpha\underline{\Lambda}^2(\bC^*)\right)\sum_{k=1}^3(d_k^{(t-1)})^2+C_0^\prime\eta\textsf{Err}_{\br}^2\Big]
 \end{align*}
 \begin{align*}
 &\le \left(1+\frac{c_0}{\kappa_0^4}\right)^2\left(1-\frac{1}{2}\eta\gamma_\alpha\underline{\Lambda}^2(\bC^*)\right)\sum_{k=1}^3(d_k^{(t-1)})^2+\left(1+\frac{c_0}{\kappa_0^2}\right)^2C_0^\prime\eta\textsf{Err}_{\br}^2\\
 &+\frac{24\bar{r}\beta_\alpha^2(\beta_\alpha+\gamma_\alpha)^2(1+c_0)^2\oLambda^4(\bC^*)}{\gamma_\alpha^2}\eta^2\cdot \sum_{k=1}^3(d_k^{(t-1)})^2 \\
 &+ \frac{6\bar{r}(4\beta_\alpha^2+9\gamma_\alpha^2)(1+c_0)^2\oLambda^2(\bC^*)}{\gamma_\alpha^2}\eta^2\textsf{Err}_{\br}^2+\frac{2\sqrt{\bar{r}}\beta_\alpha(\beta_\alpha+\gamma_\alpha)(1+c_0)^2\oLambda^2(\bC^*)}{\gamma_\alpha}\eta\sum_{k=1}^3d_k^{(t-1)}\Big[\sum_{k=1}^3(d_k^{(t-1)})^2+C_0^\prime\eta\textsf{Err}_{\br}^2\Big]\\
 &+\frac{\sqrt{\bar{r}}(\beta_\alpha+3\gamma_\alpha)(1+c_0)^2}{\gamma_\alpha}\eta\Big[\oLambda^2(\bC^*)\Big(\sum_{k=1}^3(d_k^{(t-1)})^2\Big)^2+\textsf{Err}_{\br}^2\Big]\\
 &+\frac{2\sqrt{\bar{r}}(\beta_\alpha+3\gamma_\alpha)(1+c_0)^2\oLambda(\bC^*)}{\gamma_\alpha}C_0^\prime\eta^2\textsf{Err}_{\br}^3 \numberthis \label{finalrel}
 \end{align*}
 Now we can choose $c_0= \gamma_\alpha c_2/48$, it follows that
 $$\left(1+\frac{c_0}{\kappa_0^4}\right)^2\left(1-\frac{1}{2}\eta\gamma_\alpha\underline{\Lambda}^2(\bC^*)\right)\le 1-\frac{7}{16}\eta\gamma_\alpha\underline{\Lambda}^2(\bC^*)$$
 By condition \ref{thmcond2}, we have
 \begin{align*}
 \sum_{k=1}^3 d_k^{(t-1)}\le\sqrt{3\sum_{k=1}^3 (d_k^{(t-1)})^2}\le \frac{\gamma_\alpha^2}{32\sqrt{\bar{r}}\beta_\alpha(\beta_\alpha+\gamma_\alpha)(1+c_0)^2\kappa_0^2}
 \end{align*}
 \begin{align*}
 \sum_{k=1}^3 (d_k^{(t-1)})^2\le \frac{\gamma_\alpha^2}{16\sqrt{\bar{r}}(\beta_\alpha+3\gamma_\alpha)(1+c_0)^2\kappa_0^2}
 \end{align*}
 and also
 \begin{align*}
 \eta\le \frac{\gamma_\alpha^3}{384\bar{r}\beta_\alpha^2(\beta_\alpha+\gamma_\alpha)^2(1+c_0)^2\kappa_0^4}\cdot\frac{1}{\underline{\Lambda}^2(\bC^*)}
 \end{align*}
 Moreover, note the last term of \eqref{finalrel} is small due to the assumption \eqref{dklambdacond}, which implies 
 $$\oLambda(\bC^*)\eta^2\textsf{Err}_{\br}^3\le \frac{\gamma_\alpha^3}{384\bar{r}\beta_\alpha^2(\beta_\alpha+\gamma_\alpha)^2(1+c_0)^2\kappa_0^3}\cdot\frac{\textsf{Err}_{\br}}{\underline{\Lambda}(\bC^*)}\eta\textsf{Err}_{\br}^2\le\frac{\gamma_\alpha^4c_0}{768\bar{r}\beta_\alpha^2(\beta_\alpha+\gamma_\alpha)^2\kappa_0^3}$$
 Finally we have the contraction property
 \begin{align*}
 \sum_{k=1}^3\|\check {U}_k^{(t)}-U_k^*O_k^{(t)}\|_F^2\le\left(1-\frac{1}{4}\eta\gamma_\alpha\underline{\Lambda}^2(\bC^*)\right)\sum_{k=1}^3(d_k^{(t-1)})^2+C_0^{\prime\prime}\eta\textsf{Err}_{\br}^2 \numberthis \label{Ucheckerror}
 \end{align*}
 where
 $$C_0^{\prime\prime}=\frac{9\gamma_\alpha}{64(\beta_\alpha^2+\gamma_\alpha^2)\kappa_0^2}+\Big(\frac{c_0}{768}+\frac{1}{6144\kappa_0}\Big)\frac{\gamma_\alpha^4}{\beta_\alpha^2(\beta_\alpha+\gamma_\alpha)^2\kappa_0^3}C_0^\prime+\frac{\sqrt{\bar{r}}(\beta_\alpha+3\gamma_\alpha)(1+c_0)^2}{\gamma_\alpha}$$

\subsubsection{Error of $\widehat {U}_k^{(t)}$ (Regularization step)}
Denote $\widehat{U}_k^{(t)}:=\textsf{Reg}_{\delta_k}(\check{U}_k^{(t)})$ and let $\widehat{O}_k^{(t)}:=\argmin_O\|\widehat{U}_k^{(t)}-U_k^*O\|_F$. 
Since $U_k^*O_k^{(t-1)}$ is $\mu_0$-incoherent, we have 
\begin{align*}
\|\widehat{U}_k^{(t)}-U_k^*\widehat{O}_k^{(t)}\|_F\le \|\widehat{U}_k^{(t)}-U_k^*{O}_k^{(t-1)}\|_F\le \|\check{U}_k^{(t)}-U_k^*O_k^{(t-1)}\|_F \numberthis \label{Uhatbound}
\end{align*}
We write ${U}_k^{(t)}=U_k^*\widehat{O}_k^{(t)}+({U}_k^{(t)}-U_k^*\widehat{O}_k^{(t)})$, then by perturbation bound for singular subspaces (see \cite{xia2019normal, xia2019confidence}) we have
$$\|{U}_k^{(t)}{U}_k^{(t)\top}-U_k^*U_k^{*\top}\|_F^2\le 2\|\widehat{U}_k^{(t)}-U_k^*\widehat{O}_k^{(t)}\|_F^2-2\sum_{j\ge 3}\langle\Xi\Xi^\top,\mathcal{S}_j\rangle$$
where 
$$\Xi\Xi^T=\left(\begin{array}{cc}
U_k^* U_k^{*\top} & 0 \\
0 & I_{r_k}
\end{array}\right)$$
and $\mathcal{S}_j$ is $j$-th order perturbation term whose Frobenius norm can be bounded by  
$$\|\mathcal{S}_j\|_F\le \left(4\sqrt{2}\|\widehat{U}_k^{(t)}-U_k^*\widehat{O}_k^{(t)}\|_F\right)^j$$
It follows that 
$$|\sum_{j\ge 3}\langle\Xi\Xi^\top,\mathcal{S}_j\rangle|\le \|\Xi\Xi^\top\|_F\sum_{j\ge 3}\|\mathcal{S}_j\|_F\le \sqrt{2r_k}\left(4\sqrt{2}\|\widehat{U}_k^{(t)}-U_k^*\widehat{O}_k^{(t)}\|_F\right)^k\le 512\sqrt{\bar{r}}\|\widehat{U}_k^{(t)}-U_k^*\widehat{O}_k^{(t)}\|_F^3$$
The last inequality is due to \eqref{Uhatbound}, from which we have
$$\|\check{U}_k^{(t)}-U_k^*O_k^{(t-1)}\|_F\le \sqrt{\sum_{k=1}^3(d_k^{(t-1)})^2+C_0^{\prime\prime}\eta\textsf{Err}_{\br}^2}\le \frac{1}{8\sqrt{2}}$$
provided that $\underline{\Lambda}(\bC^*)\ge \sqrt{256C_0^{\prime\prime} c_3/\bar{r}}\cdot \textsf{Err}_{\br}$ and $\sum_{k=1}^3(d_k^{(t-1)})^2\le 1/256$. Using the explicit formula for geodesics on the Grassmann manifold (e.g., \cite{xia2017polynomial}\cite{edelman1998geometry}), we can derive the relation between projection distance and $d_k^{(t)}$
\begin{align*}
(d_k^{(t)})^2=\|{U}_k^{(t)}-U_k^*O_k^{(t)\top}\|_F^2\le\frac{1}{2}\|{U}_k^{(t)}{U}_k^{(t)\top}-U_k^*U_k^{*\top}\|_F^2+\frac{1}{4}\|{U}_k^{(t)}{U}_k^{(t)\top}-U_k^*U_k^{*\top}\|_F^4\numberthis \label{distrel}
\end{align*}
Then by \eqref{Ucheckerror} \eqref{Uhatbound} and \eqref{distrel} we have 
\begin{align*}
\sum_{k=1}^3(d_k^{(t)})^2&\le \sum_{k=1}^3\|\widehat{U}_k^{(t)}-U_k^*\widehat{O}_k^{(t)}\|_F^2+513\sqrt{\bar{r}}\sum_{k=1}^3\|\widehat{U}_k^{(t)}-U_k^*\widehat{O}_k^{(t)}\|_F^3\\
&\le \left(1-\frac{1}{4}\eta\gamma_\alpha\underline{\Lambda}^2(\bC^*)\right)\sum_{k=1}^3(d_k^{(t-1)})^2+C_0^{\prime\prime}\eta\textsf{Err}_{\br}^2+513\sqrt{\bar{r}}\left(\sum_{k=1}^3(d_k^{(t-1)})^2+C_0^{\prime\prime}\eta\textsf{Err}_{\br}^2\right)^{3/2}\\
&\le \left(1-\frac{1}{8}\eta\gamma_\alpha\underline{\Lambda}^2(\bC^*)\right)\sum_{k=1}^3(d_k^{(t-1)})^2+C_0\eta\textsf{Err}_{\br}^2 \numberthis \label{contractioninq}
\end{align*}
provided that 
$$\sum_{k=1}^3(d_{k}^{(t-1)})^2\le \frac{c_2^{2}\gamma_\alpha^2}{513^2\cdot128\bar{r}\kappa_0^8}$$
$$\underline{\Lambda}(\bC^*)\ge\frac{513\sqrt{128C_0^{\prime\prime}\cdot \bar{r}}}{c_2\gamma_\alpha}\kappa_0^2\cdot \textsf{Err}_{\br}$$
and $C_0=(1+{513\cdot c_3\gamma_\alpha\sqrt{\bar{r}}}/{\kappa_0^4})\cdot C_0^{\prime\prime}$.
\subsubsection{Induction step} Note that the above arguments hold only when $U_k^{(0)}$'s are $2\mu_0$-incoherent and $\sum_{k=1}^3(d_k^{(t)})^2$ satisify the condition \ref{thmcond2}. To deduce the contraction inequality, it suffices to verifty these conditions hold for $t\ge2$. Suppose for $t=t_0$ ($t_0\ge 1$) we have
$$\max_j\|e_j^\top U_k^{(t_0)}\|_2\le 2\mu_0\sqrt{\frac{r_k}{n_k}},\quad \sum_{k=1}^3(d_k^{(t_0-1)})^2\le \frac{c_1}{\kappa_0^8\bar{r}}$$ 
Then for $t=t_0+1$, the regularization step guarantees that $U_k^{(t_0+1)}$'s are $2\mu_0$-incoherent given that $\sum_{k=1}^3(d_k^{(t_0-1)})^2\le 1/256$ (see \cite{keshavan2010matrix}). By \eqref{contractioninq} we have
\begin{align*}
\quad \sum_{k=1}^3(d_k^{(t_0)})^2&\le\left(1-\frac{1}{8}\eta\gamma_\alpha\underline{\Lambda}^2(\bC^*)\right)\sum_{k=1}^3(d_k^{(t_0-1)})^2+C_0\eta\textsf{Err}_{\br}^2 \\
&\le \frac{c_1}{\kappa_0^8\bar{r}}+C_0\eta\textsf{Err}_{\br}^2-\frac{1}{8}\eta\gamma_\alpha\underline{\Lambda}^2(\bC^*)\cdot \frac{c_1}{\kappa_0^8\bar{r}}\le\frac{c_1}{\kappa_0^8\bar{r}}
\end{align*}
where the last inequality holds as long as
$$\underline{\Lambda}(\bC^*)/\textsf{Err}_{\br}\ge \sqrt{\frac{8{C_0}\bar{r}}{c_1\gamma_\alpha}}\kappa_0^4 $$
By induction, \eqref{contractioninq} holds for all $t$ and the proof is completed.
\subsection{Proof of Corollary \ref{col:MMLSM_latent}}
By the definition of $\bC^*$ and  $\bar{\bC}$, we have the following estimation on $\underline{\Lambda}(\bC^*)$:
$$\underline{\Lambda}(\bC^*)\ge n^{-1}L^{-1/2}\sigma_{\min}^2(\bar{U})\underline{\Lambda}(\bar{\bC})\sqrt{\min_{1\le j\le L} L_j}\geq \frac{\sqrt{m}}{\kappa_{\bar{U}}}c_*$$
where we've used $\sigma_{\min}(\bar{U})\geq r^{-1/2}\kappa_{\bar U}^{-1}\fro{\bar U}$ and the assumption that the network cluster sizes are balanced. Combined with Theorem \ref{thm:main} and Lemma \ref{lem:xibound}, we have completed the proof.

\subsection{Proof of Theorem \ref{MMLSM:network}}
 By a similar argument to the proof of Corollary \ref{col:MMLSM_latent}, we have
$$\fro{\widehat W -L^{-1/2}W^*\widehat O}^2\le \mathcal{R}$$
where $\widehat O=\argmin_{O\in \mathbb{O}_r}\fro{\widehat W-L^{-1/2}W^*O}$ and 
\begin{align}\label{errorR}
	\mathcal{R}=C_3\zeta_\alpha^2\kappa_{\bar{U}}^4\frac{(r\vee m)\left(2nr+Lm+mr^2\right)}{c_*^2n^2L{m}} 
\end{align}
for some constant $C>0$ depending on $\alpha$. Now denote $\widehat W=[\hat w_1,\cdots,\hat w_L]^\top$ and $W_L^*=[w_1^*,\cdots, w_L^*]^\top$ where $\{\hat w_l\}_{l=1}^L$ and $\{w_l^*\}_{l=1}^L$ are rows of $\widehat W$ and $L^{-1/2}W^*$, respectively. By definition, $W^*_L$ has exactly $m$ distinct rows, denoted by $\{v_j^{*\top}\}_{j=1}^m$. Now we first consider the oracle case such that we put $m$ cluster centers at $\{\widehat O^\top v_j^*\}_{j=1}^m$, and assign nodes in network class $j$ to the cluster centroid $\widehat O^\top v_j^*$. Let $\rm WCSS^*$ denote the objective value (within-cluster sum of squares) of k-means, we have 
\begin{align}\label{WCSS}
	\text{WCSS}^* =\sum_{j=1}^m\sum_{l\in \mathbb{S}_j}\op{\hat w_l-\widehat O^\top  v_j^*}_2^2=\sum_{l=1}^L\op{\hat w_l-\widehat O^\top w_l^*}_2^2=\fro{\widehat W -W_L^*\widehat O}^2\le \mathcal{R}
\end{align}
where $\mathbb{S}_j$ denotes the index set of layers in network class $j$. We also introduce the following index set for layers:
$$J=\left\{l\in [L]:\op{\hat w_l-\widehat O^\top w_l^*}_2\le \frac{\nu}{3} \right\}$$
where $\nu=c\sqrt{m/L}$, where $c$ is the same absolute constant in the  network class sizes condition, i.e. $\abs{\mathbb{S}_j}\ge  cL/m$ for all $j\in[m]$. Then for every layer $l$ in $J^c$, $w_l$ has a distance (in $\ell_2$ norm) at least $\nu/3$ to the centroid $\widehat O^\top w_l^*$. Therefore we have the following estimate:
$$\abs{J^c}\left(\frac{\nu}{3}\right)^2\le \sum_{l\in J^c}\op{\hat w_l-\widehat O^\top w_l^*}_2^2\le \mathcal{R}$$
which leads to
\begin{align}\label{est:Jc}
\abs{J^c}\le 9\mathcal{R}/\nu^2
\end{align}
Now denote $\rm \widehat {WCSS}$ the objective value of k-means algorithm screening on the rows of $\widehat W$, and we give the following claim:
\\For each $j\in[m]$, there exists a unique cluster centroid which has a distance (in $\ell_2$ norm) at most $\nu$ to $\widehat O^\top  v_j^*$.
\\To show it, we first prove the existence using proof by contradiction. Suppose for some $j\in[m]$, the k-means algorithm assigns all centers having distances larger than $\nu$ to $\widehat O^\top  v_j^*$. Then for any $j\in J\cap \mathbb{S}_j$, let $\hat c_j$ denote the closest center to $\hat w_j$, and by triangular inequality we have
$$\op{\hat w_j-\hat c_j}_2\ge \op{\hat c_j-\hat O^\top w_j^*}_2-\op{\hat w_j-\hat O^\top w_j^*}_2\ge \nu -\frac{\nu}{3}=\frac{2\nu}{3}$$
The network class size balance condition suggest $\abs{\mathbb{S}_j}\ge  cL/m$, together with \eqref{est:Jc}, we arrive at
$$\abs{J\cap \mathbb{S}_j}=\abs{\mathbb{S}_j}-\abs{J^c\cap \mathbb{S}_j}\ge \abs{\mathbb{S}_j}-\abs{J^c}\ge \frac{cL}{m}-\frac{9\mathcal{R}}{\nu^2}=O\left(\frac{L}{m}\right)$$
where we use \eqref{errorR} and the condition that $\mathcal R=O\left ((n+L)/(n^2L)\right)\le O(1)$. Also we have
$$
\widehat {\text{WCSS}}\ge \abs{J\cap \mathbb{S}_j}\cdot \op{\hat w_j-\hat c_j}_2^2\gtrsim O(1)
$$
But \eqref{WCSS} implies that $\text{WCSS}^*\leq \mathcal R=O\left ((n+L)/(n^2L)\right)$. Sending $n,L\rightarrow\infty$ such that $L=O(n)$, we get $\text{WCSS}^*\leq \mathcal R\rightarrow 0$, which is  a contradiction. \\
Next we show the uniqueness of such centroid. Observe that for $i\in\mathbb S_k$, $j\in\mathbb S_l$ and $k\ne l$, under the network class sizes balance condition, we have
$$
\op{\widehat O^\top v_k^*-\widehat O^\top v_l^*}_2=\op{ v_k^*-v_l^*}_2=\op{w_i^*-w_j^*}_2\geq c\sqrt{\frac{m}{L}}=3\nu
$$
It follows that one cluster center cannot be within a distance of $\nu$ to $\widehat O^\top v_k^*$ and $\widehat O^\top v_l^*$ simultaneously, which implies that for each $j\in m$ the cluster centroid that has a distance at most $\nu$ to $\widehat O^\top v_j^*$ is unique and we finish the proof of the claim.\\
Now we denote the unique cluster centers in the above claim achieving $\widehat {\text{WCSS}}$ by $\{\hat v_j\}_{j=1}^m$. For each $i\in J\cap \mathbb S_j$,
$$
\op{\hat w_i-\hat v_j}_2\le \op{\hat w_i-\widehat O^\top v_j^*}_2+\op{\widehat O^\top v_j^*-\hat v_j}_2\le \frac{\nu}{3}+\nu=\frac{4\nu}{3}
$$
For any $l$ such that $l\ne j$,
$$
\op{\widehat O^\top v_j^*-\hat v_l}_2\ge\op{v_j^*-v_l^*}_2-\op{\hat v_l-\widehat O^\top v_l^*}_2\ge 3\nu -\nu=2\nu
$$
Thus we have
$$
\op{\hat w_i-\hat v_l}_2\geq \op{\widehat O^\top v_j^*-\hat v_l}_2-\op{\hat w_i-\widehat O^\top v_j^*}_2\geq2\nu-\frac{\nu}{3}=\frac{5\nu}{3}
$$
which implies that the layer $i$ is correctly assigned to the center $\hat v_j$. Therefore, the wrongly clustered layers can only belong to $J^c$, which leads to 
$$
\mathcal{L}(\widehat{\mathbb{S}}, \mathbb{S})\leq\frac{1}{L}\cdot \abs{J^c}\le C_3\zeta_\alpha^2\kappa_{\bar{U}}^4\frac{(r\vee m)\left(2nr+Lm+mr^2\right)}{c_*^2n^2L{m}^2}
$$

\subsection{Proof of Theorem \ref{DLSM:latent}}
Let $W^*_L:=L^{-1/2}W^*$. By the condition that time interval are balanced, we have for any change point $t+1 \in\{t_j\}_{j=1}^{m}$,
$$\op{[W^*_L]_{t+1,:}-[W^*_L]_{t,:}}_2=\sqrt{T_{s_{t+1}}^{-1}+T_{s_{t}}^{-1}}\asymp \sqrt{\frac{T}{m}}$$
By Theorem \ref{MMLSM:network} we have
$$\fro{\widehat W -W_L^*\widehat O}^2\le \mathcal{R}:=C_3\zeta_\alpha^2\kappa_{\bar{U}}^4\frac{(r\vee m)\left(2nr+Lm+mr^2\right)}{c_*^2n^2L{m}}$$
where $\widehat O=\argmin_{O\in \mathbb{O}_r}\fro{\widehat W-W_L^*O}$. Hence for $t+1\in\{t_j\}_{j=1}^{m}$, by triangular inequality we get
\begin{align*}
	c\sqrt{\frac{T}{m}}&\le \op{[W^*_L]_{t+1,:}-[W^*_L]_{t,:}}_2\le \op{[W^*_L]_{t+1,:}-[\widehat W]_{t+1,:}\widehat O^T}_2+\op{[\widehat W]_{t+1,:} -[\widehat W]_{t,:}}_2+\op{[W^*_L]_{t,:}-[\widehat W]_{t,:}\widehat O^T}_2\\
	&\le 2\sqrt{\mathcal R} +\op{[\widehat W]_{t+1,:} -[\widehat W]_{t,:}}_2
\end{align*}
Hence we have
$$\op{[\widehat W]_{t+1,:} -[\widehat W]_{t,:}}_2\ge c\sqrt{\frac{T}{m}}-2\sqrt{\mathcal R}$$
On the other hand, for $t+1\notin\{t_j\}_{j=1}^{m}$, since $\op{[W^*_L]_{t+1,:}-[W^*_L]_{t,:}}_2=0$, we have
$$\op{[\widehat W]_{t+1,:}-[\widehat W]_{t,:}}_2\le \op{[\widehat W]_{t+1,:}-[W^*_L]_{t+1,:}\widehat O}_2+\op{[\widehat W]_{t,:}-[W^*_L]_{t,:}\widehat O}_2\le 2\sqrt{\mathcal R}$$
Then if $nT\ge C\kappa_{\bar U}^2(r\vee m)^{1/2}\left(2nr+Tm+{mr^2}\right)^{1/2}\cdot c_*^{-1}$ for some constant $C>0$ depending only on $\alpha$, we arrive at
$$3\sqrt{\mathcal R}< c\sqrt{\frac{T}{m}}-2\sqrt{\mathcal R}$$
Hence choosing $\epsilon \in[0.4c(T/m)^{1/2},0.6c(T/m)^{1/2}]$ completes the proof.

\end{document}